\documentclass[12pt]{article}

\usepackage{arxiv}

\usepackage{authblk} % for authors

\usepackage[utf8]{inputenc} % allow utf-8 input
\usepackage{booktabs}       % professional-quality tables

\usepackage{epsfig}
\usepackage{graphicx}
\usepackage{rotating}
\usepackage{amsmath,bm,amsthm}
\usepackage{amsfonts}
\usepackage{amssymb}
\usepackage{array}
\usepackage{verbatim}
\usepackage{algorithmic}
\usepackage{rotating}
\usepackage{color}
\usepackage{multirow}
\usepackage{comment}

\makeatletter
\newcommand{\thickhline}{%
    \noalign {\ifnum 0=`}\fi \hrule height 1pt
    \futurelet \reserved@a \@xhline
}
\newcolumntype{"}{@{\hskip\tabcolsep\vrule width 1pt\hskip\tabcolsep}}
\makeatother

\theoremstyle{definition}

\usepackage{hyperref}

\begin{document}

%\begin{center}
\title {Evaluating Point-Prediction Uncertainties in Neural Networks for Drug Discovery}
%\vspace{0.3in}

\author[1]{Ya Ju Fan}
\author[2]{Jonathan E. Allen}
\author[2]{Kevin S. McLoughlin}
\author[3]{Da Shi}
\author[4]{Brian J. Bennion}
\author[4]{Xiaohua Zhang}
\author[4]{Felice C. Lightstone}

\affil[1]{Center for Applied Scientific Computing, Lawrence Livermore National Laboratory, Livermore, California, USA}
\affil[2]{Biological Science and Security Center, Lawrence Livermore National Laboratory, Livermore, California, USA}
\affil[3]{Biomedical Informatics and Data Science Directorate, Frederick National Laboratory for Cancer Research, Frederick, Maryland, USA}
\affil[4]{Physical and Life Sciences Directorate, Lawrence Livermore National Laboratory, Livermore, California, USA}
%\affil[d]{Biochemical and Biophysical Systems Group, Lawrence Livermore National Laboratory, Livermore, CA 94551, USA.}
\date{}
\renewcommand\Affilfont{\itshape\small}

%\author{%
	%Ya Ju Fan\footnote{Center for Applied Scientific Computing, Lawrence Livermore National Laboratory, Livermore, California, USA}%
	%\and Jonathan E. Allen\footremember{GS}{Global Security Computing, Lawrence Livermore National Laboratory, Livermore, California, USA}%
	%\and Kevin S. McLoughlin\footrecall{GS}%
	%\and Da Shi\footnote{Biomedical Informatics and Data Science Directorate, Frederick National Laboratory for Cancer Research, Frederick, Maryland, USA}%
	%\and Brian Bennion\footrecall{GS}%
	%\and Xioahua Zhang\footrecall{GS}%
	%\and Felice Lightstone\footrecall{GS}%
%	}

\maketitle

\begin{abstract}

Neural Network (NN) models provide potential to speed up the drug discovery process and reduce its failure rates. The success of NN models require uncertainty quantification (UQ) as drug discovery explores chemical space beyond the training data distribution. Standard NN models do not provide uncertainty information. Methods that combine Bayesian models with NN models address this issue, but are difficult to implement and more expensive to train. Some methods require changing the NN architecture or training procedure, limiting the selection of NN models. Moreover, predictive uncertainty can come from different sources. It is important to have the ability to separately model different types of predictive uncertainty, as the model can take assorted actions depending on the source of uncertainty. In this paper, we examine UQ methods that estimate different sources of predictive uncertainty for NN models aiming at drug discovery. We use our prior knowledge on chemical compounds to design the experiments. By utilizing a visualization method we create non-overlapping and chemically diverse partitions from a collection of chemical compounds. These partitions are used as training and test set splits to explore NN model uncertainty. We demonstrate how the uncertainties estimated by the selected methods describe different sources of uncertainty under different partitions and featurization schemes and the relationship to prediction error.      

% Their point prediction alone is not sufficient. 

\end{abstract}

%\thispagestyle{empty}
%\vspace{2.0in}
%\vfill \eject
%\thispagestyle{empty}
%\clearpage

\pagestyle{plain}
\pagenumbering{arabic}

%=====================================
%
\section{Background and motivation}
%
%=====================================

Traditional drug discovery examines bioactivities between drug and protein via a high throughput screening experiment~\cite{Cohen2002,Noble2004}. The process is time-consuming and expensive. Deep learning (DL) models try to capture intricate nonlinear relationships between input data (such as drug compounds) and the associated output (such as protein inhibition) for large scale computational screens~\cite{Stevenson2021}. The models then predict the properties of new compounds that help determine a feasible set of compounds for synthesis and evaluation. Computational screens save time and effort, and provide the potential to speed up the drug discovery process and reduce its failure rates~\cite{Vamathevan2019,Jimenez-Luna2020}. However, drug discovery requires exploring chemical space beyond the training data distribution. Predictions on unknown regions are prone to pathological failure. The success of adopting DL methods across a range of drug-design settings requires better communication of uncertainty~\cite{Ching2018, Hie2020}. Detecting regions of chemical space with high uncertainty could help design the experiments to expand a model's applicability domain. 

Recent studies have emphasized the importance and challenges of uncertainty quantification (UQ) in deep learning for drug discovery. Mervin et al. (2020)~\cite{Mervin2020} reviewed classic and modern UQ methods and discussed their usage for drug design (e.g, including the empirical, frequentist and Bayesian approaches). The study points out that the chemoinformatic data being modeled is overall heavily biased with respect to the amount, degree of diversity and distribution of data points. There are issues surrounding data quality and assay variability~\cite{Yang2019}, as well as the skewed proportion of protein-target complexes~\cite{Mervin2016,Bosc2019} and the imbalance between active and inactive compound-target labels~\cite{Rodriguez-Perez2017}. Hence, most models for drug discovery are not able to provide realistic probability estimates while providing single point predictions. A common misleading phenomenon is in the classification models where the predictions for two classes, such as labels of `activity' and `inactivity', come from probability-like fractions. One example is the softmax function used in the last layer of a neural network model, which gives an output value between zero and one. These function values only mean to separate output classes, not to provide confidence of the prediction. Applying explicit uncertainty quantification methods could provide an alternative confidence measure to the classification probability estimates.

Uncertainty of deep learning could come from two sources: data (aleatoric) uncertainty and model (epistemic) uncertainty~\cite{Abdar2021}. The \textit{data uncertainty} reflects a lack of confidence due to the imprecision of molecular measurements. The aleatory concept involves unknown outcomes that can differ each time one runs an experiment under similar conditions~\cite{Fox2011}. Data collected for the drug discovery process could have experimental variabilities due to natural biological changes in the samples, measurement fidelity, instrumentation, sampling procedures, etc. The data uncertainty is irreducible to the model and is an inherent property of their distribution. 

In contrast, \textit{model uncertainty} measures the uncertainty in model parameters given the training data. Particularly, inadequate knowledge contributes to the model uncertainty. This is often the case for the chemoinformatic data where there is a limited number of drug compounds available with unbalanced activity classes for a complex biological process. UQ methods for deep neural networks (NN) that evaluate how changes in model parameters effect the NN predictions are aimed at capturing the model uncertainty. Bayesian neural networks (BNN)~\cite{Kononenko1989, MacKay1992} quantify posterior uncertainty on NN model parameters (i.e. weights) and express predictions in terms of expectations with respect to this posterior distribution~\cite{Bishop1995}. Other methods generate a set of point predictions and use its mean and variance to represent the optimal prediction and its corresponding uncertainty. Monte-Carlo dropout~\cite{Gal2016} passes a test sample multiple times through the NN model and generate a collection of such predictions. For each iteration, the model assigns the value zero to a fraction of randomly selected weights. Similarly, deep ensemble~\cite{Lakshminarayanan2017} and bootstrap~\cite{Du2021} train multiple models and compute the mean and spread of the ensemble. 

Apart from data uncertainty and model uncertainty, Malinin and Gales (2018) presented distributional uncertainty as a separate source of uncertainty~\cite{Malinin2018}. \textit{Distributional uncertainty} occurs when the test data is foreign to the model due to mismatch between the training and test distributions. Applicability domain (AD) estimates whether a model's prediction for a chemical compound is applicable based on the model's training set properties. Distance-based methods used to evaluate AD play a similar role as distributional uncertainty modeling methods~\cite{Wang2021}. 

Residual estimation with an I/O kernel (RIO)~\cite{Qiu2020} directly estimates prediction residuals using modified Gaussian Processes (GP). GP models offer a mathematically grounded approach to reason about the predictive uncertainty~\cite{Rasmussen2006}. RIO uses a new composite I/O kernel that makes use of both inputs and outputs of the NN, meaning that the method examines both the distance to the nearest training data and the prediction errors on the training set. It provides predictive uncertainty estimation without modifying the NN training or formulation. Hie et al. 2020~\cite{Hie2020} demonstrated that RIO uncertainty estimation enables successful iterative learning across a broad spectrum of experimental scales for biological discovery and design.  

It is important to have the ability to separately model the different types of predictive uncertainty, as the model can take assorted actions depending on the source of uncertainty~\cite{Malinin2018}. In this article, we present a carefully designed experiment with publicly available drug data to gain insights into uncertainty quantification methods. We employ the calculated end-point binding free energy with MM/GBSA (the molecular mechanics generalized Born surface area) as the response values for the NN models. The simulated MM/GBSA scores are not from the experimental measurements, and hence reduce the aleatoric uncertainty, which allows the study to focus on modeling epistemic uncertainty. We select the UQ methods that examine different sources of uncertainty. Additionally, they can be directly applied to any standard NN without having to modify the model training formulation. We use the ATOM Modeling PipeLine (AMPL)~\cite{Minnich2020} developed by the Accelerating Therapeutics for Opportunities in Medicine (ATOM) Consortium to provide a rigorous pipeline for training and evaluating drug discovery oriented NN models. We demonstrate how the selected UQ methods provide uncertainty estimations on point predictions of NN models. Since the goal of UQ is to detect unanticipated imprecision of model predictions, we inspect how the UQ methods reflect prediction errors. 

This paper is organized as follows. First, we collect related work in Section~\ref{sec:related}. Next, we describe the selected uncertainty quantification methods and how they present different sources of uncertainty in Section~\ref{sec:methods}. In Section~\ref{sec:design}, we explain the preparation of the chemical compound inputs, NN model building and the experimental design to evaluate the UQ methods. Finally, we present the experimental results in Section~\ref{sec:results} and conclude our findings in Section~\ref{sec:conclusion}.

\section{Related work}
\label{sec:related}
%
%=====================================

There has been significant demand in quantifying prediction uncertainty for DNN models, especially for drug discovery where mistakes may be expensive. Hie et al. (2020) ~\cite{Hie2020} leveraged Gaussian process-based uncertainty prediction to identify and validate experimental compounds with nanomolar affinity for diverse kinases and whole-cell growth inhibition of Mycobacterium tuberculosis. They showed that the GP uncertainty estimation enables successful iterative learning across a broad spectrum of experimental scales. Scalia et al. (2020)~\cite{Scalia2020} compared scalable UQ methods, including MC-dropout, Deep Ensembles and bootstrapping, for graph convolutional neural networks (GCNN), designed for deep learning-based molecular property prediction. They introduced a set of quantitative criteria, including ranking-based methods and uncertainty calibration methods, to capture different uncertainty aspects. Wang et al. (2021) ~\cite{Wang2021} combine both distance-based and Bayesian UQ approaches together for improved uncertainty quantification in QSAR (Quantitative Structure-Activity Relationship) regression modeling. The hybrid method quantitatively assesses the ranking and calibration ability of the selected UQ methods, including applicability domain (AD) methods, mean-variance estimation of the graph convolutional neural networks~\cite{Scalia2020} and deep ensembles~\cite{Lakshminarayanan2017}. 

% Add other uq methods that need to change model formulation

There has been notable progress made on predictive uncertainty for deep learning through the formulation of neural networks. One class of approaches stems from the combination of a Bayesian approach and neural networks~\cite{MacKay1992, Hinton1993, Neal1996,Wilson2016,Lakshminarayanan2017}. All such methods require significant modifications to the model infrastructure and training procedure. Malinin and Gales (2018) developed Prior Networks for modeling predictive uncertainty, which explicitly models distributional uncertainty~\cite{Malinin2018}. Tang and de Jong (2019)~\cite{Tang2019} developed marginalized graph kernel specifically for computing similarity between molecules. The framework employs GP regression to perform prediction on the atomization energy of molecules.

We investigate a set of the UQ methods separately, which examine different sources of uncertainty. These methods also provide UQ estimations without requiring specific NN formulations, allowing broader applications.

\section{Uncertainty quantification methods for neural networks}
\label{sec:methods}
%
%==================================================
%==================================================

We select uncertainty quantification approaches for deep neural networks that do not limit the choice of model. Consider a neural network model with $L$ layers. We denote $\mathbf{W}_i$ as the NN's weight matrices for each layer $i=1, 2, \dots, L$. We denote $y_i$ as the observed output corresponding to input $\mathbf{x}_i$, $i=1, 2, \dots, n$, where $n$ is the number of data points.  Let $\mathbf{X}, \mathbf{y}$ be the input and output sets and let the training dataset $\mathcal{D} = (\mathbf{X}, \mathbf{y}) = \{(\mathbf{x}_i,y_i)\}_{i=1}^{n}$. The predictive probability of a NN model can be parameterized as

\begin{equation}
P(y|\mathbf{x},\mathcal{D}) = \int \underbrace{P(y|\mathbf{x},\omega)}_{\text{Data}} \underbrace{p(\omega|\mathcal{D})}_{\text{Model}}\mathrm{d}\omega.
\label{eq:nnpred}
\end{equation}

\noindent In a Bayesian framework the predictive uncertainty of a NN model $P(y|x,\mathcal{D})$ trained on a finite dataset $\mathcal{D}$ will result from data (aleatoric) uncertainty and model (epistemic) uncertainty as shown in (\ref{eq:nnpred}). The posterior distribution over responses $y$ given a set of model parameters $\omega$ describes a model's estimates of data uncertainty, and the posterior distribution over the parameters given data describes model uncertainty~\cite{Malinin2018}.

%======================================
\subsection{Monte-Carlo dropout (MC-dropout)}
%======================================

Deep neural networks contain multiple nodes and layers that try to learn complicated relationships between the inputs and outputs. Dropout is a method used to prevent overfitting and lower generalization error for NN models~\cite{Srivastava2014}. Dropout means temporarily removing a node from the network along with all its incoming and outgoing connections, which we can simply set zero on the weight of the node. The choice of which nodes to drop is random. In most cases, we choose a fixed value to indicate the fraction of the nodes to drop. Gal and Ghahramani (2016) further applied the concept to generate ensembles for evaluating model uncertainty. 

Let $\mathbf{W}_i$ be a random matrix for each layer $i$ and  set $\omega = \{\mathbf{W}_i\}_{i=1}^{L}$. In Equation (\ref{eq:nnpred}) obtaining the true posterior $p(\omega|\mathcal{D})$ using Bayes' rule is intractable, and it is necessary to use either an explicit or implicit variational approximation $q(\omega)$~\cite{Graves2011, Kingma2015, Louizos2016}. Monte-Carlo (MC) dropout utilizes the $q(\omega)$ as a distribution over matrices whose columns are randomly set to zero (called dropout).  Furthermore, the integral in equation (\ref{eq:nnpred}) for the predictive probability is also intractable for neural networks. MC dropout employs sampling to approximate it. As shown in Equation (\ref{eq:dropout}) each term in the sum is approximated by Monte-Carlo integration with a single sample  $\omega^{(i)} \sim q(\omega)$ to obtain an unbiased estimate~\cite{Gal2016}. 

\begin{equation}
P(y|x,\mathcal{D}) \approx \displaystyle \frac{1}{n}\sum_{i-1}^{n} P(y|x,\omega^{(i)}), \quad \text{where } \omega^{(i)} \sim q(\omega).  
\label{eq:dropout}
\end{equation} 

\noindent Training a neural network with dropout is as training a collection of thinned networks with extensive weight sharing. The process does not change the dropout NN model itself. In testing we use the original model by simply cutting off the node's weights. MC dropout estimates the predictive mean and predictive uncertainty by collecting the results of stochastic forward passes through the NN model. 

%==================================
\subsection{Applicability domain (AD)}
%==================================

The Applicability Domain (AD) assessment is based on the numerical vector representation of chemical compounds in the training set. We select two AD methods that are distance-based and suitable for novelty (or outlier) detection. 

\subsubsection{Empirical distance distribution (AD-DD)}
For distance-based novelty detection we need a threshold to decide whether or not the unseen object is actually novel~\cite{Mathea2016}. The empirical distance distribution of the training set molecules can help establish a threshold, using the $1-\alpha$ quantile (for $\alpha = 0.01,0.05$ or $0.1$) of the $k$-nearest neighbors within the training set distance distribution~\cite{Knorr97}. We collect the mean distances from every data point to its $k$-nearest neighbors in the training set. We fit a normal distribution on these mean distances and obtain its total mean, $\mu_{\text{knn}}$, and standard deviation, $\sigma_{\text{knn}}$. For a new chemical compound $x$, we compute its mean distance to its $k$-nearest neighbors in the training set $\overline{d}_{\text{knn}}(x)$. The score of the applicability domain using the empirical distance distribution is the rectified Z-score of the mean $k$-nearest neighbor distance of the new point based on the distribution of the training set:
\begin{equation}
\rho_{D}(x) = \max \left\{ 0, \frac{\overline{d}_{\text{knn}}(x) - \mu_{\text{knn}}}{\sigma_{\text{knn}}} \right\}.
\label{eq:ADDD}
\end{equation}

\subsubsection{Local density (AD-LD)}

Denote the distance of a data point $x$ to its $k^{th}$-nearest neighbor in the training set $NN^{train}_k(x)$ as $\|x - NN^{train}_k(x) \|$. The local density is inversely related to the distance. The higher the local density is, the lower the distance is to the neighbors. The score of the applicability domain using local density \cite{Tax98outlierdetection} is:
\begin{equation}
\rho_{L}(x) =  \displaystyle \frac{ \frac{1}{k}\sum_{i=1}^k \|x - NN_i^{train}(x) \| }{ \frac{1}{k^2}\sum_{i=1}^k\sum_{j=1}^k    \|NN_i^{train}(x) - NN_j^{train}(NN_i^{train}(x)) \| }.
\label{eq:ADLD}
\end{equation}

\noindent The denominator is the average distance from the nearest neighbors $NN^{train}(x)$ to their nearest neighbors in the training set $NN^{train}(NN^{train}(x))$.

%==================================================
\subsection{Residual estimation with an I/O kernel (RIO)}
%==================================================

Residual estimation with an I/O kernel (RIO) is a Gaussian processes (GP) method designed to be applied on top of any trained NN model~\cite{Qiu2020}. RIO quantifies the uncertainty in the point-predictions of NN models without retraining or modifying any component of them. 

Let $\hat{y}_i$ be the point prediction of a NN model given $\mathbf{x}_i$. RIO models the residuals between observed outcomes $y$ and NN predictions $\hat{y}$ using GP with a composite kernel. The residuals $\{r_i\}_{i=1}^{n}$ between observed outcomes and NN predictions on the training dataset $\mathcal{D}$ is 
\begin{equation}
r_i = y_i - \hat{y}_i, \quad \text{for } i=1, 2, \dots, n.
\end{equation} 
\noindent Let $\mathbf{r}$ be the vector of all residuals and $\hat{\mathbf{y}}$ be the vector of all NN predictions. RIO trains a GP with a composite kernel assuming $\mathbf{r} \sim \mathcal{N}(0, \mathbf{K}_{\theta}((\mathbf{X},\hat{\mathbf{y}}),(\mathbf{X},\hat{\mathbf{y}})) + \sigma_n^2\mathbf{I})$, where $\mathcal{N}$ denotes a multivariate Gaussian distribution with mean 0 and covariance matrix $\mathbf{K}_{\theta}((\mathbf{X},\hat{\mathbf{y}}),(\mathbf{X},\hat{\mathbf{y}})) + \sigma_n^2\mathbf{I})$, and $\sigma_n^2$ is the noise variance of observations. The composite kernel $\mathbf{K}_{\theta}((\mathbf{X},\hat{\mathbf{y}}),(\mathbf{X},\hat{\mathbf{y}}))$ is an $n\times n$ covariance matrix at all pairs of training points whose elements are:
\begin{equation}
k_{\theta}\big( (\mathbf{x}_i,\hat{y}_i),(\mathbf{x}_j,\hat{y}_j)\big) = k_{\theta_{\text{in}}}(\mathbf{x}_i,\mathbf{x}_j)+ k_{\theta_{\text{out}}}(\hat{y}_i,\hat{y}_j), \quad \text{for }i, j = 1, 2, \dots, n.
\end{equation}
\noindent This kernel design enables the evaluation on both the novelty of a data point and the prediction residual. The composite kernel is the sum of two kernels, one for the NN input data and the other for the NN output data. RIO applies the composite kernel as the covariance function of GP. The GP optimizes the hyperparameters of the covariance function, $\theta{\text{in}}$ and $\theta{\text{out}}$, by maximizing the log marginal likelihood $\log p(\mathbf{r}|\mathbf{X},\hat{\mathbf{y}})$.

%============================================
%
\section{Experimental design} 
\label{sec:design}
%
%============================================
%

%The database primarily captures the association between a ligand and a biological target in the form of an experimentally measured activity end-point, e.g. half maximal inhibitory concentration (IC50).

We randomly select 9000 drug compounds available from the ChEMBL database~\cite{Mendez2018,Davies2015}, a large, open-access bioactivity data source. The drug compounds  are presented in the SMILES (simplified molecular-input line-entry system) strings, which describes the structure of the chemical compounds in the dataset. We use two featurization schemes to map the SMILES strings into model input features. We first compute the chemical descriptors using Molecular Operating Environment (MOE) software~\cite{moe2022}. Many MOE descriptors were strongly correlated with each other due to the fact that they scaled with molecular size, as measured by the total number of atoms, denoted as $\texttt{a\_count}$ ~\cite{McLoughlin2021}. We replace all descriptors, $d$, having Pearson correlation $r(d, \texttt{a\_count}) > 0.5$ with the computed $d/\texttt{a\_count}$ to remove the dependency. We further eliminate descriptors that are duplicated, have constant values or are linear functions of other descriptors. The final set of the MOE descriptors contain 306 features. The other featurization scheme that we also compute is the extended connectivity fingerprint (ECFP4). We use RDKit~\cite{rdkit} to generate the ECFP4 bit vectors with length 1024.

\begin{table}[htbp]
\scriptsize
\caption{For visualization we select the subset of MOE features whose values are highly correlated to the MM/GBSA scores. There are 22 features having their Pearson correlation coefficient larger than 0.5 or lower than -0.5. }
\vspace{0in}
\begin{center}
\begin{tabular}{ccr}
\hline
\textbf{Rank } & \textbf{MOE feature} & \textbf{Correlation} \\
\hline
1&BCUT\_SMR\_0\_per\_atom&-0.738\\
2&BCUT\_SLOGP\_0\_per\_atom&-0.735\\
3&GCUT\_SMR\_3\_per\_atom&0.729\\
4&GCUT\_SLOGP\_3\_per\_atom&0.725\\
5&BCUT\_SMR\_3\_per\_atom&0.722\\
6&VAdjEq\_per\_atom&0.721\\
7&a\_count&-0.694\\
8&vsurf\_R\_per\_atom&0.688\\
9&vsurf\_CW1\_per\_atom&0.686\\
10&PEOE\_RPC-\_per\_atom&0.682\\
11&RPC-\_per\_atom&0.674\\
12&opr\_leadlike\_per\_atom&0.668\\
13&vsurf\_G\_per\_atom&0.664\\
14&balabanJ\_per\_atom&0.662\\
15&weinerPath&-0.658\\
16&VAdjMa\_per\_atom&0.644\\
17&VDistMa\_per\_atom&0.621\\
18&VDistEq\_per\_atom&0.568\\
19&ast\_violation&-0.552\\
20&ast\_fraglike&0.538\\
21&std\_dim1&-0.528\\
22&pmi\_per\_atom&-0.506\\
\hline
\end{tabular}
\label{tb:topCorrMOEfeatures}
\end{center}
\end{table}
\begin{figure}[thp]
\begin{center}
\includegraphics[width = 0.5\columnwidth, trim=0.2cm 0.3cm 0.45cm 0.3cm, clip=false]{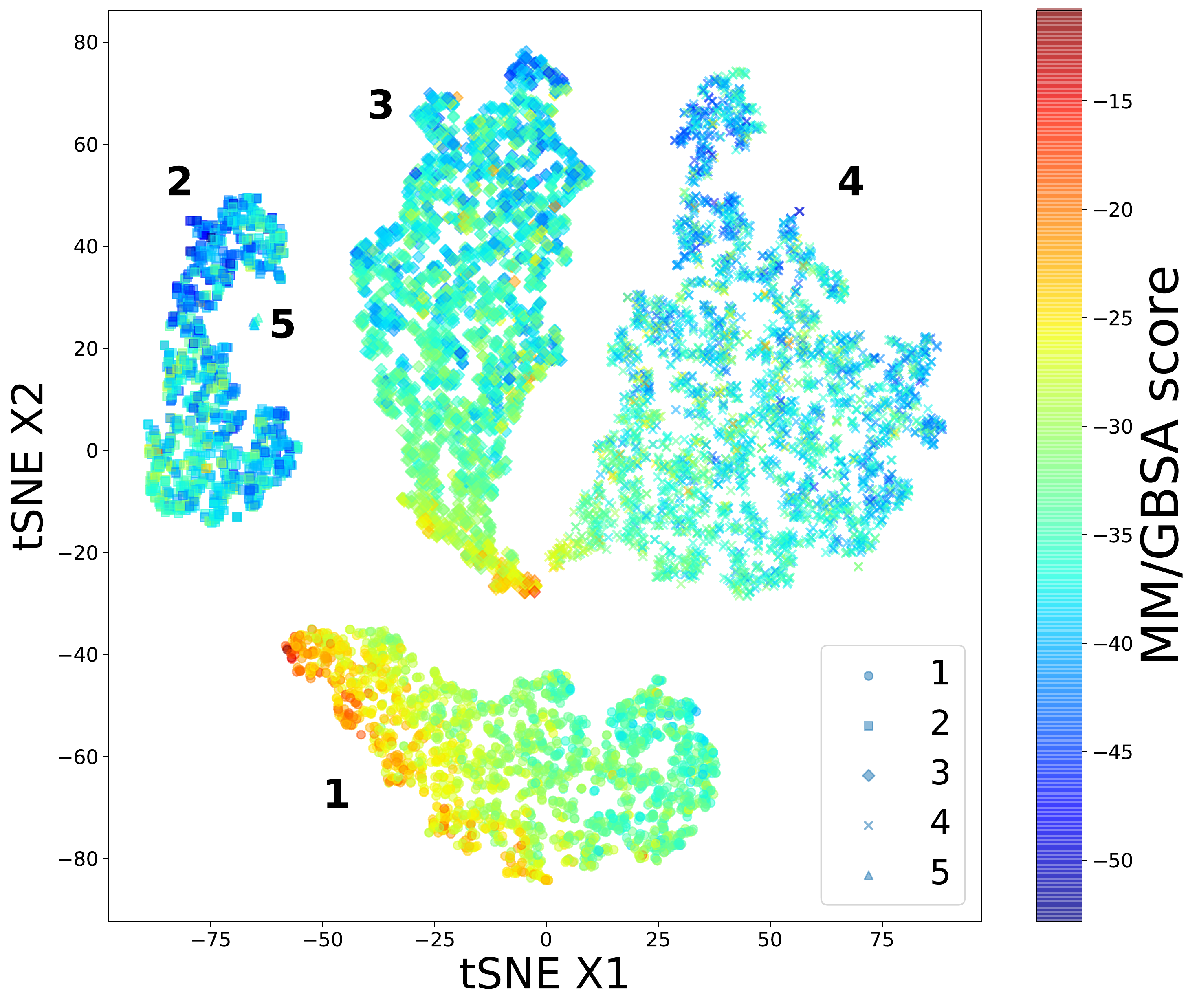}
\end{center}
\vspace{0in}
\caption{\label{fig:DBclusters} Clusters}
\end{figure}
\begin{figure}[thp]
\begin{center}
\includegraphics[width = 0.5\columnwidth, trim=0.2cm 0.3cm 0.45cm 0.3cm, clip=false]{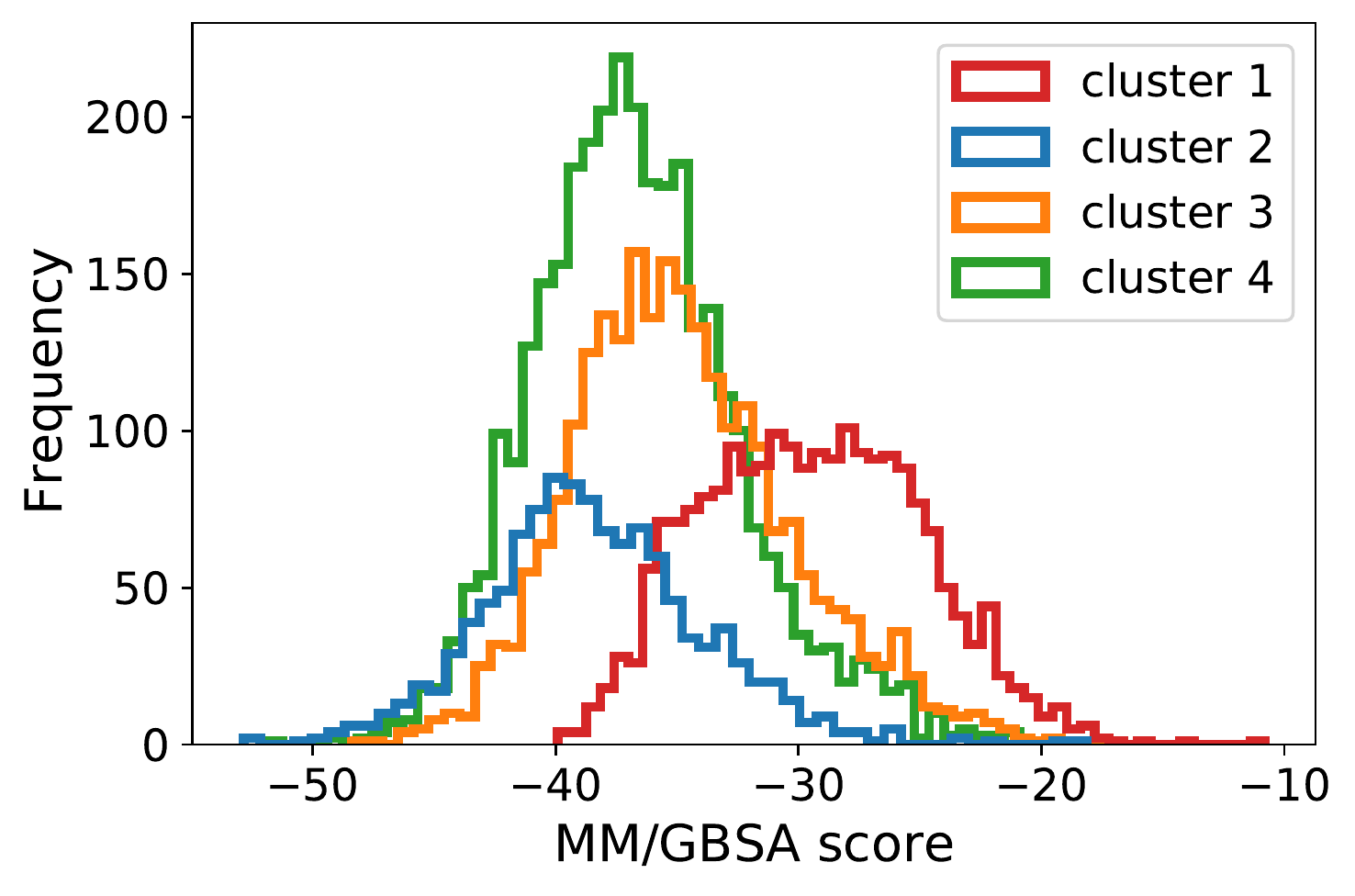}
\end{center}
\vspace{0in}
\caption{\label{fig:DBclusterFreq} MM/GBSA score distribution in each cluster.}
\end{figure}
%

\begin{comment}
%
\begin{figure}[th]
\begin{center}
\begin{tabular}{cc}
\multicolumn{2}{c}{Model trained with MOE descriptors}\\
\includegraphics[width = 0.45\columnwidth, trim=0.3cm 1.9cm 0.45cm 0.3cm, clip=false]{GBSA_DBclusters.pdf}
&\includegraphics[width = 0.45\columnwidth, trim=0.3cm 1.9cm 0.3cm 0.3cm, clip=false]{GBSA_DBcluster_frequency.pdf}\\
\end{tabular}
\end{center}
\vspace{-0.17in}
\caption{\label{fig:DBclusters} Clusters and the MM/GBSA score distribution.}
\end{figure}
%
\end{comment}

We utilize the molecular mechanics generalized Born surface area (MM/GBSA) continuum solvent approach to rescore the binding affinities of the chemical compounds~\cite{Massova2000, Mongan2007}. The computed MM/GBSA scores are not from experimental measurements but rather estimate binding affinity between a compound and a target protein. The target protein used for this study is the main spike protein for SARS-CoV-2. Applying them as the responses of the NN models reduces their aleatoric uncertainty. Since the aleatoric uncertainty is irreducible to the model, making it more controllable could help observing other uncertainty sources. We have also observed that the MOE descriptors are highly correlated to the MM/GBSA scores in the randomly selected 9000 drug compounds. Among the total 306 MOE features there are 22 features having correlation larger than 0.5 as shown in Table~\ref{tb:topCorrMOEfeatures}. 

To better understand how the uncertainty quantification methods work for drug discovery tasks, we make use of the known correlation between the MOE descriptors and the MM/GBSA scores to design the experiments. We visualize the correlation in the dataset using the t-distributed stochastic neighbor embedding (t-SNE). We project the top 22 correlated MOE features employing ten principle components before computing the two coordinates of the t-SNE, according to the suggested usage of the method~\cite{tsne}. Figure~\ref{fig:DBclusters} displays the two-dimensional t-SNE scatter plot of the dataset. 

It appears that the dataset consists of five clusters. The color on the data points represents the MM/GBSA scores where a more negative value indicates stronger binding. Cluster 2, 3 and 4 marked on the plot contain data points with large negative MM/GBSA scores, while cluster 1 and 3 contain more high value MM/GBSA scores. Figure~\ref{fig:DBclusterFreq} shows the distribution of the MM/GBSA scores for each cluster. 

We split the dataset into training set and test set based on the clusters. The numbers of the data points in the clusters are 2132, 2455, 3252, 1154 and 7, respectively, from cluster 1 to cluster 5. We exclude cluster 5 due to its small size. We randomly select 1000 data points in each cluster as the training set, another 100 data points as the validation set and the rest as the test set. We apply the same splits to both the MOE features and the ECFP features, creating eight splittings of training, validation and test sets (from four clusters and two featurization methods). We include all of the 306 MOE features in the dataset as the 22 high correlation features discussed earlier are for visualization purposes only. 

%molecular mechanica l and generalized Born/Surface Accessible (MM/MM/GBSA)
%molecular mechanics energies combined with generalized Born and surface area continuum solvation (MM/PBSA and MM/MM/GBSA) 

We train the Neural Network models using a data-driven modeling pipeline, AMPL, developed by our group at the ATOM Consortium~\cite{Minnich2020}. The underlying models were implemented with the DeepChem package~\cite{deepchem}. The architecture of the neural networks consists  of one, two or three fully connected hidden layers. Each layer contains varying numbers of rectified linear unit (ReLU) nodes. During training and evaluation we randomly drop out 30\% of nodes to avoid overfitting. To optimize the performance of neural networks, we run hyperparameter searches, varying the numbers of hidden layers, numbers of nodes per layer, and learning rates. We evaluate an average of 1500 regression models for each training set and select the best model for further analysis on quantifying the point-prediction uncertainties. We select the best model parameters based on validation set performance following standard machine learning practices. We evaluate the performance of regression models using the coefficient of determination ($R^2$)~\cite{Barrett1974} defined as 
\[
R^2 = 1 - \displaystyle\frac{\sum_{i=1}^n (y_i-\hat{y}_i)^2}{\sum_{i=1}^n (y_i-\bar{y}_i)^2}
\]
\noindent where $y_i$ is the actual MM/GBSA score for a compound $i$, $\bar{y}$ is the average value of the actual scores and $\hat{y}_i$ is the predicted value of $y_i$. An $R^2$ close to 1 implies an almost perfect case when modeled values nearly match the observed values. An $R^2$ equal to 0 implies a baseline model, which always predicts the mean. A negative $R^2$ indicates that a model has worse predictions than the baseline.

\begin{comment}
%===============================
\subsection{Public datasets}
%===============================
We use combined datasets curated by the ATOM Consortium from publicly available sources: ChEMBL~\cite{Mendez2018}, DTC (Drug Target Commons)~\cite{Tang2018} and ExCAPE-DB~\cite{Sun2017ExCAPE}. 

An IC50 measurement tells us the concentration where a drug is able to inhibit a particular biological process by 50\%. In order to capture the logarithmic nature of compound potency, we convert IC50 values to pIC50 values, where $\text{pIC50} = -\log_{10}(\text{IC50})$ when IC50 is expressed as molar concentration.
\end{comment}

%==========================
%
\section{Experimental results}
\label{sec:results}
%
%==========================

%--------------------------------------------
%
\subsection{Model performance}
%
%--------------------------------------------

\begin{comment}
%
\begin{table}[htbp]
\scriptsize
\caption{Model performance in $R^2$. }
\vspace{0in}
\begin{center}
\begin{tabular}{crrrrrr}
\toprule
\textbf{} & \multicolumn{3}{c}{\textbf{MOE}} &\multicolumn{3}{c}{\textbf{ECFP}} \\
\cmidrule(r){2-4} \cmidrule{5-7}
\textbf{Training set} & \textbf{Training} & \textbf{Validation} & \textbf{Test} & \textbf{Training} & \textbf{Validation} & \textbf{Test} \\  
\hline
Cluster 1 & \bf{0.869} & {0.872} & 0.540       & \bf{0.898}   & {0.627} & 0.201\\
Cluster 2 & 0.748       & 0.383   & -0.968       & 0.412         & 0.199   & -0.524\\
Cluster 3 & 0.644       & 0.567   & \bf{0.664}  & 0.618         & 0.614   & \bf{0.353} \\
Cluster 4 & 0.739       & 0.477   &-0.103        & 0.437         & 0.350   & 0.076 \\
\bottomrule
\end{tabular}
\label{tb:performance}
\end{center}
\end{table}
%
\end{comment}

%
\begin{table}[tbhp]
\scriptsize
\caption{Model performance in $R^2$ on each cluster using MOE descriptors. }
\vspace{0in}
\begin{center}
\begin{tabular}{crrrr}
\toprule
\textbf{MOE} & \multicolumn{4}{c}{\textbf{Test set}} \\
\cmidrule(r){2-5} 
\textbf{Training set} & {Cluster 1} & {Cluster 2} & {Cluster 3} & {Cluster 4} \\  
\hline
Cluster 1 &\textbf{0.841}	&0.088	&0.485	&0.237\\	
Cluster 2 &-5.898	&\textbf{0.721}	&-0.652	&-0.029	\\
Cluster 3 &\textbf{0.685}	&0.256	&0.580	&0.363	\\
Cluster 4 &-3.332	&0.182	&0.323	&\textbf{0.455}	\\
\bottomrule
\end{tabular}
\label{tb:perfClusterMoe}
\end{center}
\end{table}
\begin{table}[tbhp]
\scriptsize
\caption{Model performance in $R^2$ on each cluster using ECFP features. }
\vspace{0in}
\begin{center}
\begin{tabular}{crrrr}
\toprule
\textbf{ECFP} & \multicolumn{4}{c}{\textbf{Test set}} \\
\cmidrule(r){2-5} 
\textbf{Training set} & {Cluster 1} & {Cluster 2} &{Cluster 3} &{Cluster 4} \\  
\hline
Cluster 1 &\textbf{0.767}	&-0.834	&0.222	&-0.317\\
Cluster 2 &-3.917	&\textbf{0.399}	&-0.453	&0.046\\
Cluster 3 &0.036	&-0.497	&\textbf{0.455}	&-0.054\\
Cluster 4 &-1.885	&-0.053	&0.159	&\textbf{0.276}\\
\bottomrule
\end{tabular}
\label{tb:perfClusterECFP}
\end{center}
\end{table}
%

\begin{comment}
%
\begin{figure}[th]
\begin{center}
\small
\begin{tabular}{cccc}
\multicolumn{4}{c}{Models trained with MOE descriptors}\\
\includegraphics[width = 0.24\columnwidth, trim=0.3cm 1.9cm 0.45cm 0.3cm, clip=false]{GBSA_DBset1_predError.pdf}
&\includegraphics[width = 0.24\columnwidth, trim=0.3cm 1.9cm 0.3cm 0.3cm, clip=false]{GBSA_DBset2_predError.pdf}
&\includegraphics[width = 0.24\columnwidth, trim=0.3cm 1.9cm 0.45cm 0.3cm, clip=false]{GBSA_DBset3_predError.pdf}
&\includegraphics[width = 0.24\columnwidth, trim=0.3cm 1.9cm 0.3cm 0.3cm, clip=false]{GBSA_DBset4_predError.pdf}\\
\multicolumn{4}{c}{Models trained with ECFP}\\
\includegraphics[width = 0.24\columnwidth, trim=0.3cm 1.9cm 0.45cm 0.3cm, clip=false]{GBSA_DBset1_ecfp_predError.pdf}
&\includegraphics[width = 0.24\columnwidth, trim=0.3cm 1.9cm 0.3cm 0.3cm, clip=false]{GBSA_DBset2_ecfp_predError.pdf}
&\includegraphics[width = 0.24\columnwidth, trim=0.3cm 1.9cm 0.45cm 0.3cm, clip=false]{GBSA_DBset3_ecfp_predError.pdf}
&\includegraphics[width = 0.24\columnwidth, trim=0.3cm 1.9cm 0.3cm 0.3cm, clip=false]{GBSA_DBset4_ecfp_predError.pdf}\\
\end{tabular}
\end{center}
\vspace{-0.17in}
\caption{\label{fig:predError} Prediction errors of the eight NN models. The number on the plot indicates the training cluster.}
\end{figure}
%
\end{comment}

The best regression model after hyperparameter tuning, measured by $R^2$ scores, is kept for each cluster and each type of chemical representation (MOE or ECFP). Table~\ref{tb:perfClusterMoe} and Table~\ref{tb:perfClusterECFP} collect the separated test set performances of the best model trained on each cluster. Recall that we take 1000 randomly selected data points from each cluster for model training. We use the rest of the data points as the test set in the cluster to show how each model perform on single clusters. 

We found that the models trained on MOE descriptors perform better than the ones trained on ECFP. The clusters used for the training and test sets are created from the most correlated MOE descriptors. Since the clusters are not explicitly separated by their ECFP distances, there would be more similarity between the training and test sets in ECFP. It is surprising that the ECFP models do more poorly in the cross cluster prediction task. 

The results also indicate that the NN models perform the best in their own clusters, except on the model trained on cluster 3 with MOE descriptors. Data points in cluster 3 cover a wider range of the MM/GBSA scores. Interestingly the model trained on cluster 3 using MOE descriptors performed the best on cluster 1 with $R^2$ at 0.685, while performed the second best on the test set from its own cluster with $R^2$ at 0.580.  We also observed that the models trained on cluster 2 and cluster 4 do not perform well on predicting data points in other single clusters. Overall, most models have worst performance on their test set, and best performance on their training set. This aligns with what we expect when designing the splits of the data. Chemical compounds outside a cluster tend to be more different from those in the cluster. It is a challenge to the model trained on a cluster to test its performance outside the cluster. Next, we investigate how the selected UQ methods capture such challenges.

%Models trained with ECFP features have worse $R^2$ scores than the models using MOE features on cluster 2 and cluster 4. The model trained on cluster 1 gives the best training performance with $R^2$ at 0.869 and 0.898 using MOE and ECFP features, respectively. Reflecting on the best prediction performance on the test set with $R^2$ at 0.664, the model trained on cluster 3 with MOE features has overall lower point-prediction errors than the other models. 

%Figure~\ref{fig:predError} displays the point-prediction errors of the eight regression models. 
%\clearpage

%--------------------------------------------------
%
\subsection{Uncertainty evaluation}
%
%--------------------------------------------------

%-------------------------
% Cluster 1
%-------------------------

%
\begin{figure}[th]
\begin{center}
\small
\begin{tabular}{cc}
{\large (a) Model trained on cluster 1 with MOE} & {\large (b) Model trained on cluster 1 with ECFP}\\ 
\includegraphics[width = 0.48\columnwidth, trim=0.3cm 0cm 0.3cm 0cm, clip=false]{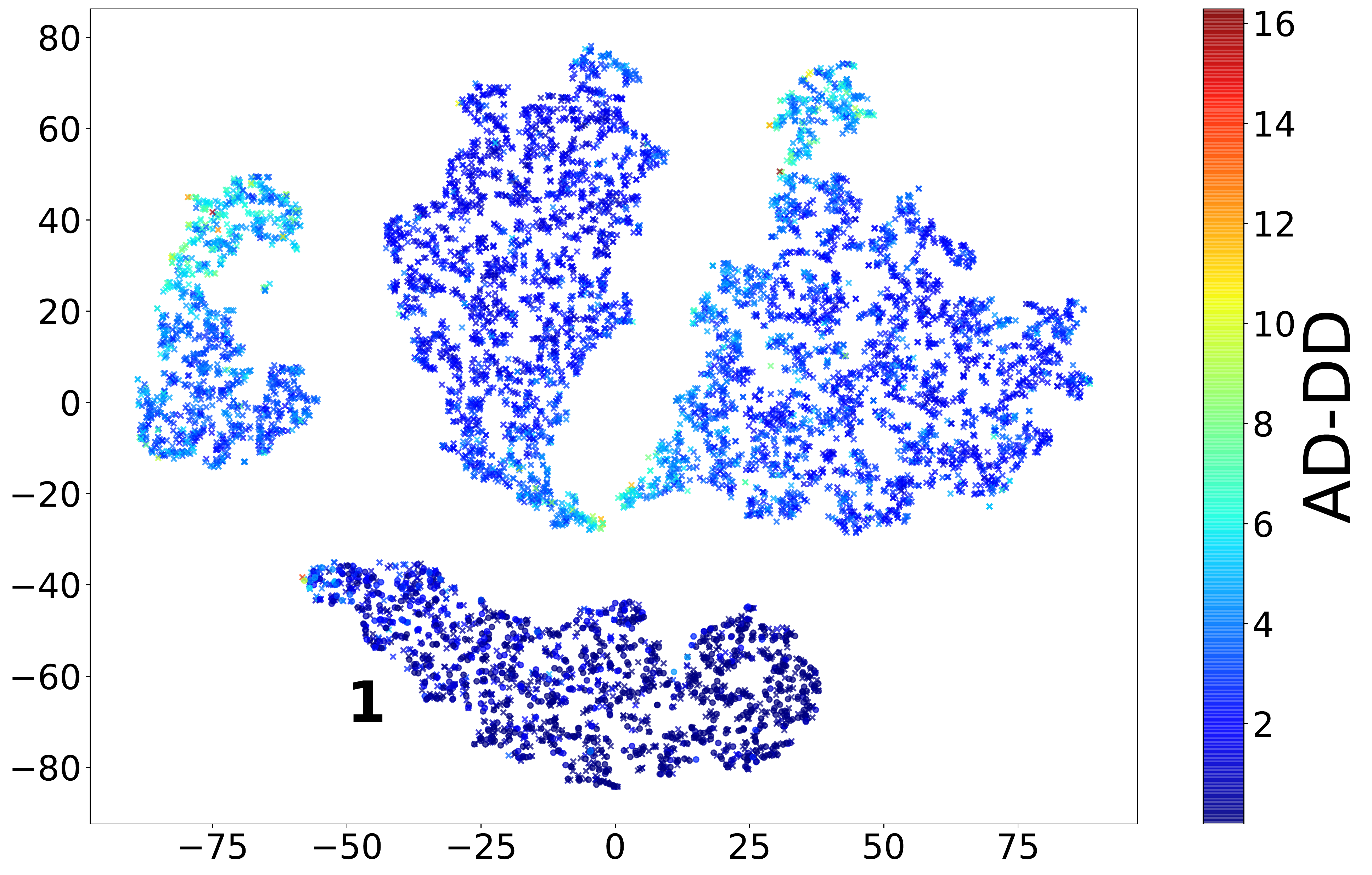}
& \includegraphics[width = 0.48\columnwidth, trim=0.3cm 0cm 0.3cm 0cm, clip=false]{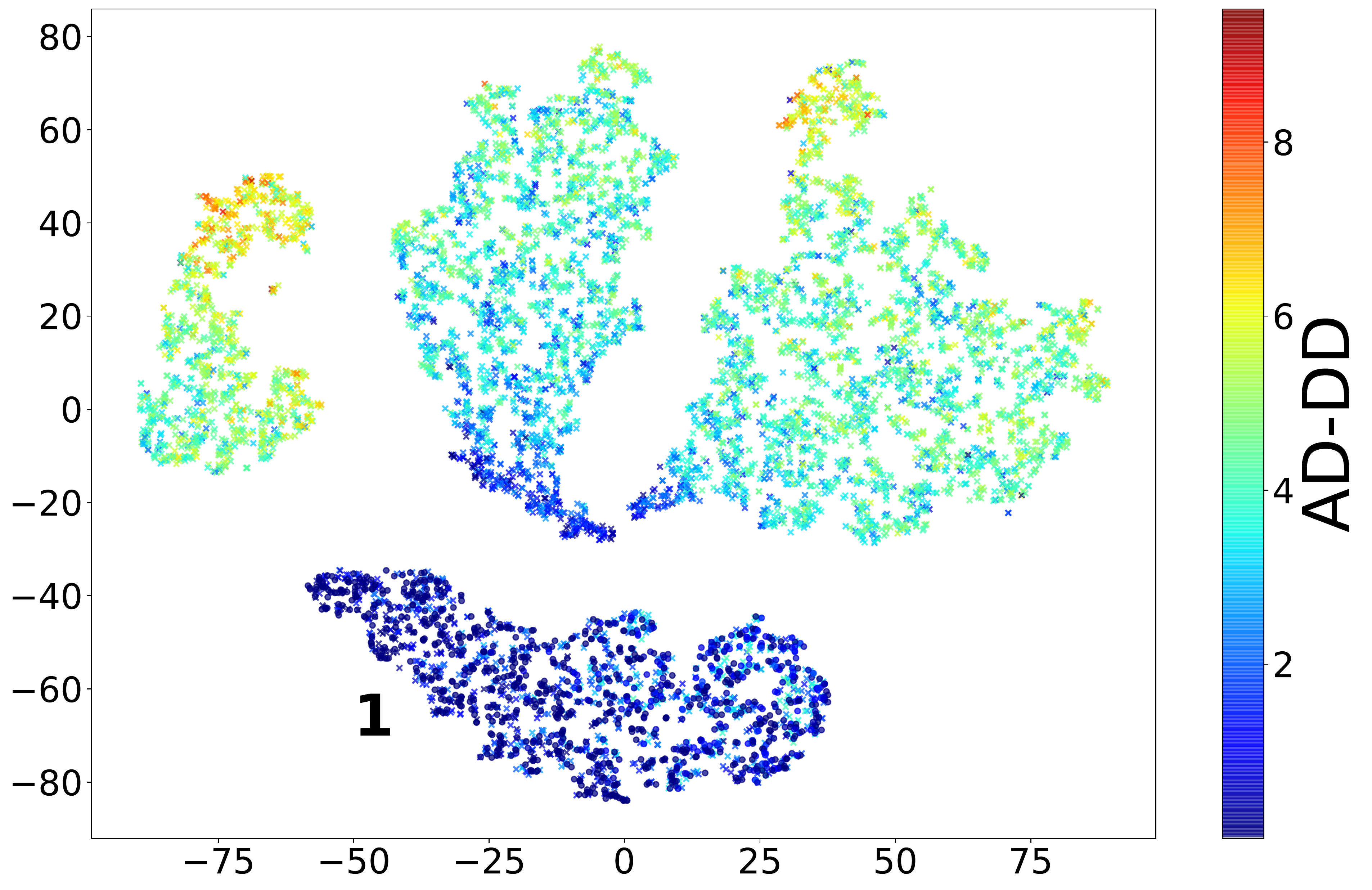}\\
\includegraphics[width = 0.48\columnwidth, trim=0.3cm 0.3cm 0.3cm 0.3cm, clip=false]{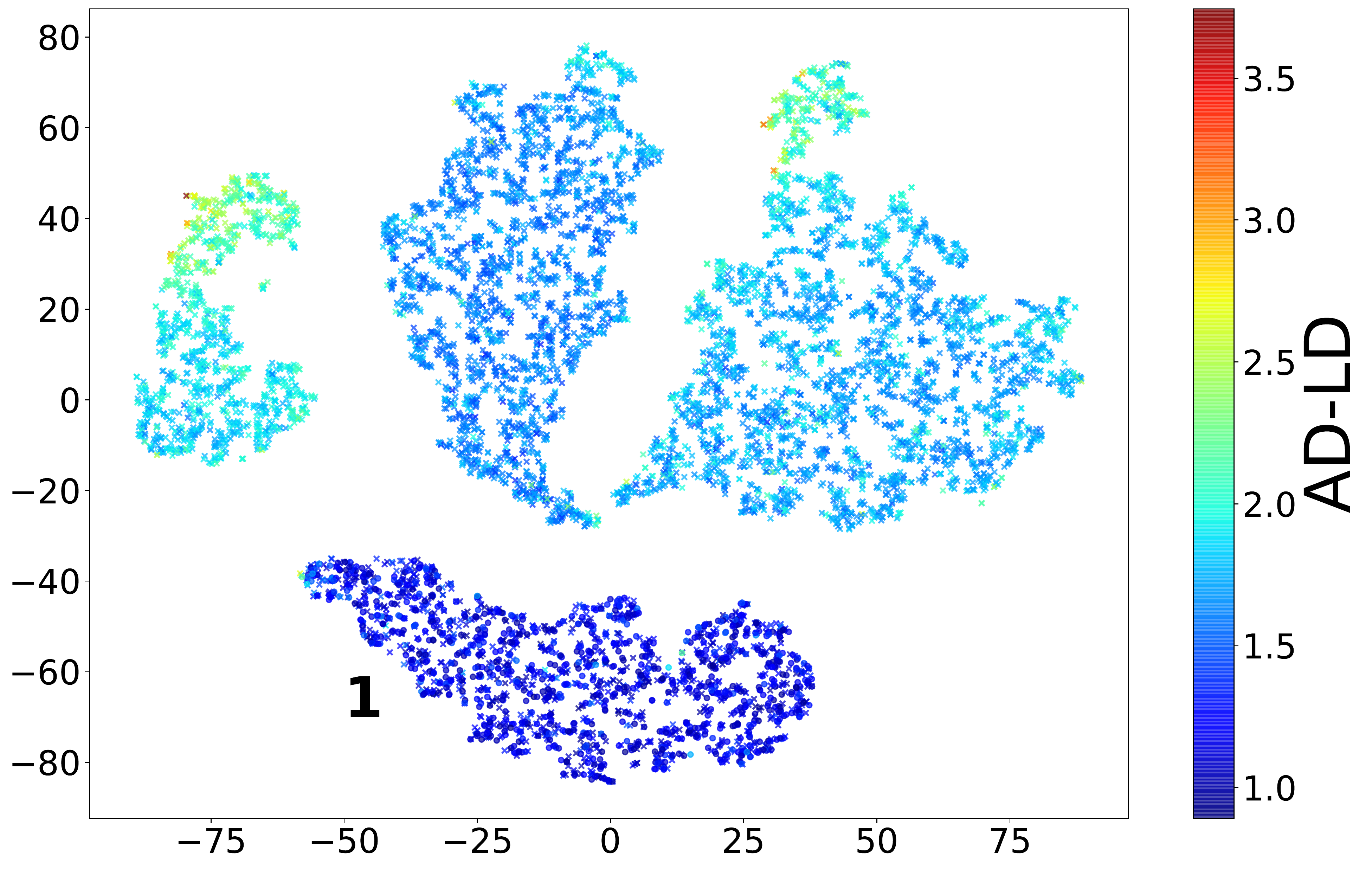}
&\includegraphics[width = 0.48\columnwidth, trim=0.3cm 0.3cm 0.3cm 0.3cm, clip=false]{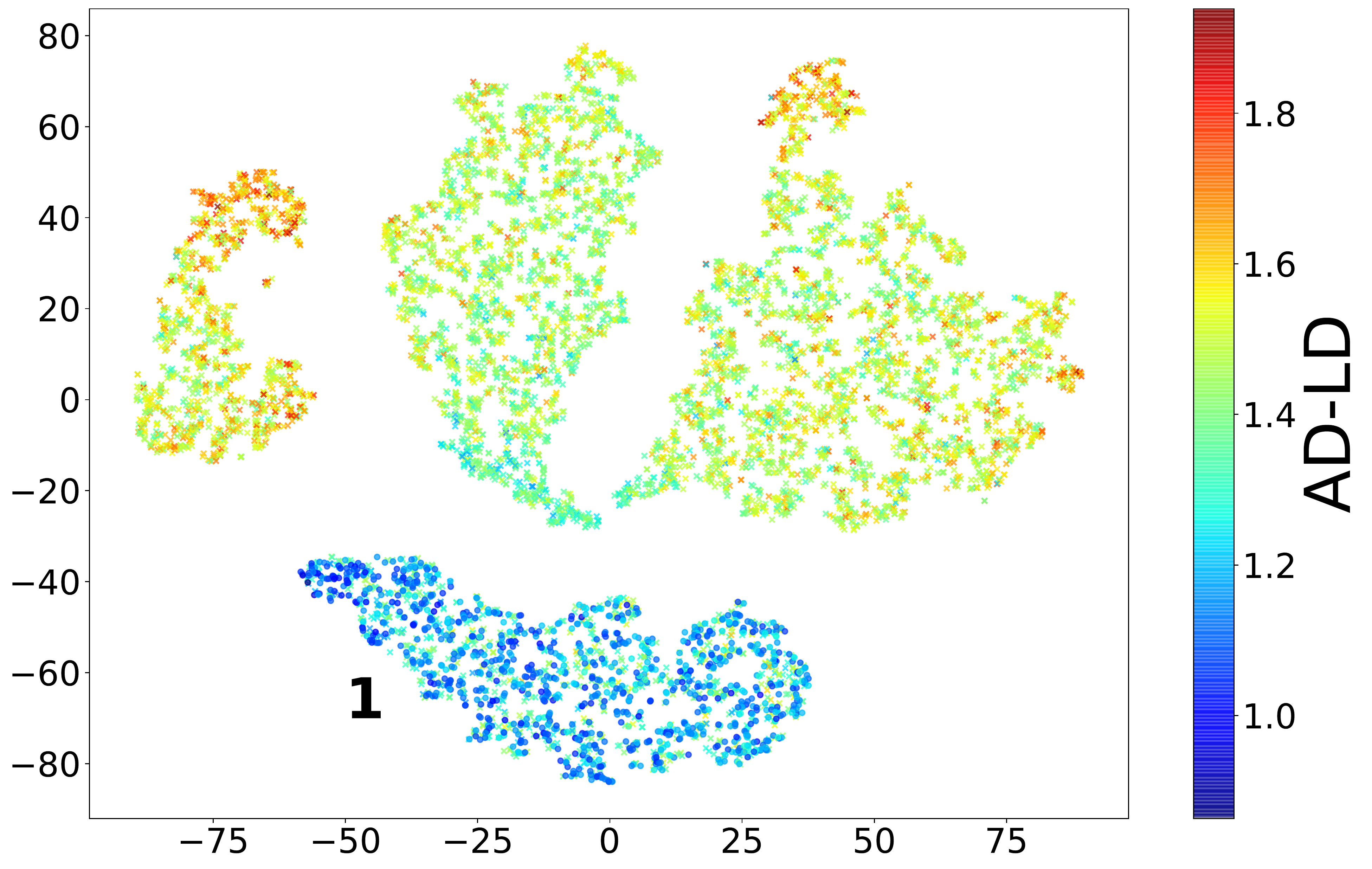}\\
\includegraphics[width = 0.48\columnwidth, trim=0.3cm 0.3cm 0.3cm 0.3cm, clip=false]{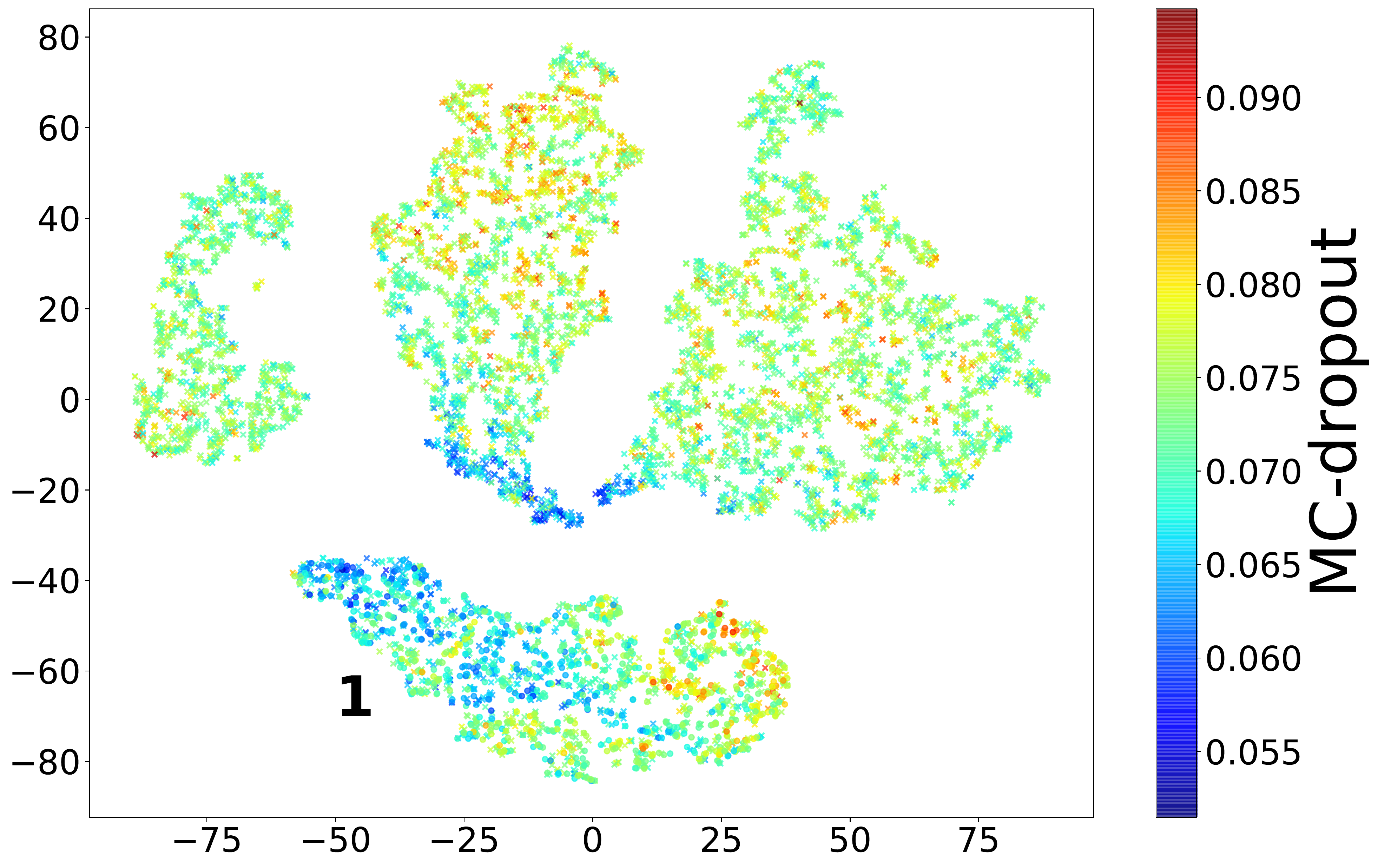}
&\includegraphics[width = 0.48\columnwidth, trim=0.3cm 0.3cm 0.3cm 0.3cm, clip=false]{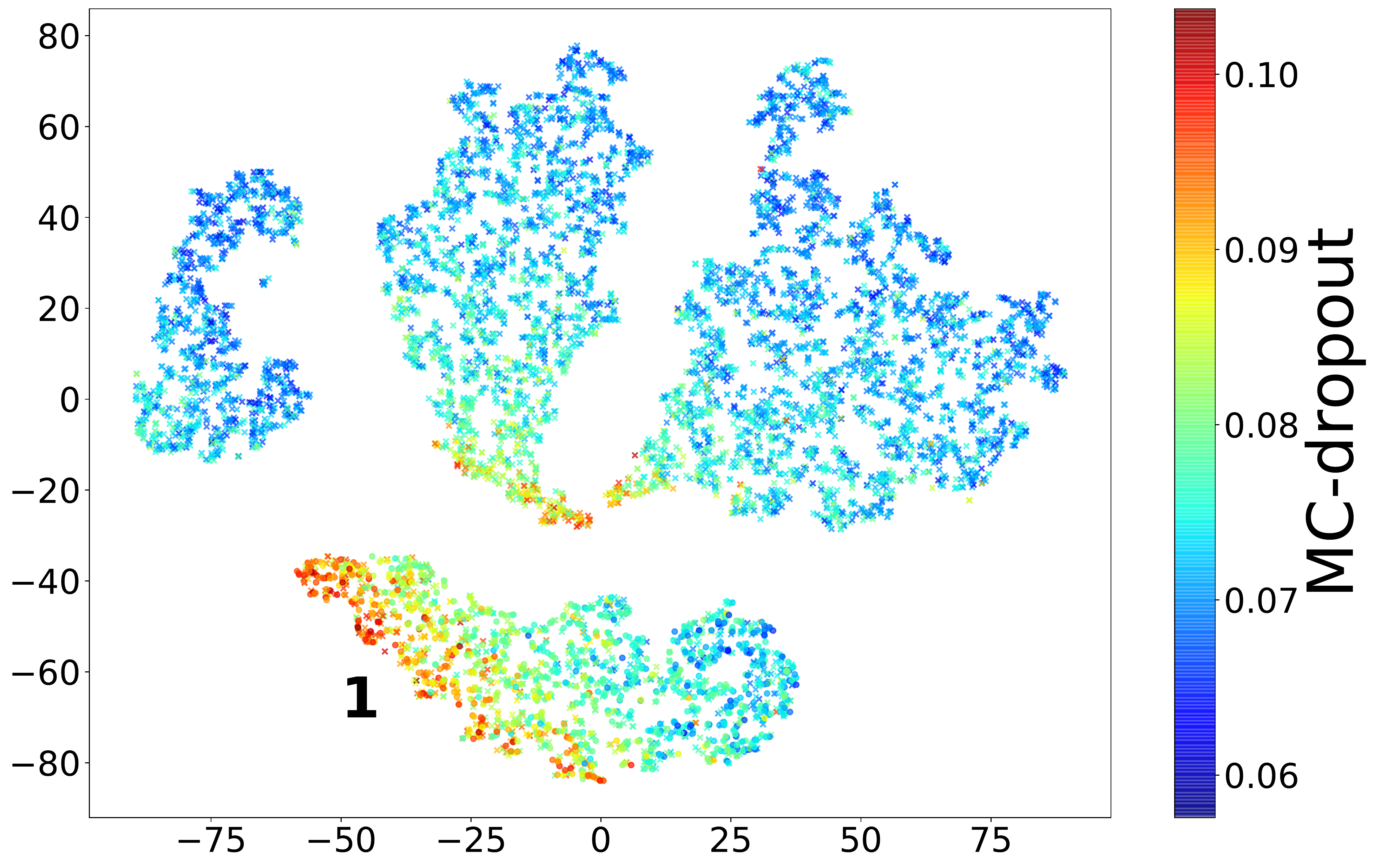}\\
\includegraphics[width = 0.48\columnwidth, trim=0.3cm 0.3cm 0.3cm 0.3cm, clip=false]{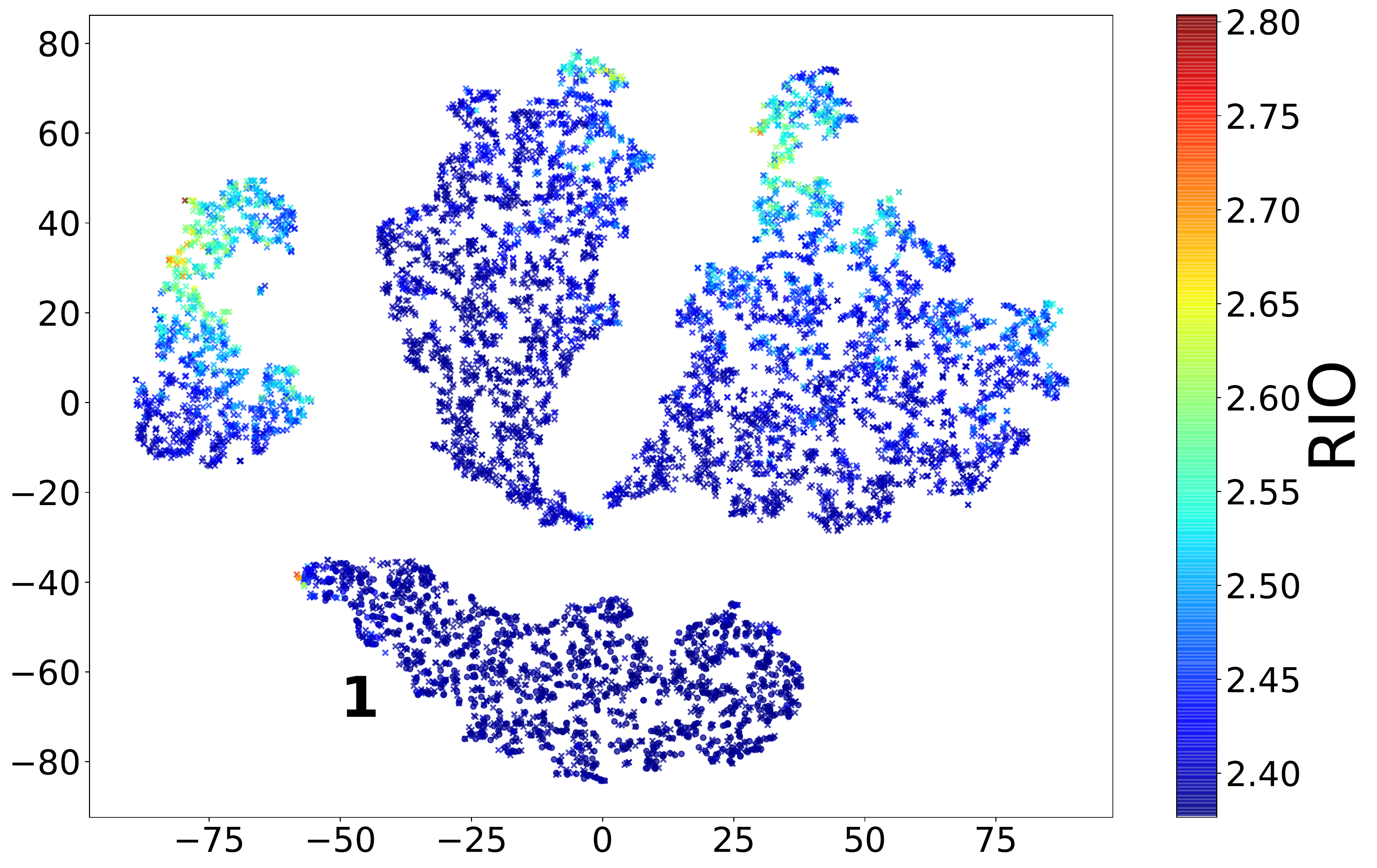}
&\includegraphics[width = 0.48\columnwidth, trim=0.3cm 0.3cm 0.3cm 0.3cm, clip=false]{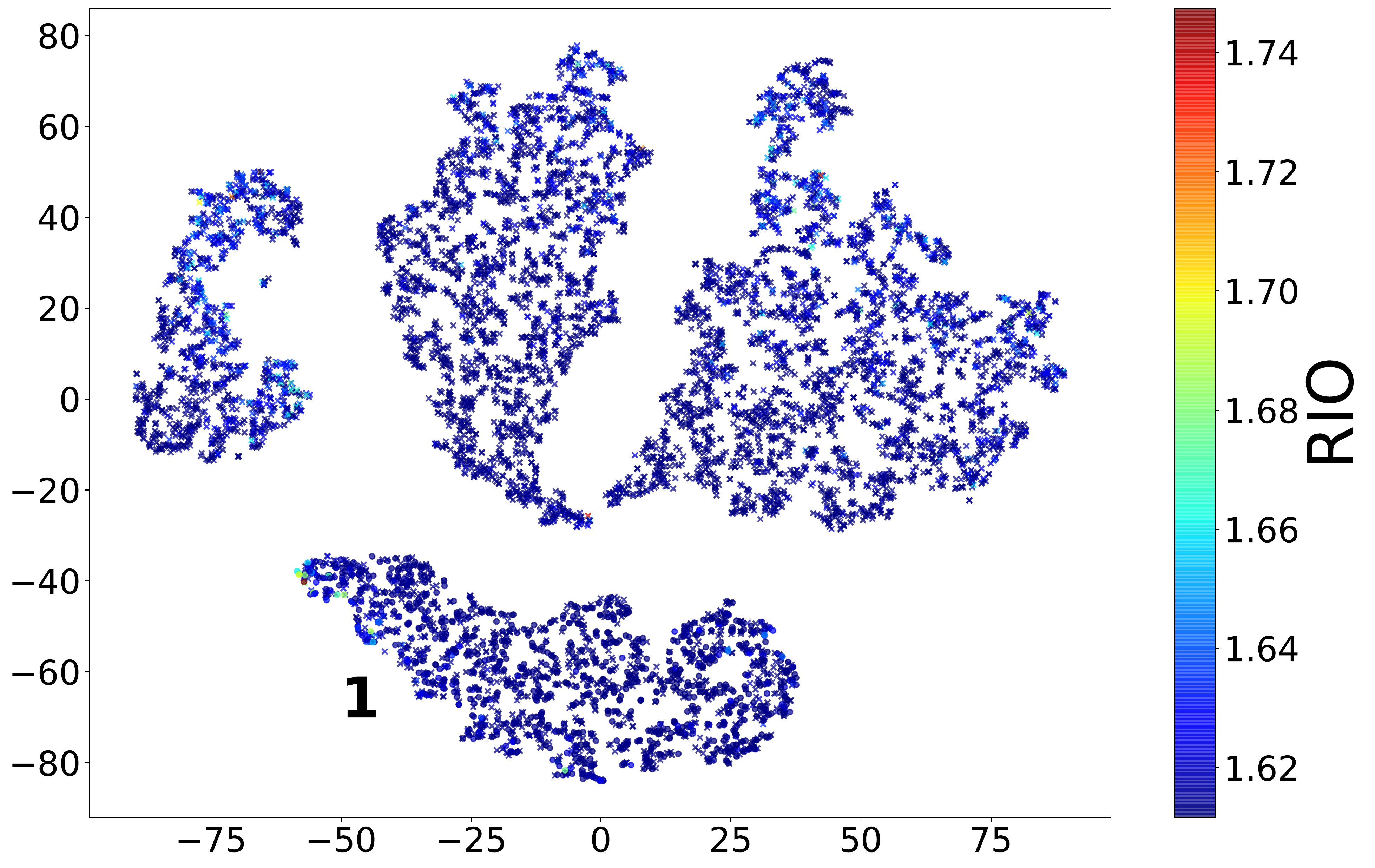}
\end{tabular}
\end{center}
\vspace{0in}
\caption{\label{fig:cluster1uq} Uncertainty values for the models trained on cluster 1. The bold number on the plot indicates the training set cluster. The x and y axes are the two coordinates from tSNE using top correlated MOE features. }
\end{figure}

\begin{figure}
\begin{center}
\small
\begin{tabular}{cc}
{\large (a) Model trained on cluster 1 with MOE} & {\large (b) Model trained on cluster 1 with ECFP}\\ 
\includegraphics[width = 0.46\columnwidth, trim=0.3cm 0.cm 0.3cm 0cm, clip=false]{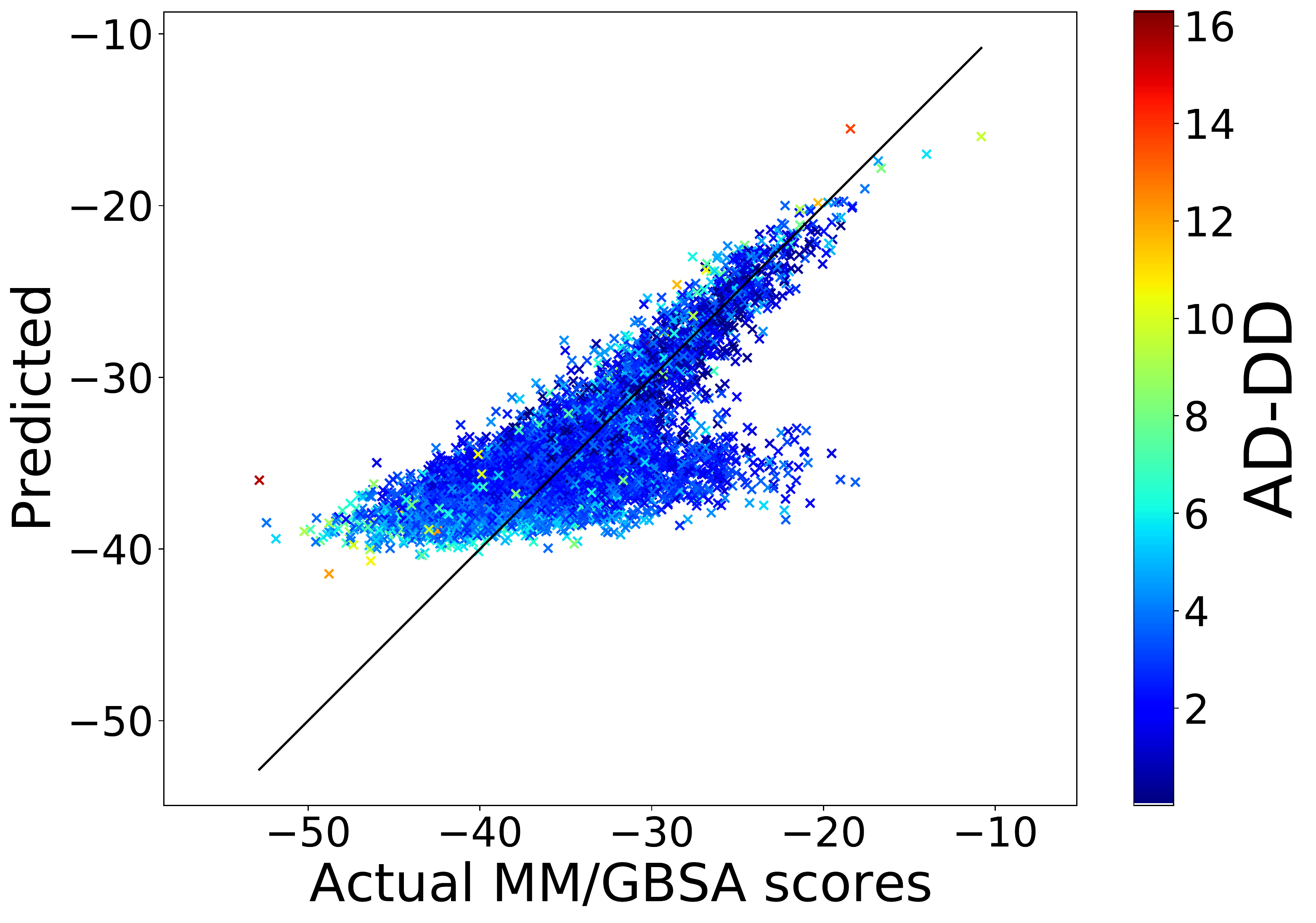}
&\includegraphics[width = 0.46\columnwidth, trim=0.3cm 0.cm 0.3cm 0cm, clip=false]{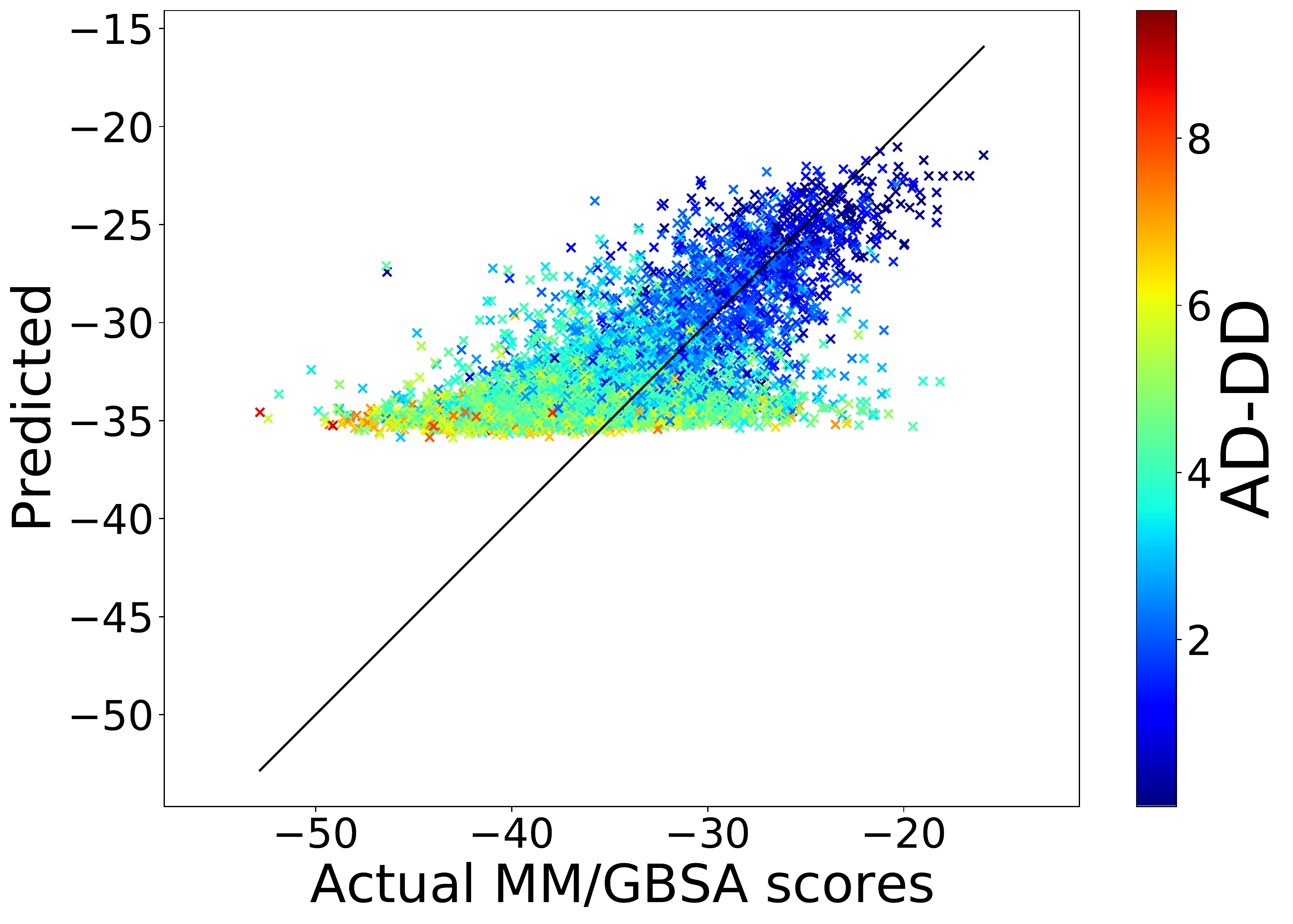}\\
\includegraphics[width = 0.46\columnwidth, trim=0.3cm 0.3cm 0.3cm 0.3cm, clip=false]{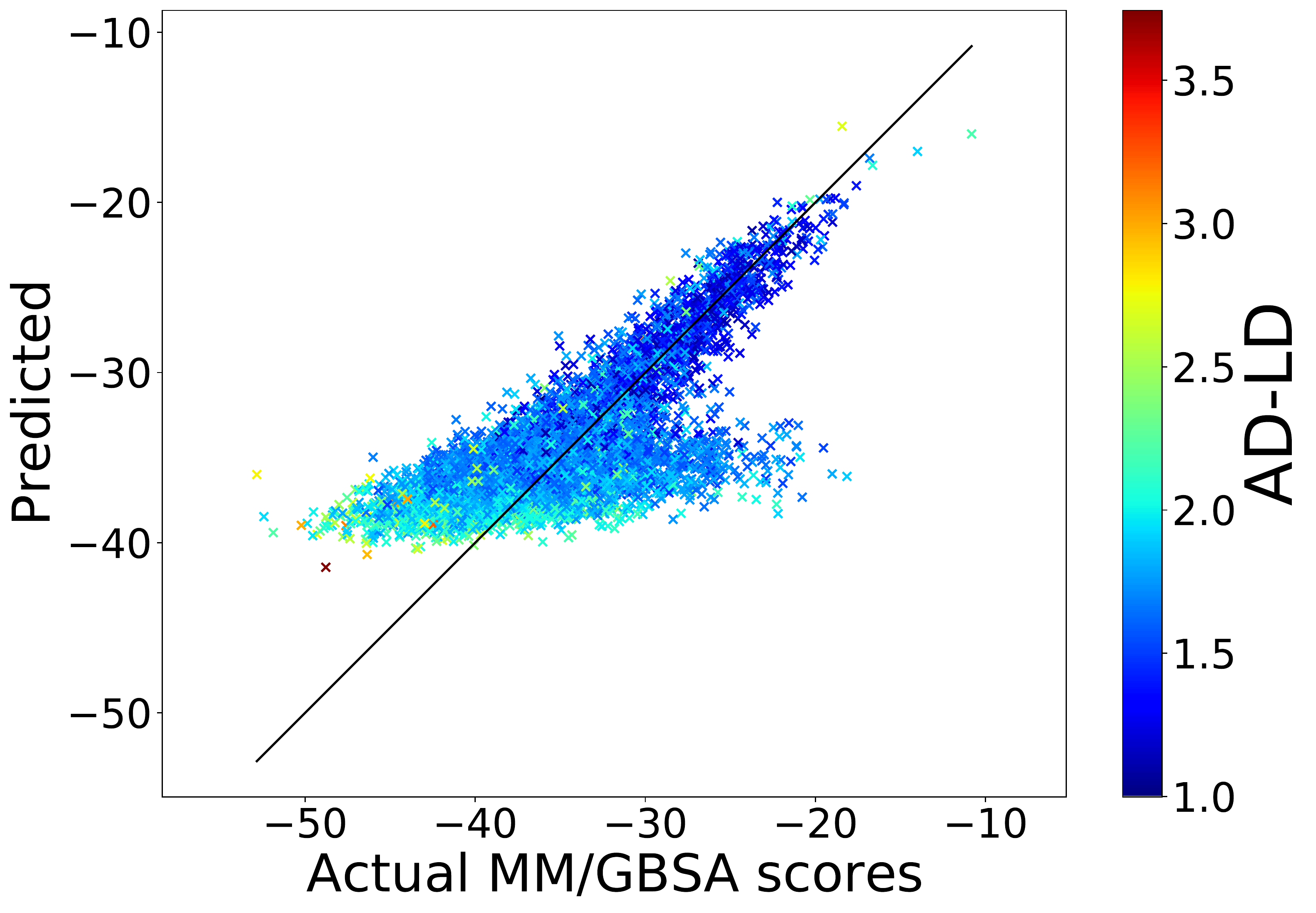}
&\includegraphics[width = 0.46\columnwidth, trim=0.3cm 0.3cm 0.3cm 0.3cm, clip=false]{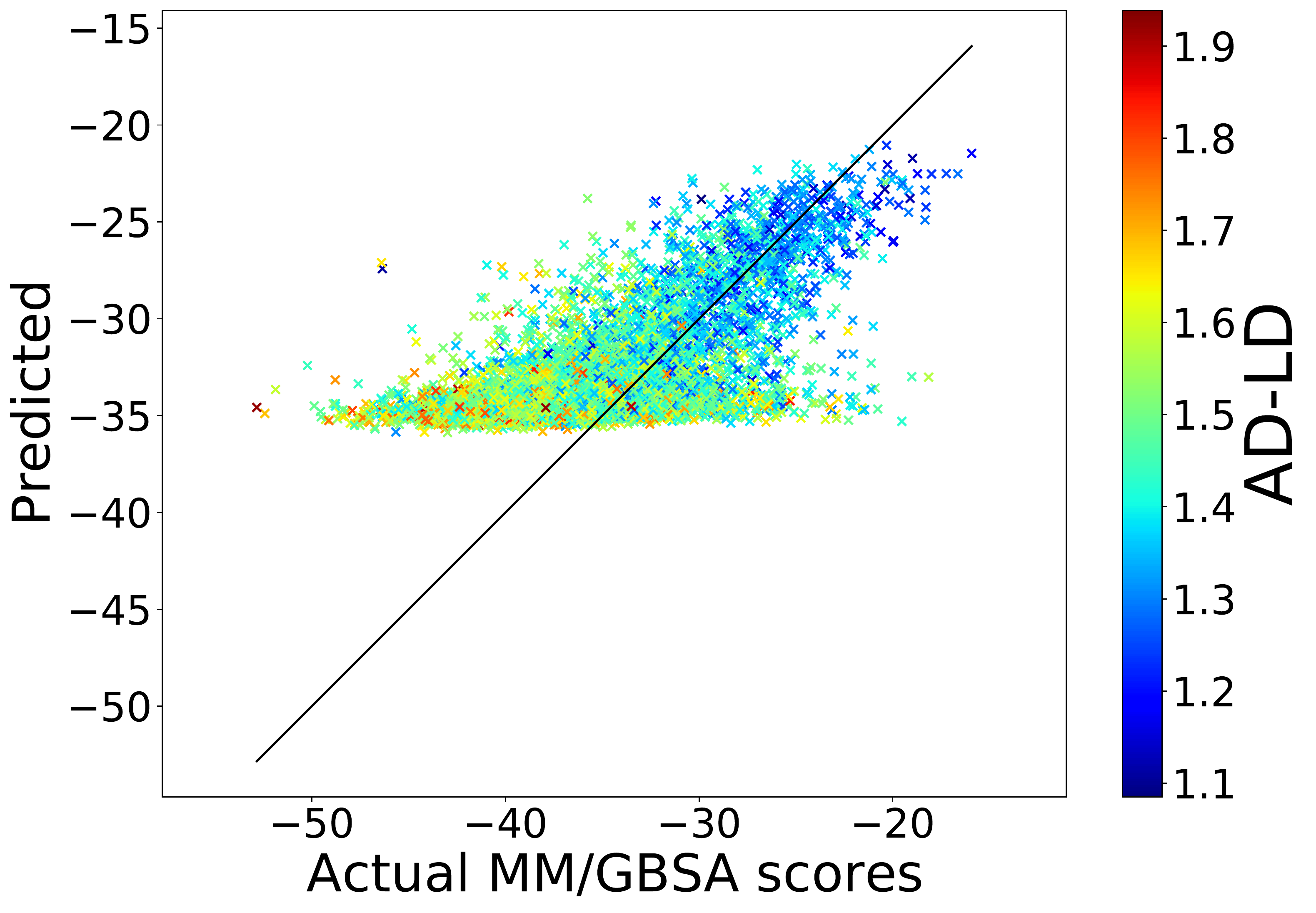}\\
\includegraphics[width = 0.46\columnwidth, trim=0.3cm 0.3cm 0.3cm 0.3cm, clip=false]{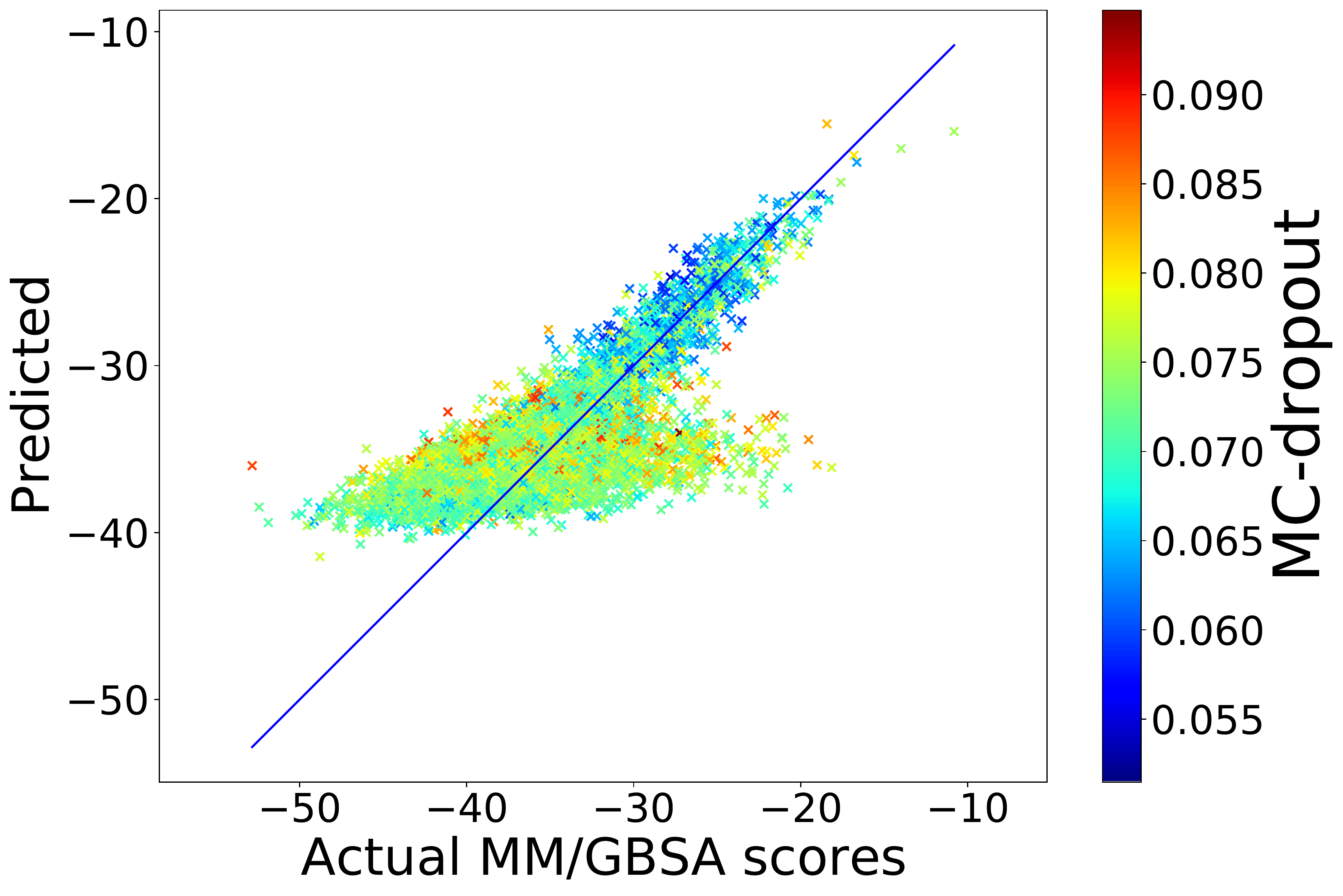}
&\includegraphics[width = 0.46\columnwidth, trim=0.3cm 0.3cm 0.3cm 0.3cm, clip=false]{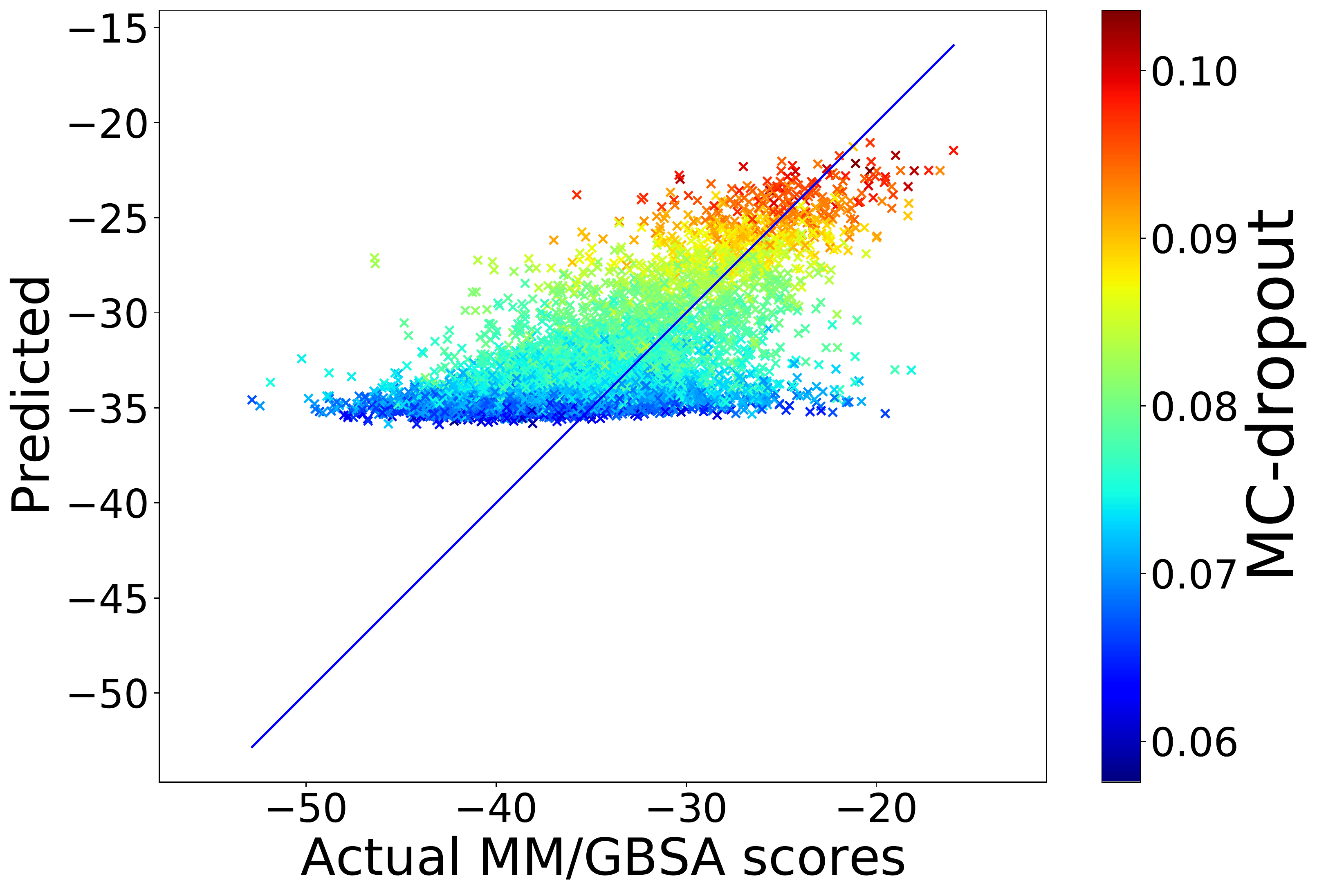}\\
\includegraphics[width = 0.46\columnwidth, trim=0.3cm 0.3cm 0.3cm 0.3cm, clip=false]{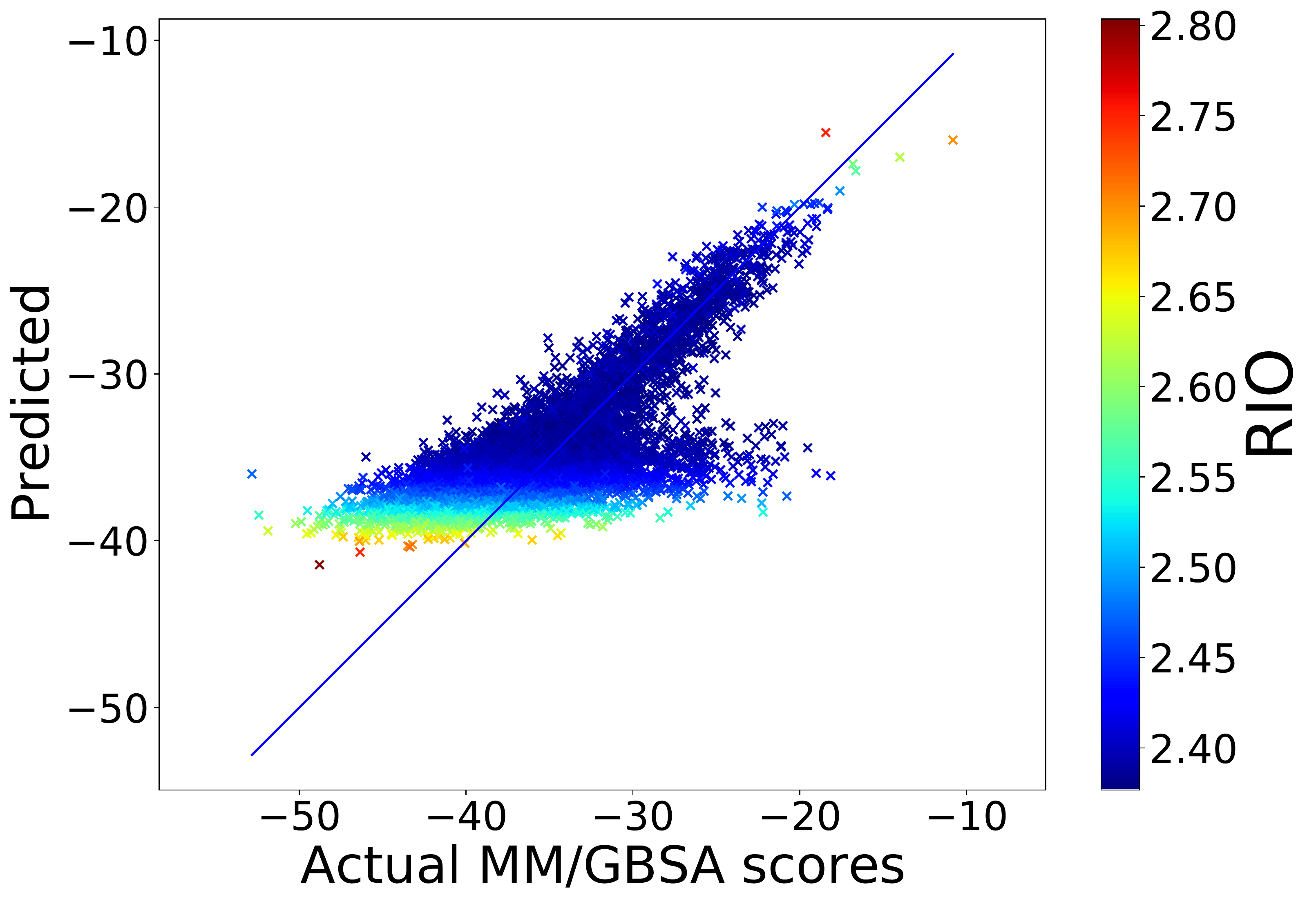}
&\includegraphics[width = 0.46\columnwidth, trim=0.3cm 0.3cm 0.3cm 0.3cm, clip=false]{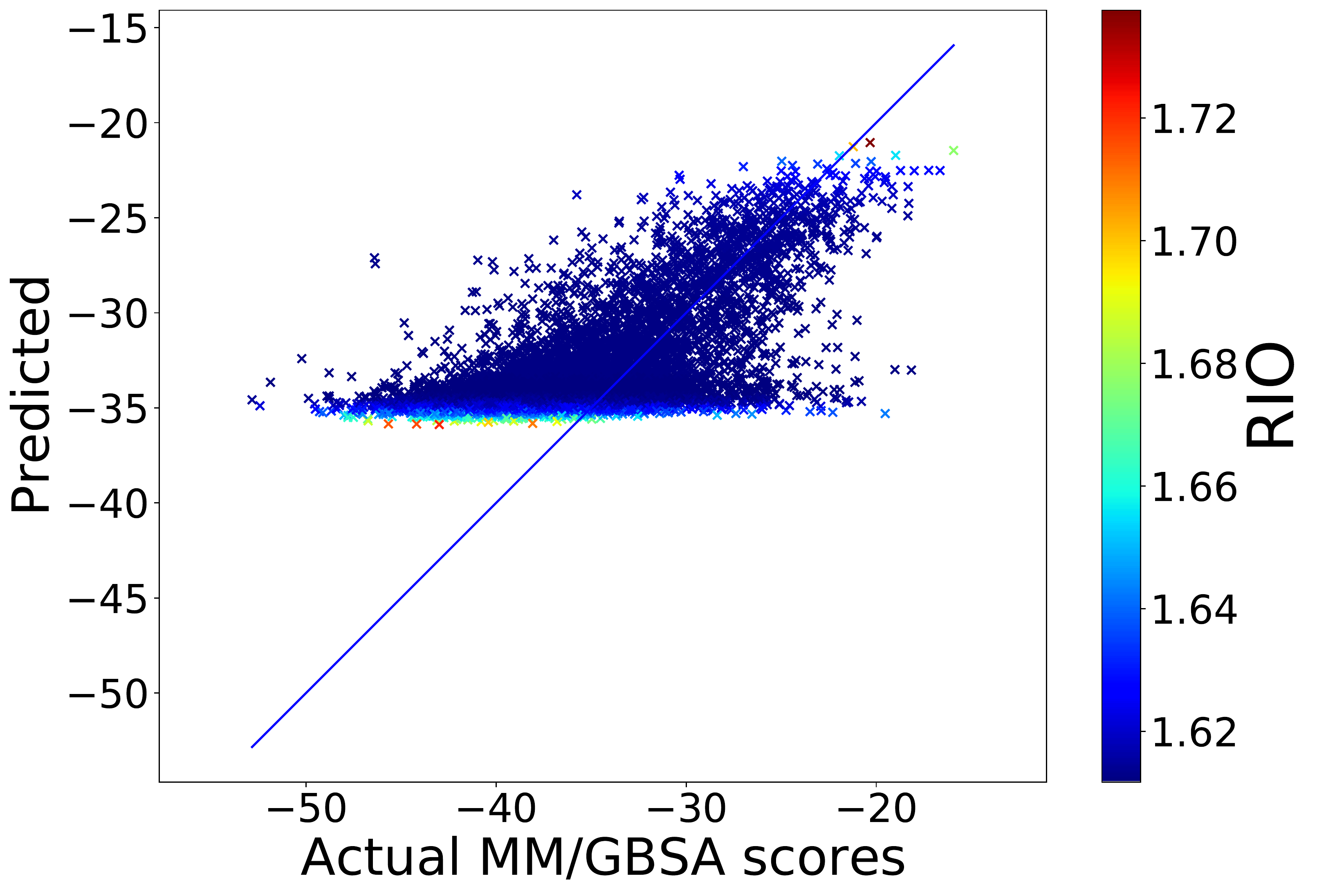}
\end{tabular}
\end{center}
\vspace{0in}
\caption{\label{fig:cluster1truePred} Test set actual MM/GBSA scores versus predicted plot from the model trained on cluster 1. The diagonal represents the values where the predicted is equal to the actual MM/GBSA score. }
\end{figure}
%

%------------------------------
% Cluster 2
%------------------------------

%
\begin{figure}[th]
\begin{center}
\begin{tabular}{cc}
{\large (a) Model trained on cluster 2 with MOE} & {\large (b) Model trained on cluster 2 with ECFP}\\ 
\includegraphics[width = 0.48\columnwidth, trim=0.3cm 0.3cm 0.3cm 0.3cm, clip=false]{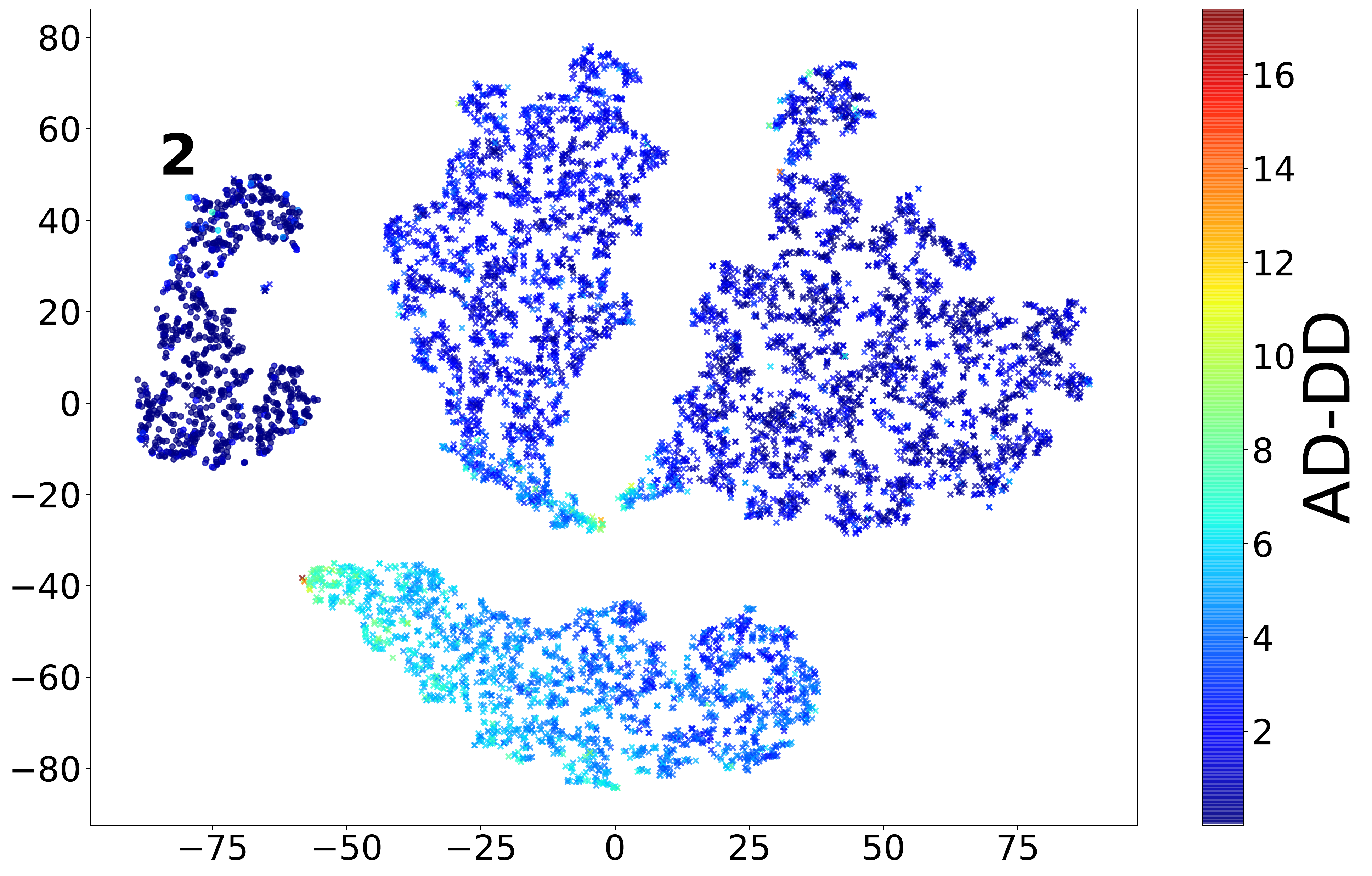}
&\includegraphics[width = 0.48\columnwidth, trim=0.3cm 0.3cm 0.3cm 0.3cm, clip=false]{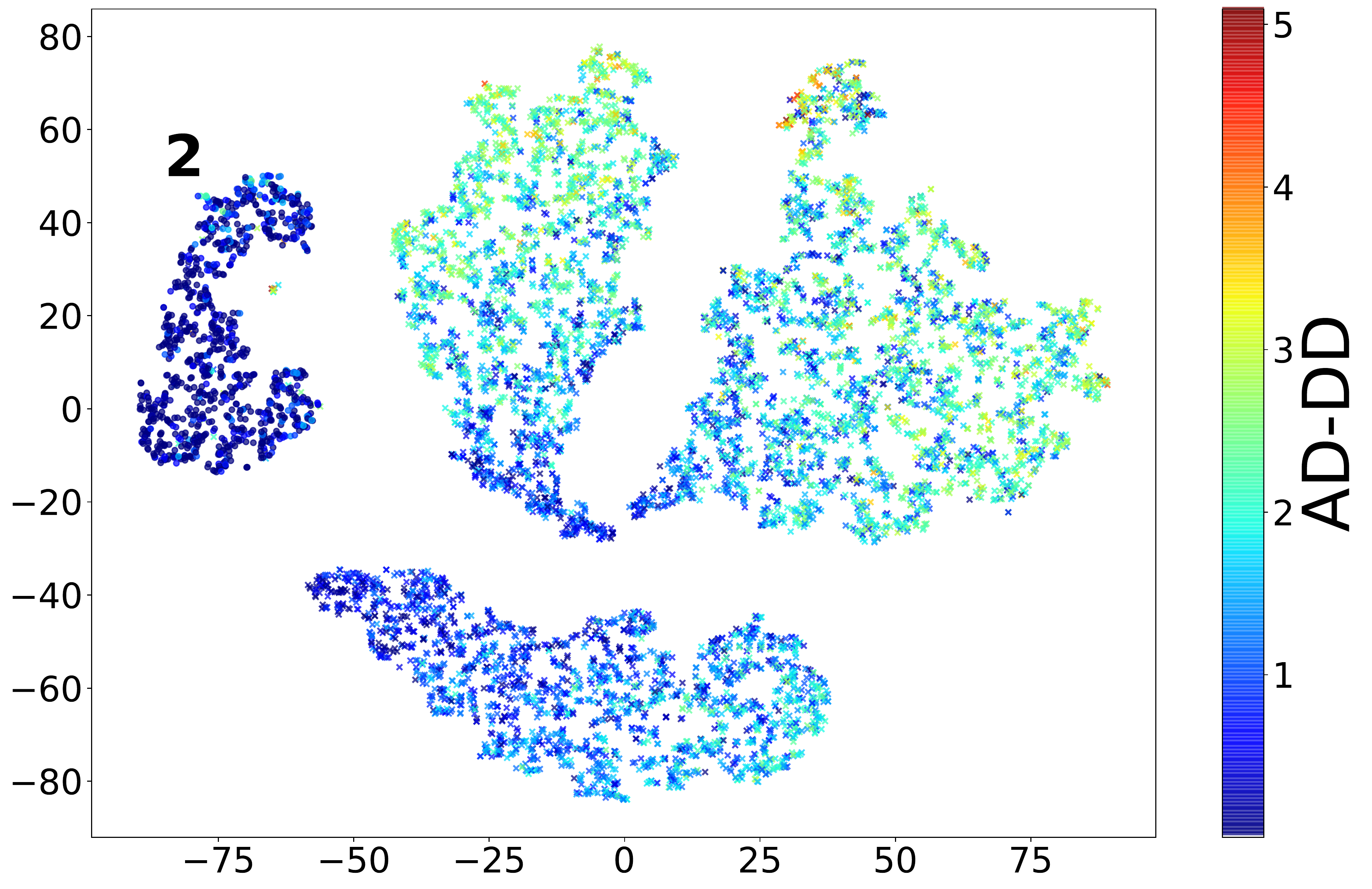}\\
\includegraphics[width = 0.48\columnwidth, trim=0.3cm 0.3cm 0.3cm 0.3cm, clip=false]{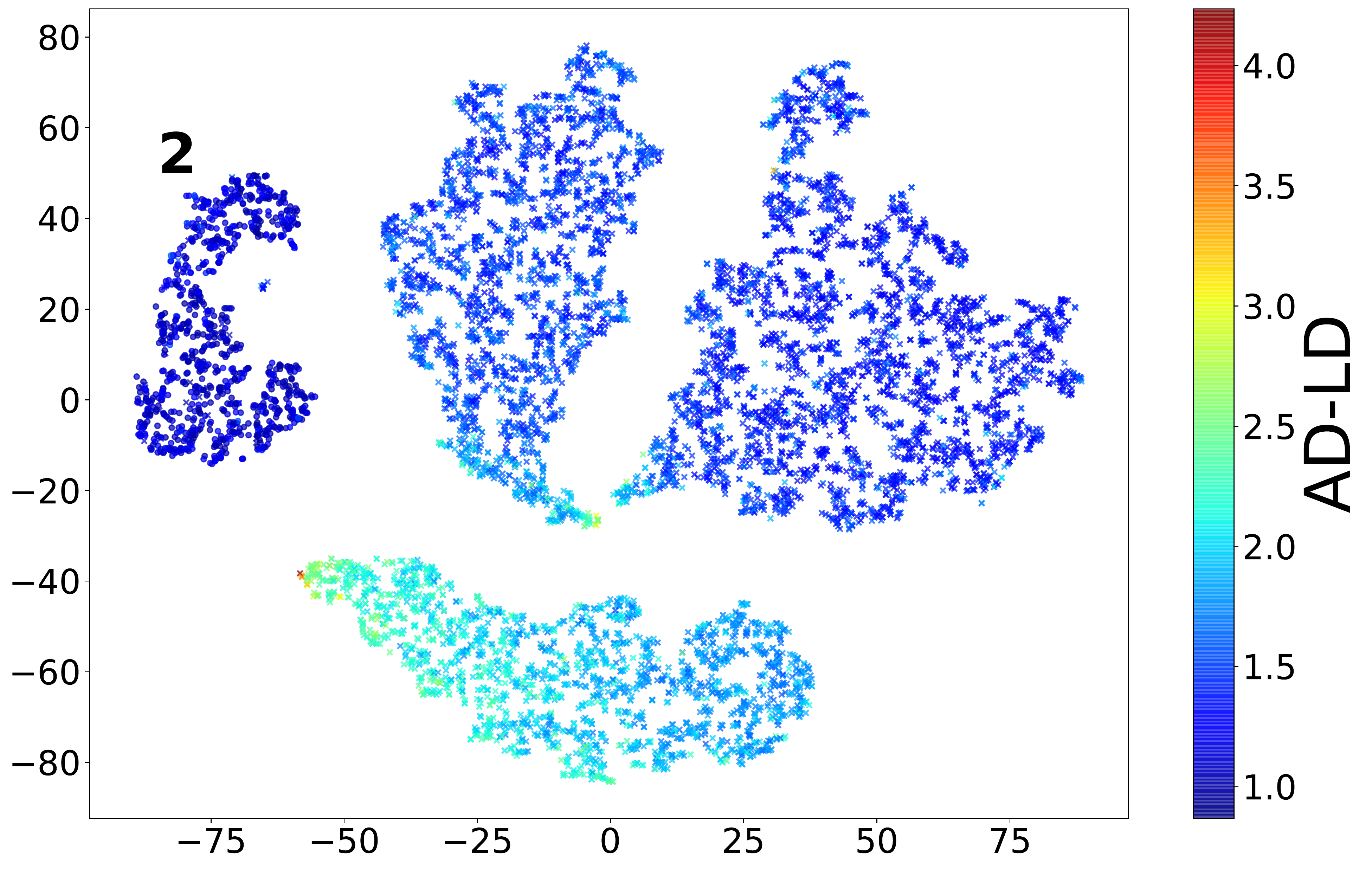}
&\includegraphics[width = 0.48\columnwidth, trim=0.3cm 0.3cm 0.3cm 0.3cm, clip=false]{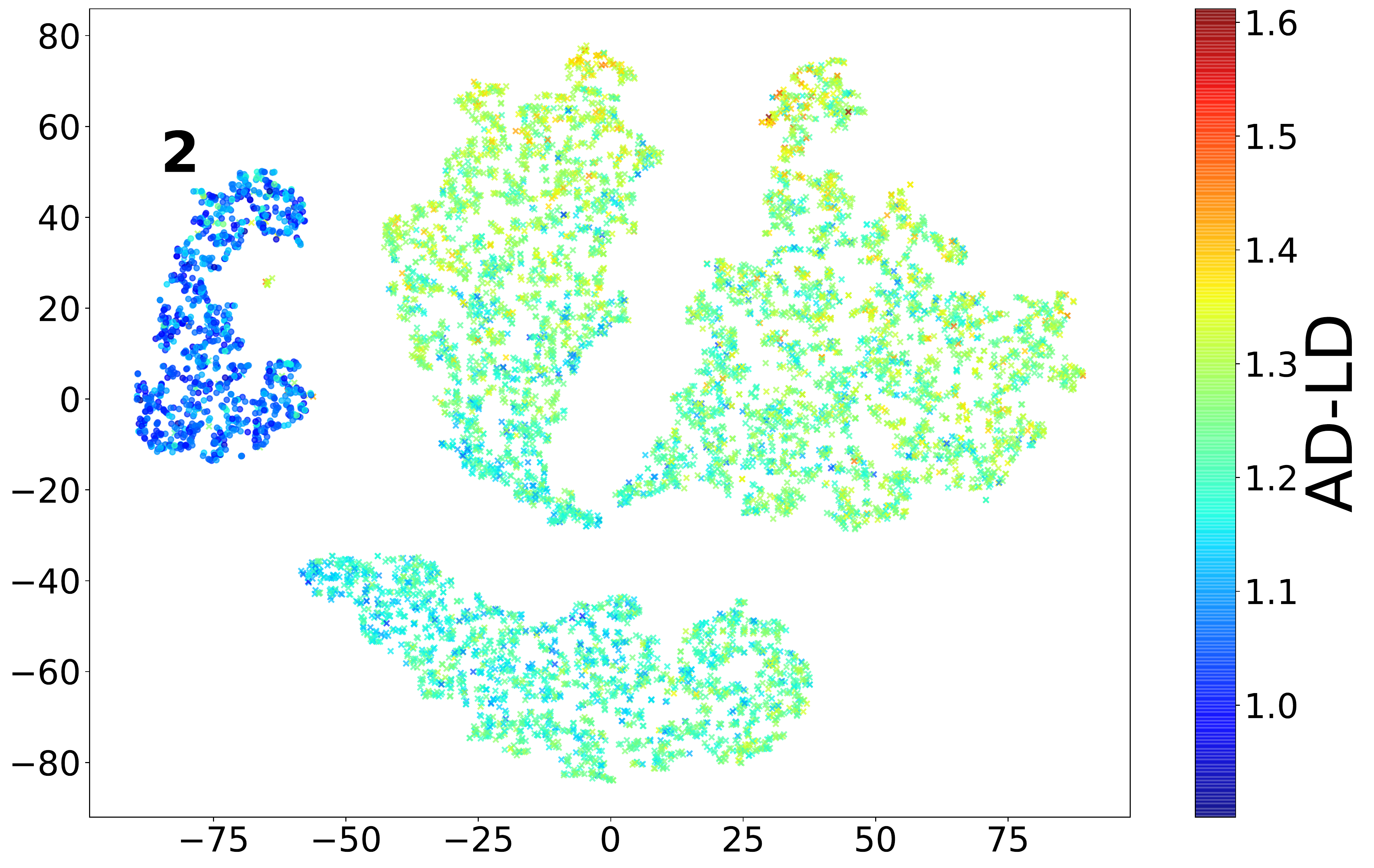}\\
\includegraphics[width = 0.48\columnwidth, trim=0.3cm 0.3cm 0.3cm 0.3cm, clip=false]{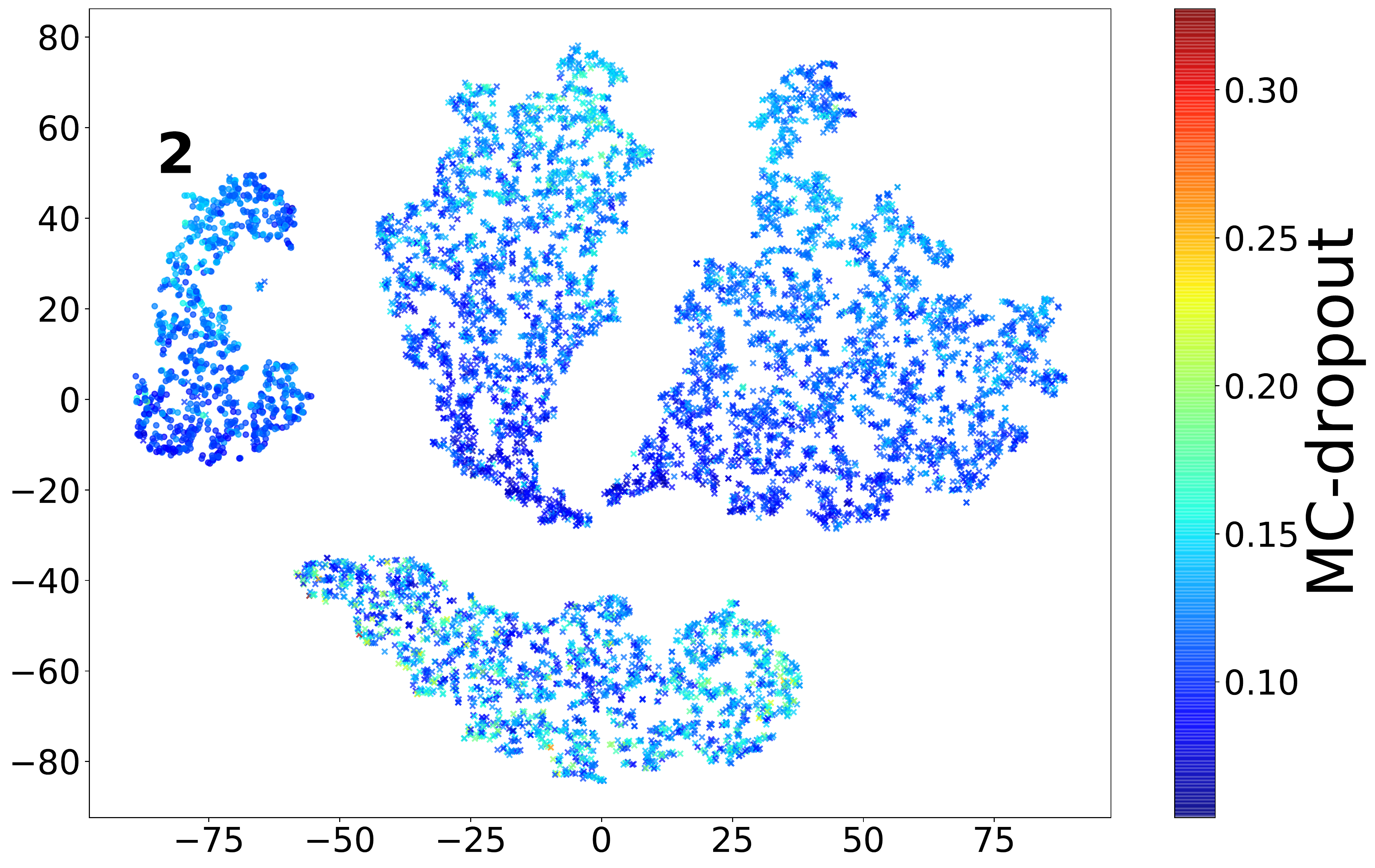}
&\includegraphics[width = 0.48\columnwidth, trim=0.3cm 0.3cm 0.3cm 0.3cm, clip=false]{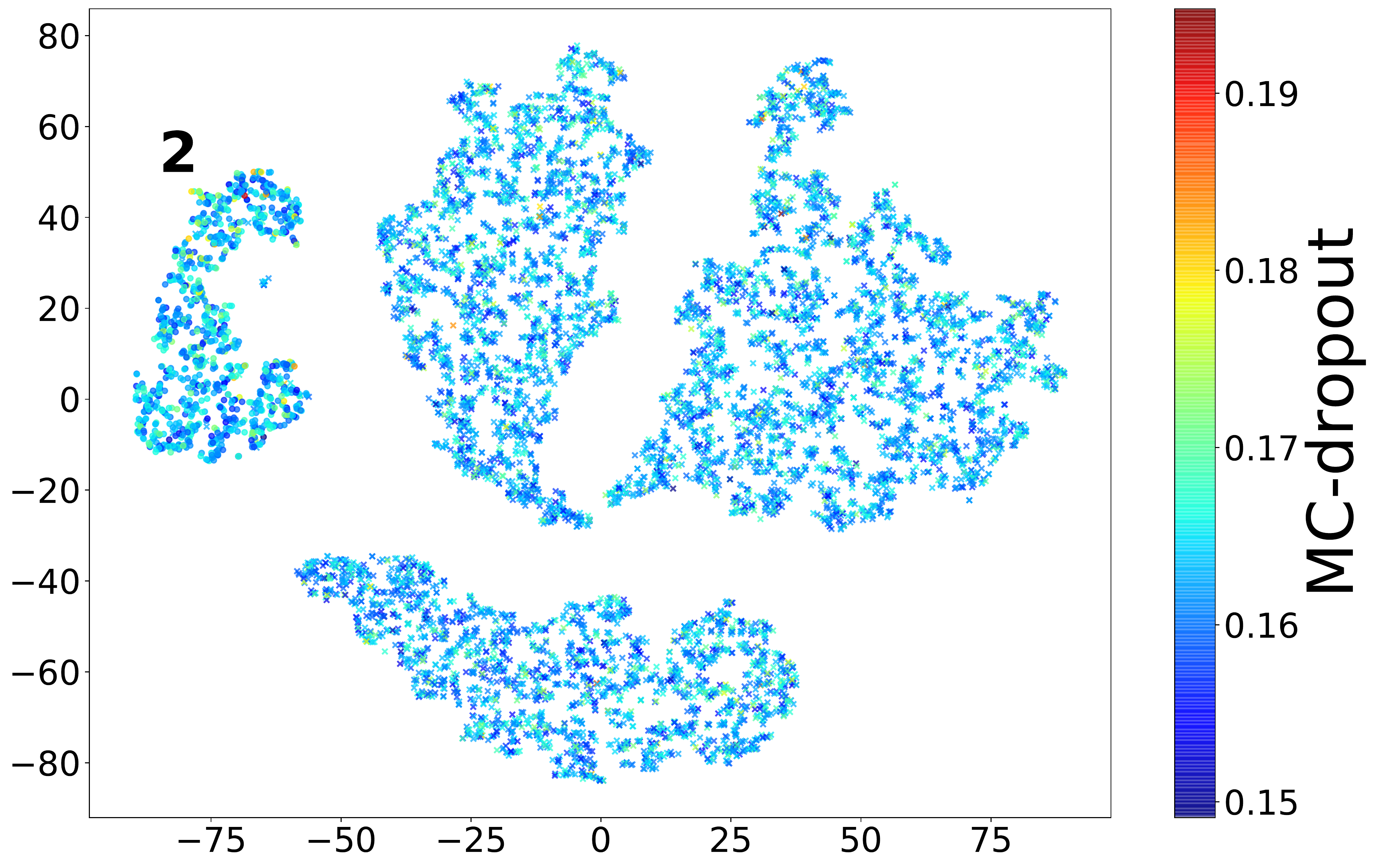}\\
\includegraphics[width = 0.48\columnwidth, trim=0.3cm 0.3cm 0.3cm 0.3cm, clip=false]{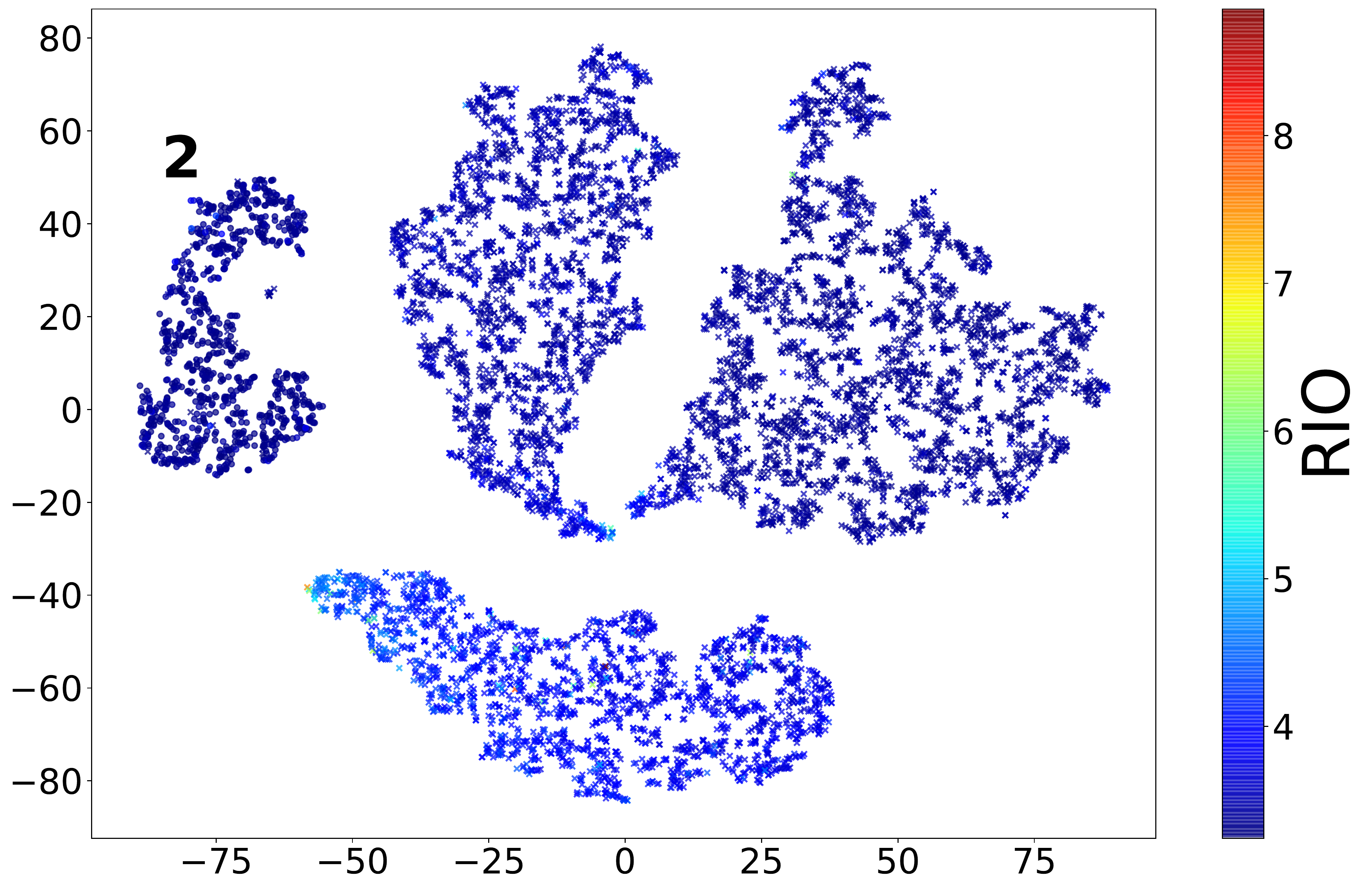}
&\includegraphics[width = 0.48\columnwidth, trim=0.3cm 0.3cm 0.3cm 0.3cm, clip=false]{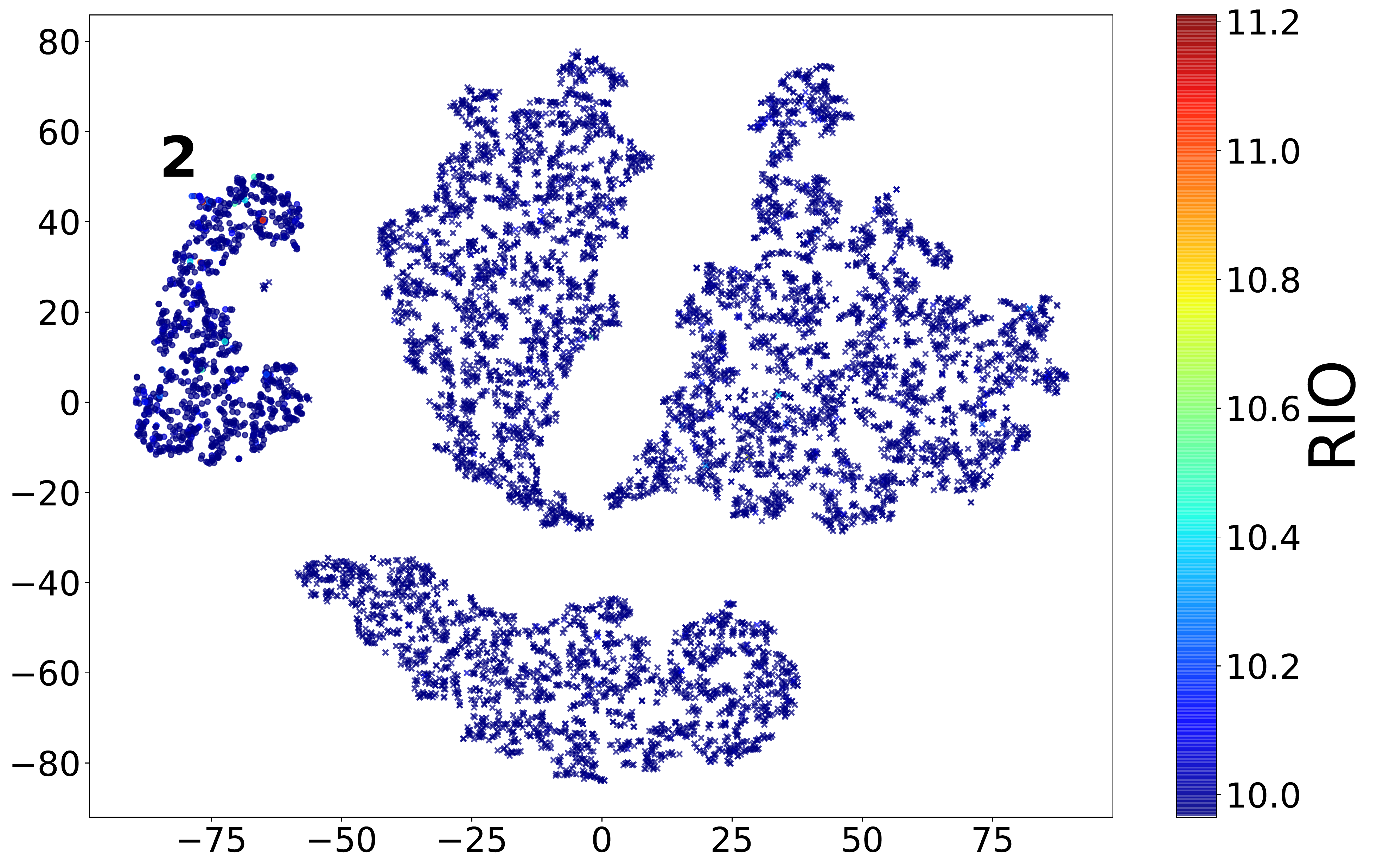}
\end{tabular}
\end{center}
\vspace{0in}
\caption{\label{fig:cluster2uq} Uncertainty values for the model trained on cluster 2.}
\end{figure}

\begin{figure}
\begin{center}
\begin{tabular}{cc}
{\large (a) Model trained on cluster 2 with MOE} & {\large (b) Model trained on cluster 2 with ECFP}\\ 
\includegraphics[width = 0.46\columnwidth, trim=0.3cm 0.3cm 0.3cm 0.3cm, clip=false]{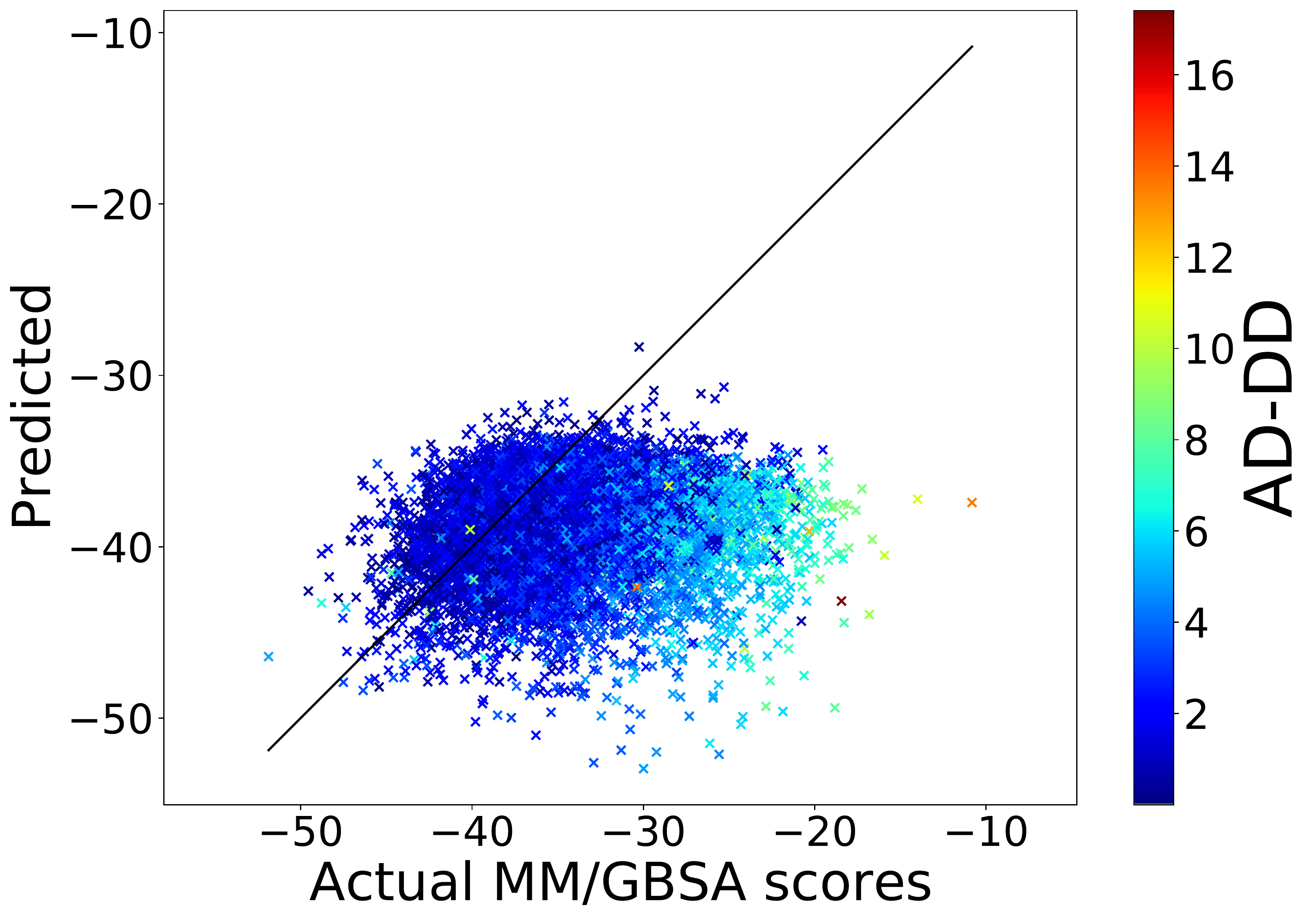}
&\includegraphics[width = 0.46\columnwidth, trim=0.3cm 0.3cm 0.3cm 0.3cm, clip=false]{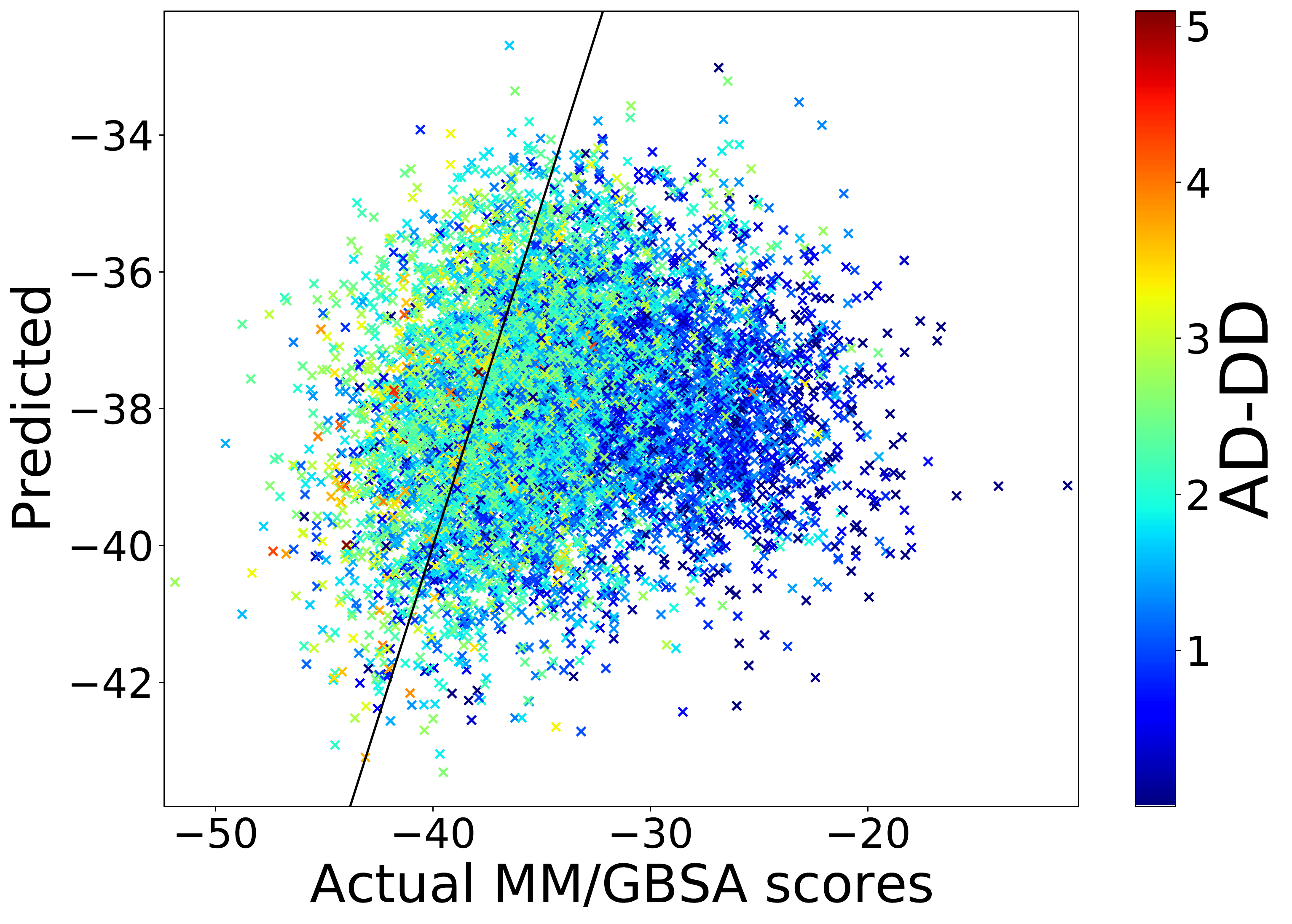}\\
\includegraphics[width = 0.46\columnwidth, trim=0.3cm 0.3cm 0.3cm 0.3cm, clip=false]{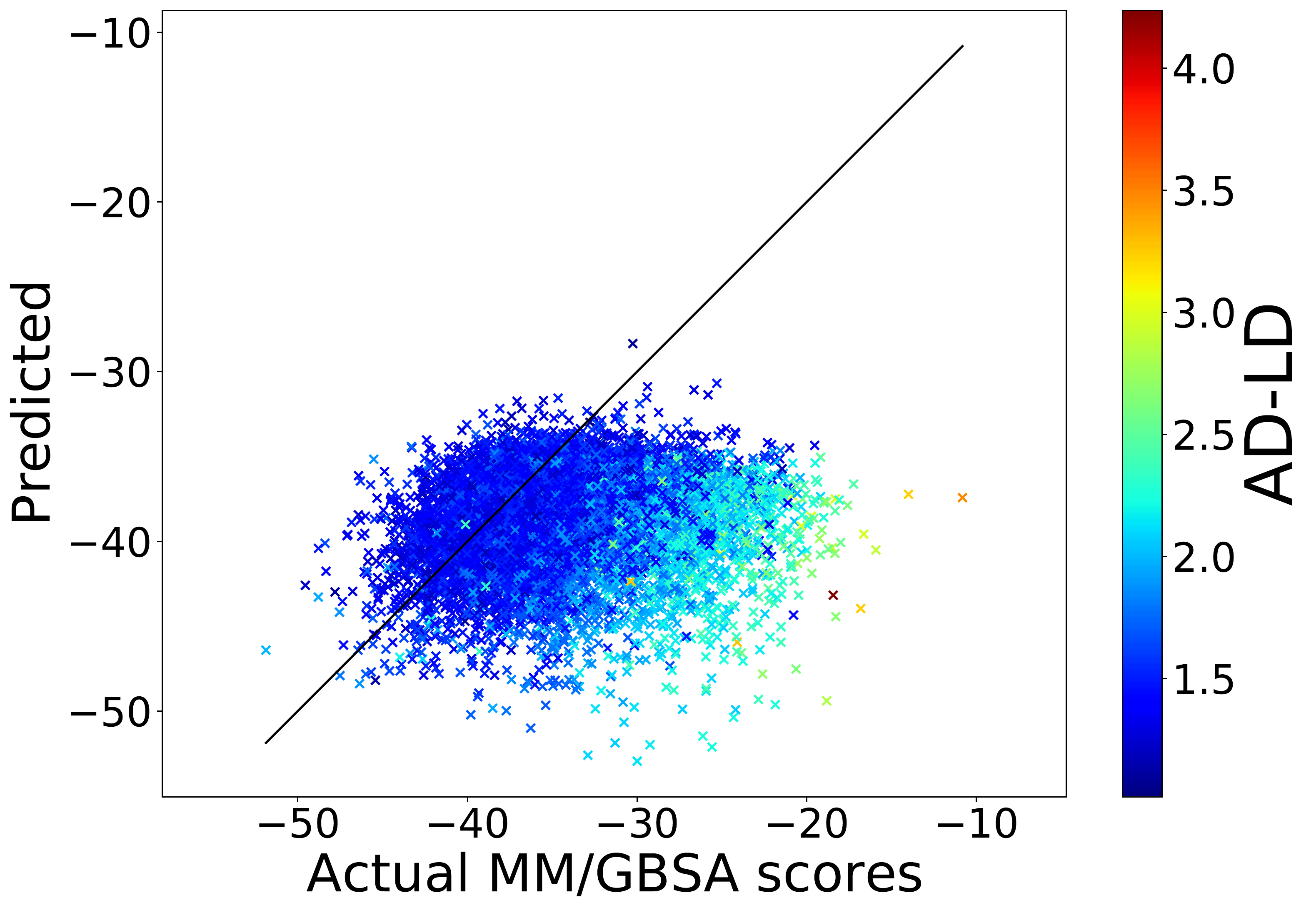}
&\includegraphics[width = 0.46\columnwidth, trim=0.3cm 0.3cm 0.3cm 0.3cm, clip=false]{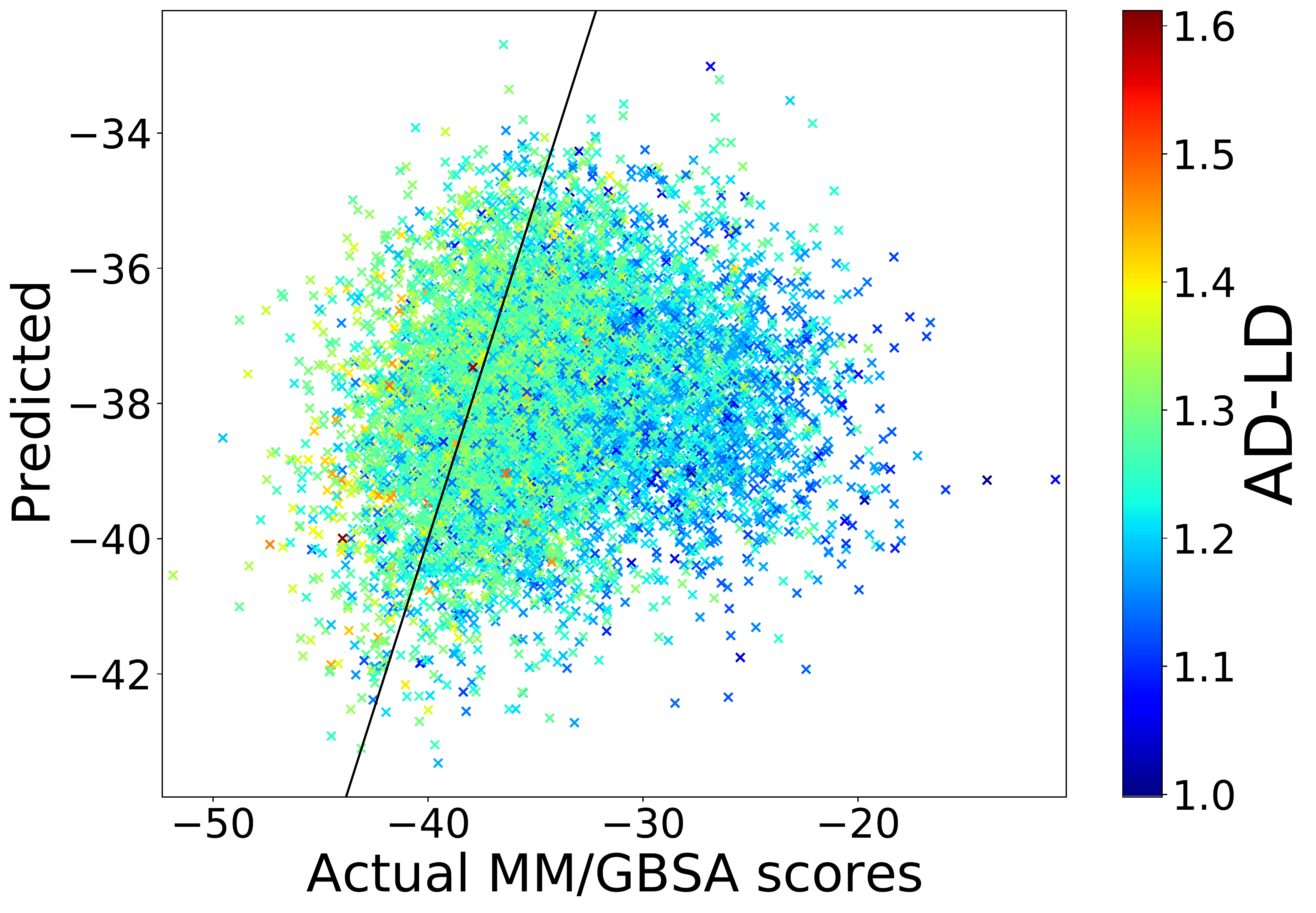}\\
\includegraphics[width = 0.46\columnwidth, trim=0.3cm 0.3cm 0.3cm 0.3cm, clip=false]{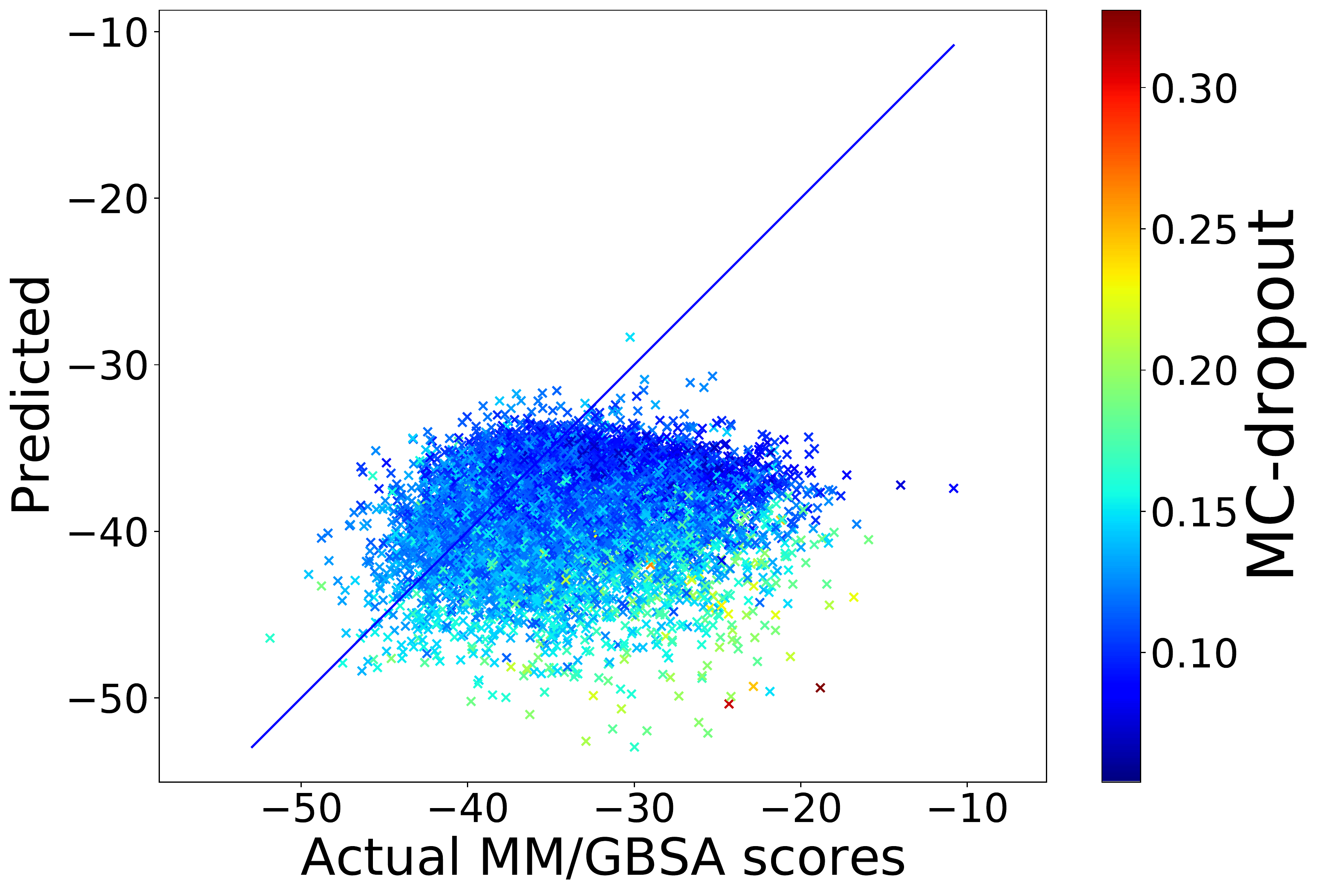}
&\includegraphics[width = 0.46\columnwidth, trim=0.3cm 0.3cm 0.3cm 0.3cm, clip=false]{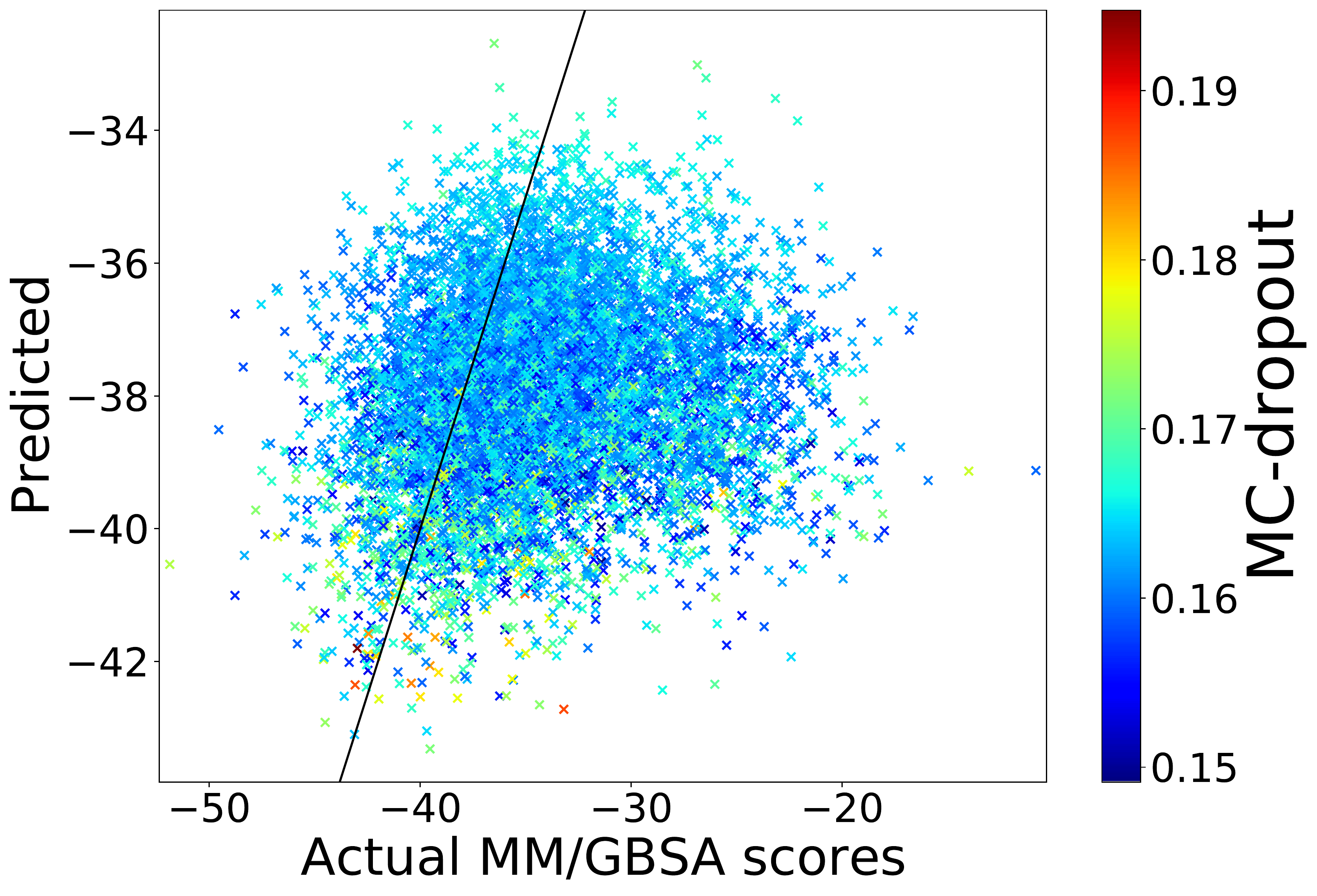}\\
\includegraphics[width = 0.46\columnwidth, trim=0.3cm 0.3cm 0.3cm 0.3cm, clip=false]{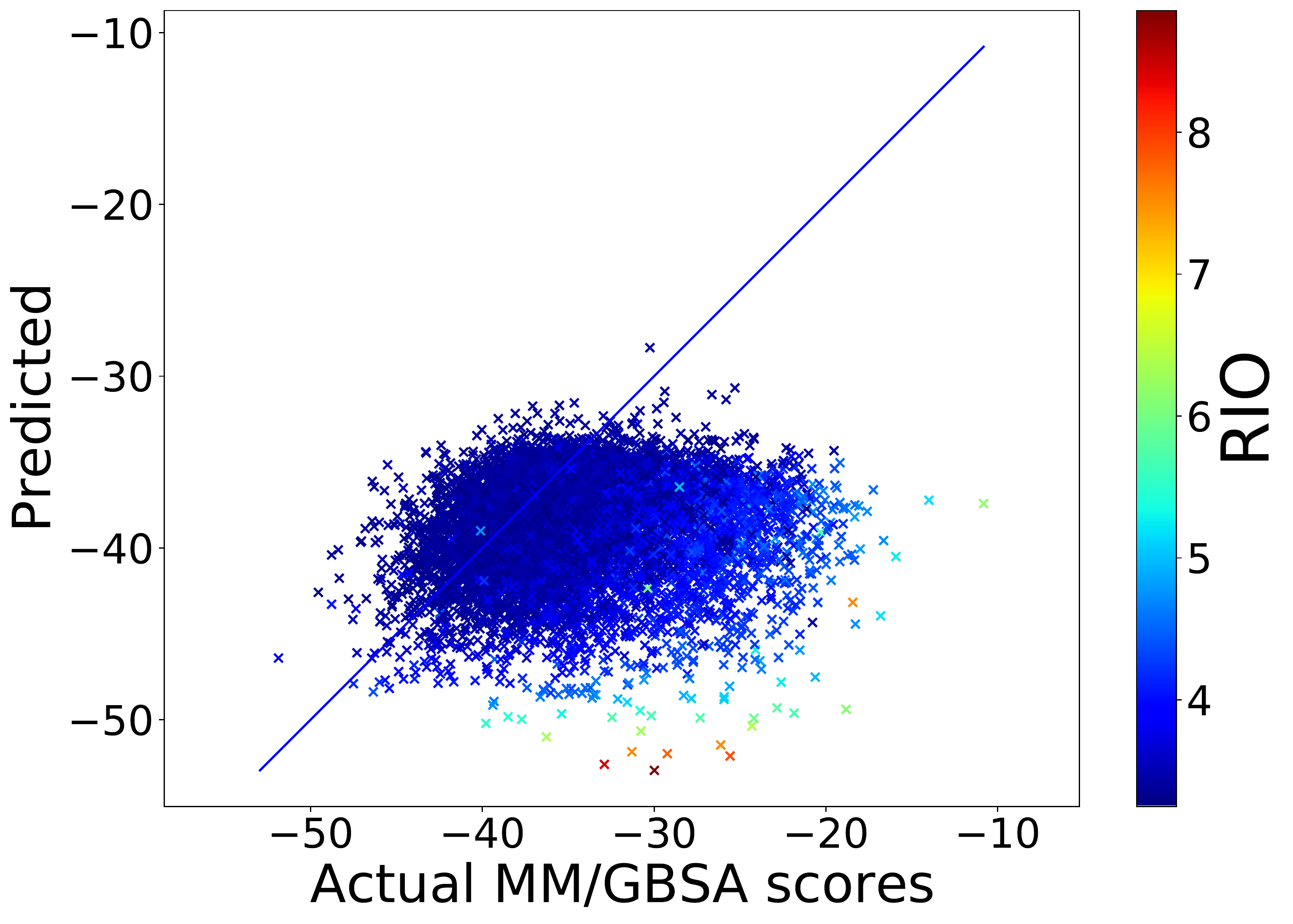}
&\includegraphics[width = 0.46\columnwidth, trim=0.3cm 0.3cm 0.3cm 0.3cm, clip=false]{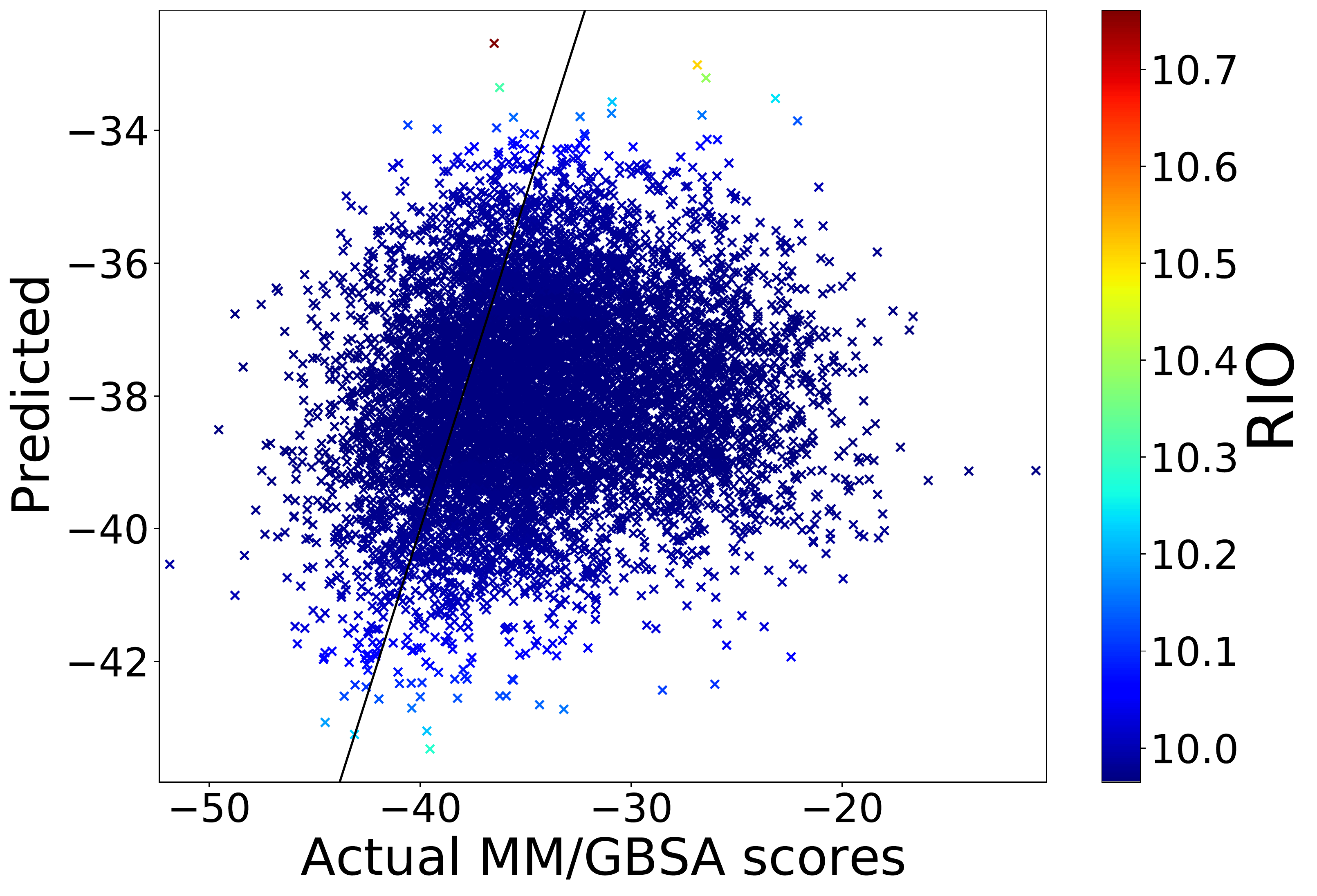}
\end{tabular}
\end{center}
\vspace{0in}
\caption{\label{fig:cluster2truePred} Test set actual MM/GBSA scores versus predicted plot from the model trained on cluster 2.}
\end{figure}
%

%------------------------------
%Cluster 3
%------------------------------

%
\begin{figure}[th]
\begin{center}
\begin{tabular}{cc}
{\large (a) Model trained on cluster 3 with MOE} & {\large (b) Model trained on cluster 3 with ECFP}\\ 
\includegraphics[width = 0.48\columnwidth, trim=0.3cm 0.3cm 0.3cm 0.3cm, clip=false]{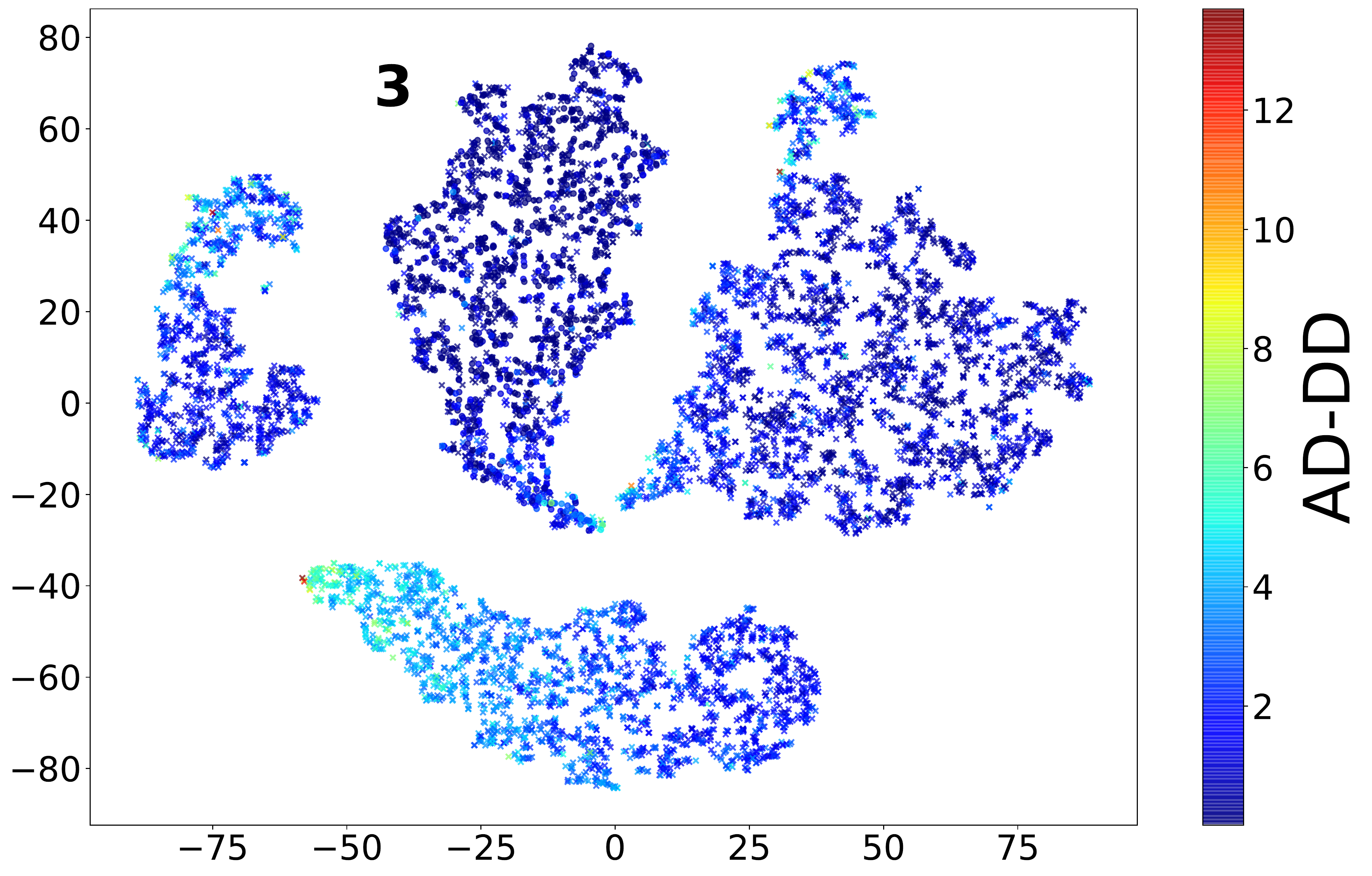}
&\includegraphics[width = 0.48\columnwidth, trim=0.3cm 0.3cm 0.3cm 0.3cm, clip=false]{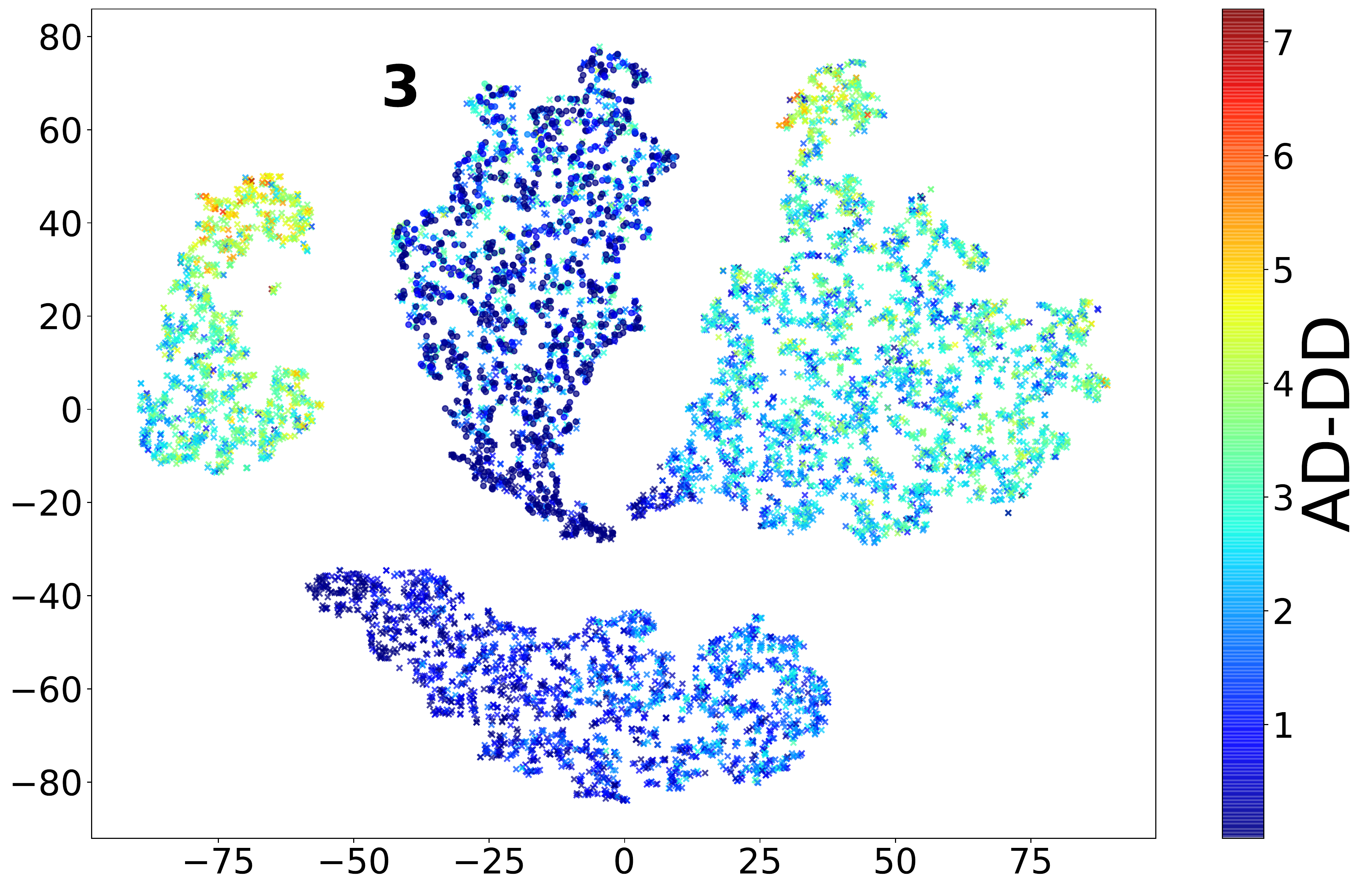}\\
\includegraphics[width = 0.48\columnwidth, trim=0.3cm 0.3cm 0.3cm 0.3cm, clip=false]{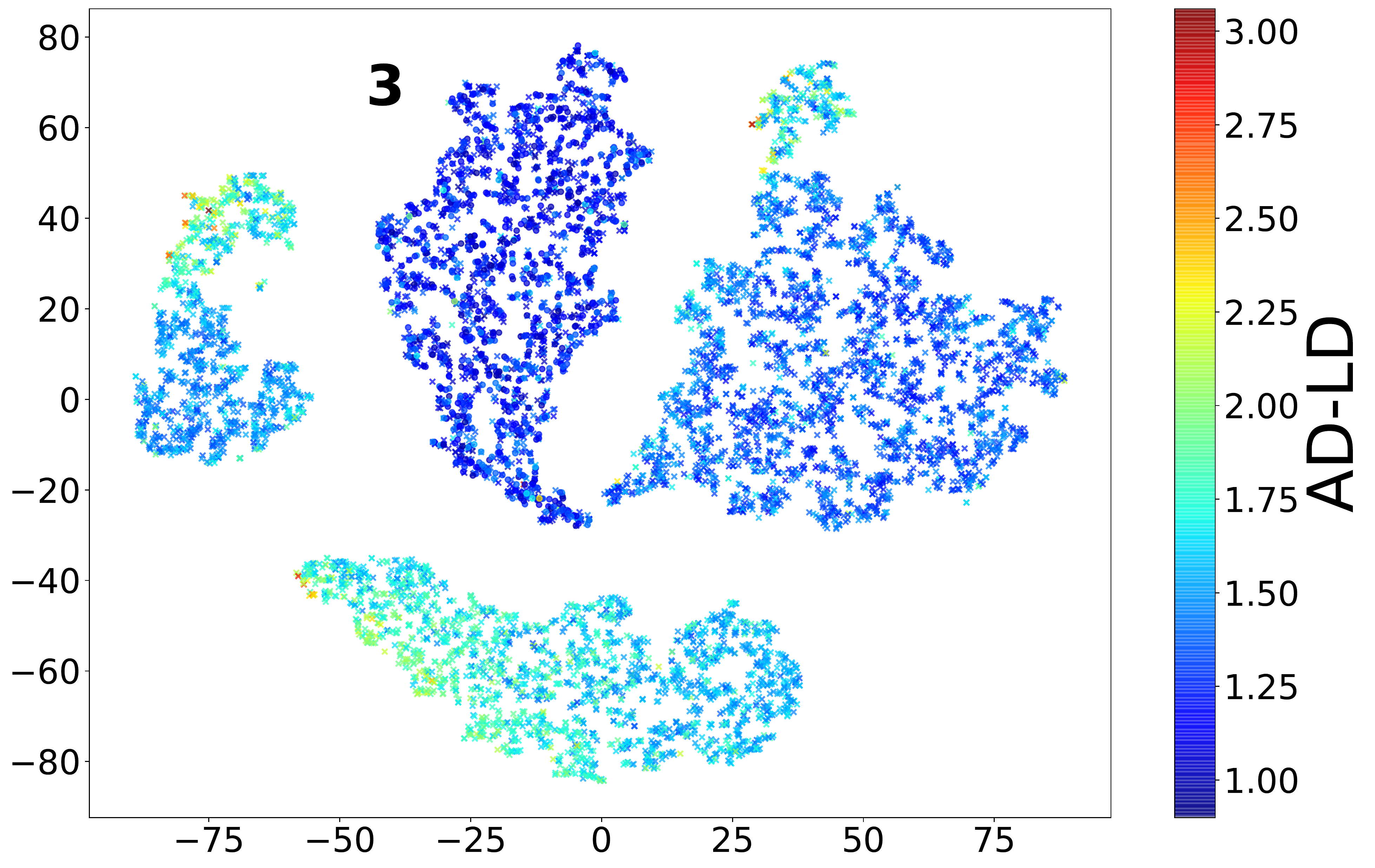}
&\includegraphics[width = 0.48\columnwidth, trim=0.3cm 0.3cm 0.3cm 0.3cm, clip=false]{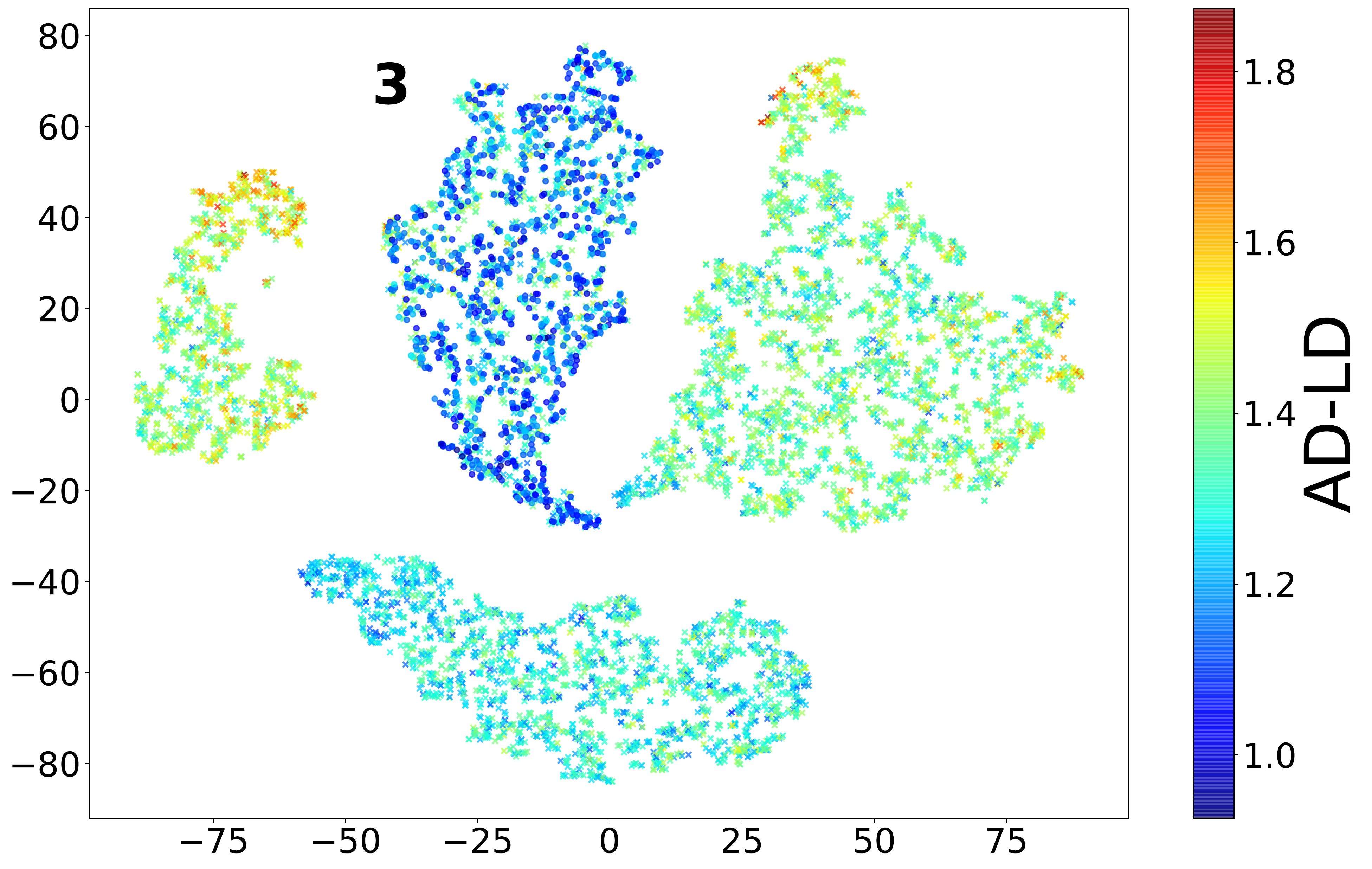}\\
\includegraphics[width = 0.48\columnwidth, trim=0.3cm 0.3cm 0.3cm 0.3cm, clip=false]{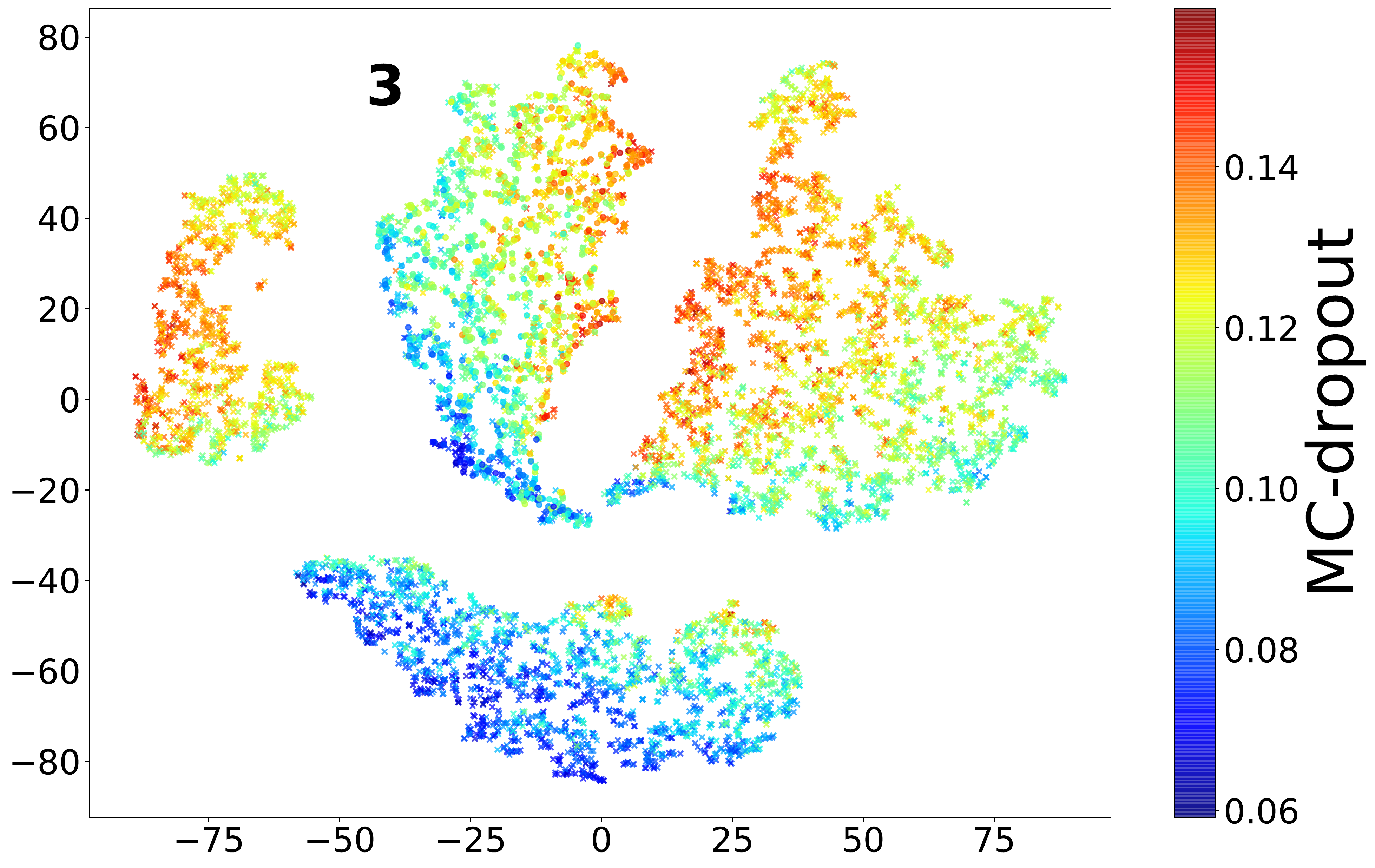}
&\includegraphics[width = 0.48\columnwidth, trim=0.3cm 0.3cm 0.3cm 0.3cm, clip=false]{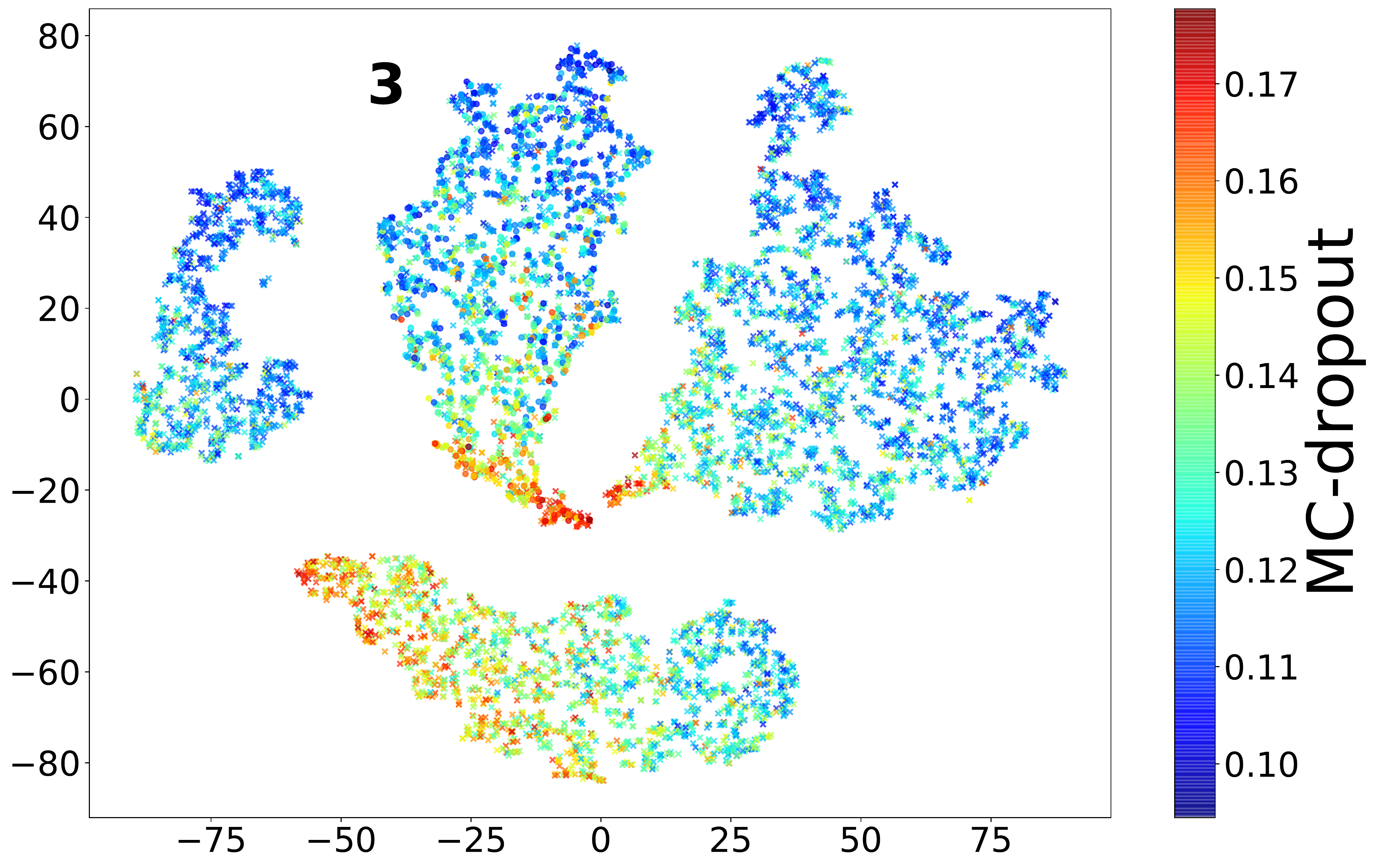}\\
\includegraphics[width = 0.48\columnwidth, trim=0.3cm 0.3cm 0.3cm 0.3cm, clip=false]{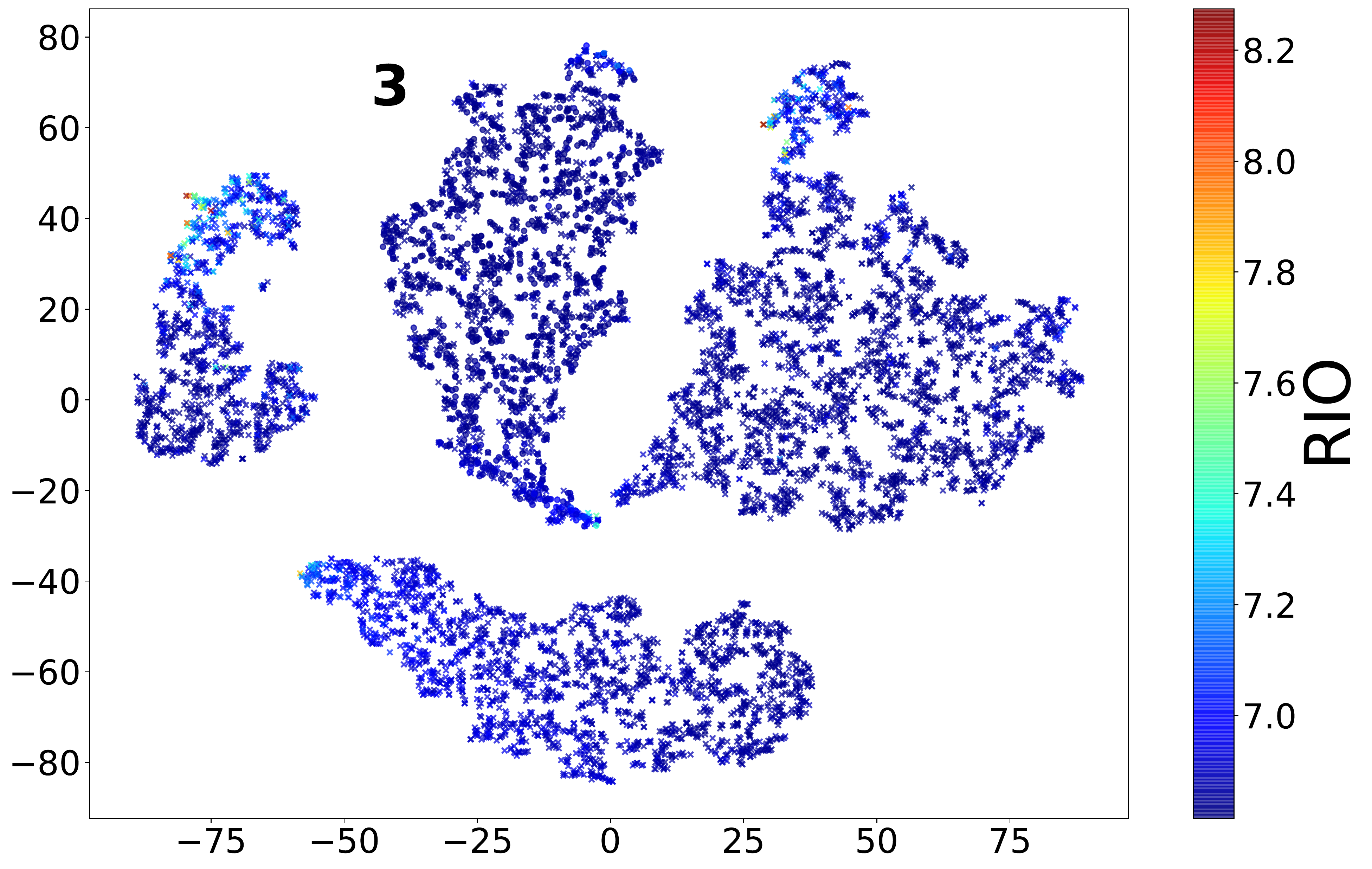}
&\includegraphics[width = 0.48\columnwidth, trim=0.3cm 0.3cm 0.3cm 0.3cm, clip=false]{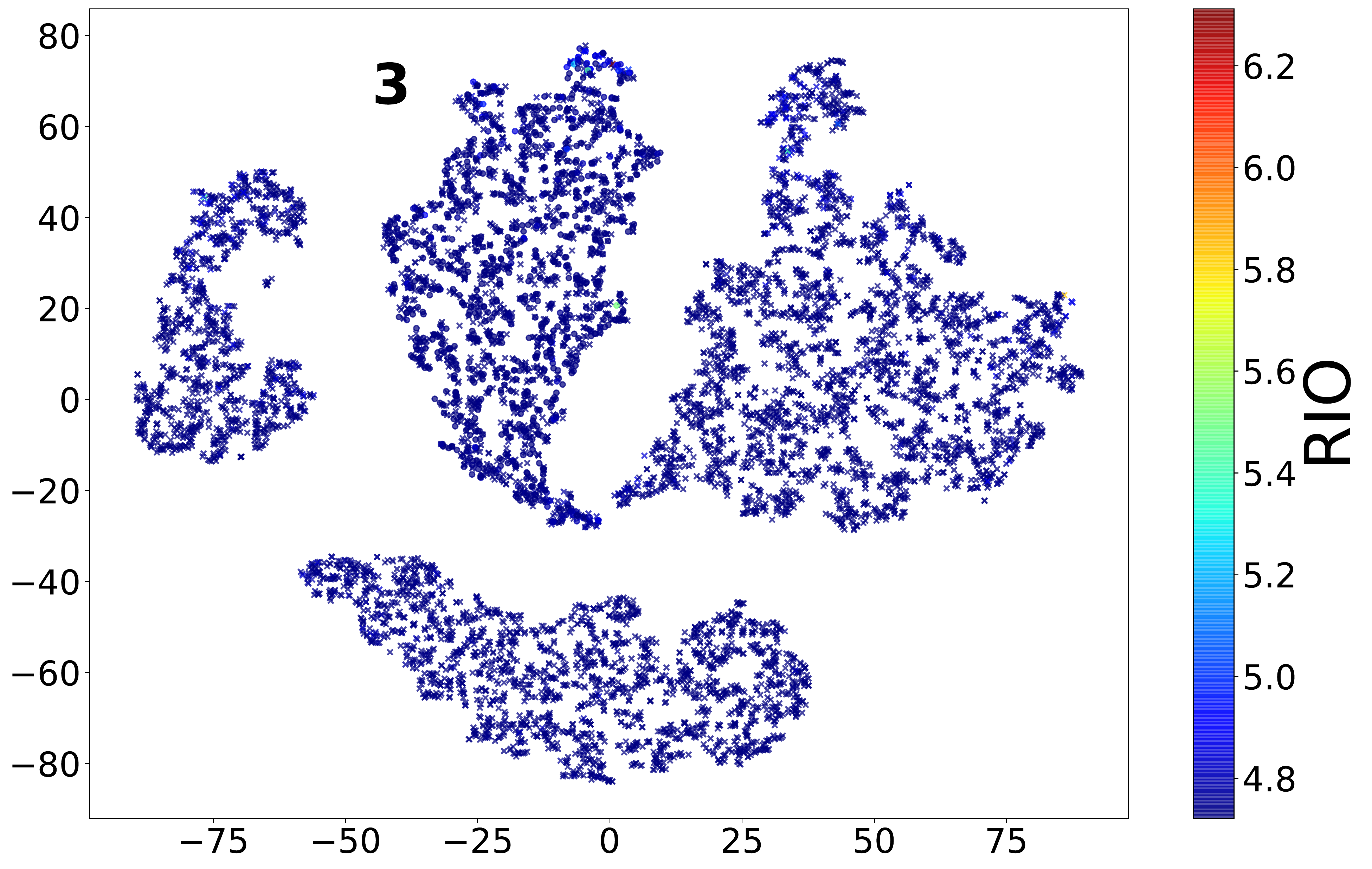}
\end{tabular}
\end{center}
\vspace{0in}
\caption{\label{fig:cluster3uq} Uncertainty values for the model trained on cluster 3.}
\end{figure}

\begin{figure}
\begin{center}
\begin{tabular}{cc}
{\large (a) Model trained on cluster 3 with MOE} & {\large (b) Model trained on cluster 3 with ECFP}\\ 
\includegraphics[width = 0.46\columnwidth, trim=0.3cm 0.3cm 0.3cm 0.3cm, clip=false]{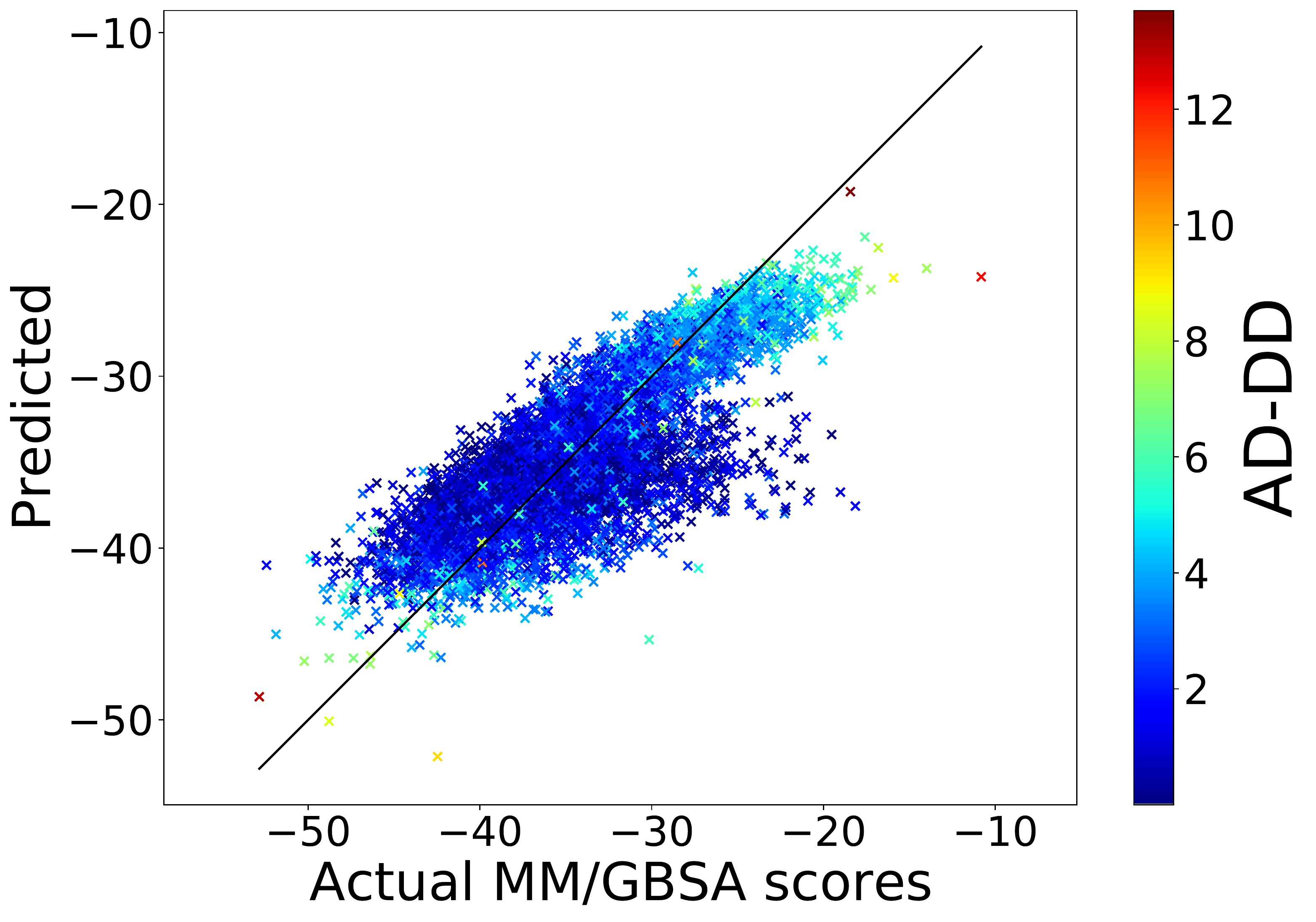}
&\includegraphics[width = 0.46\columnwidth, trim=0.3cm 0.3cm 0.3cm 0.3cm, clip=false]{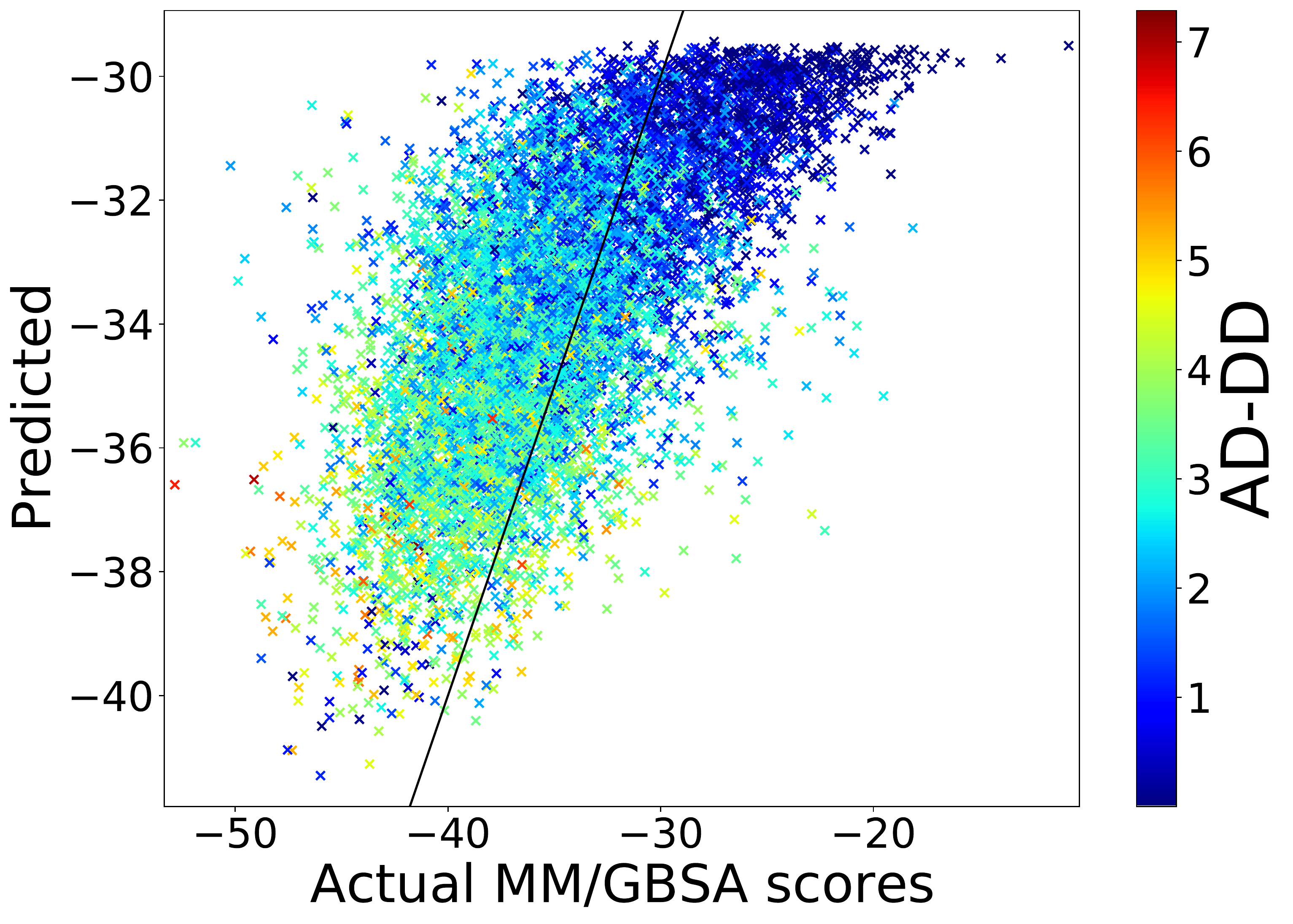}\\
\includegraphics[width = 0.46\columnwidth, trim=0.3cm 0.3cm 0.3cm 0.3cm, clip=false]{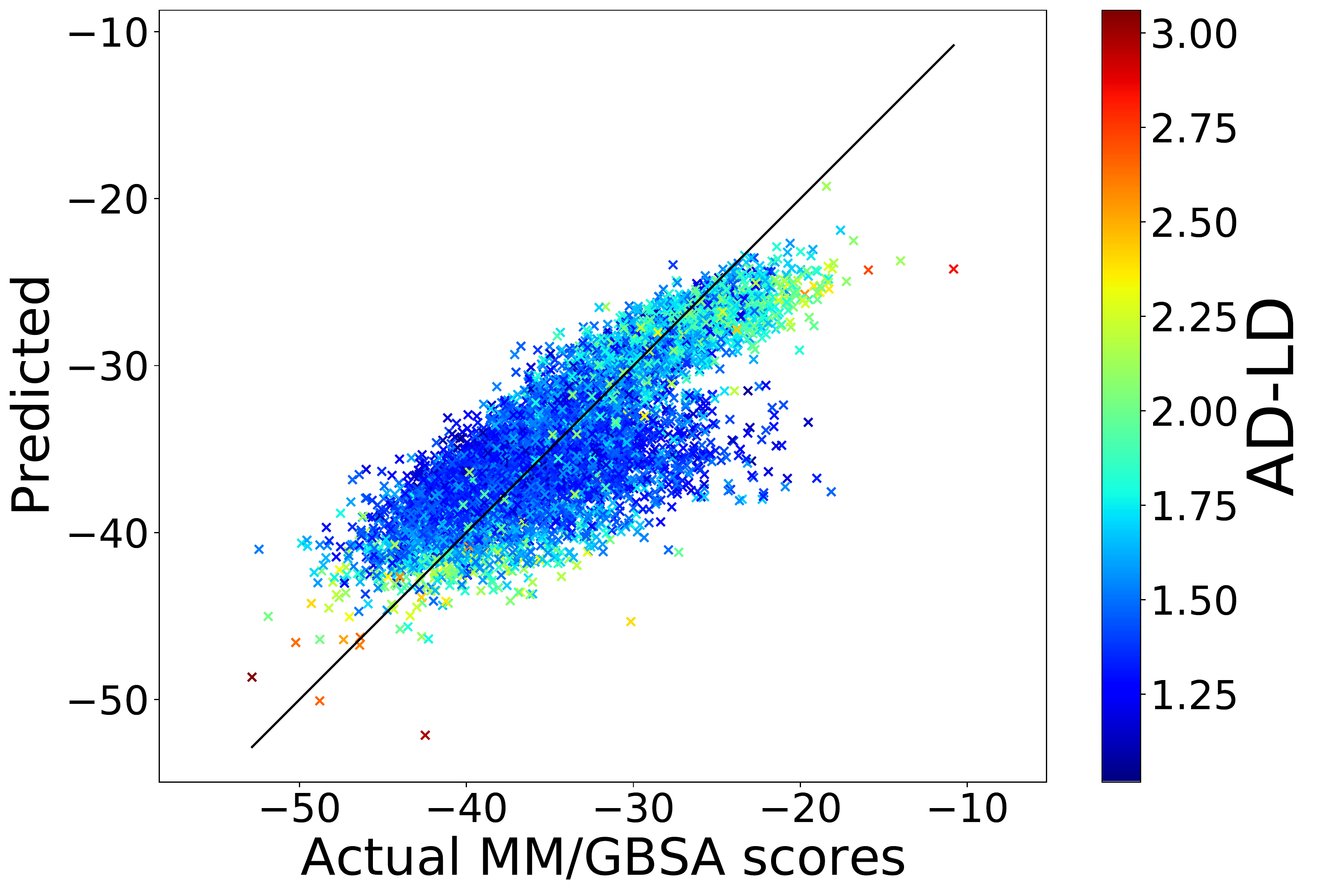}
&\includegraphics[width = 0.46\columnwidth, trim=0.3cm 0.3cm 0.3cm 0.3cm, clip=false]{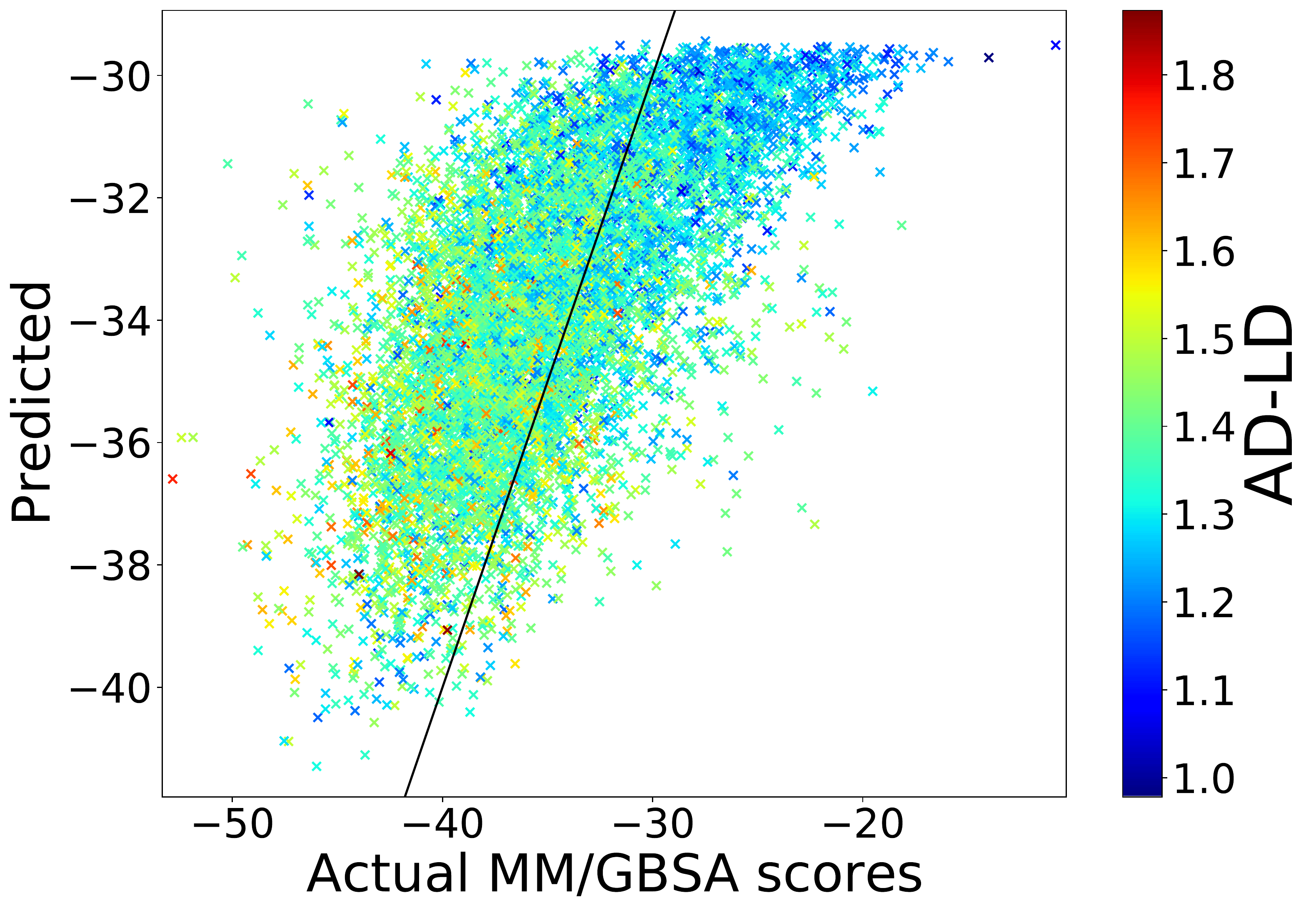}\\
\includegraphics[width = 0.46\columnwidth, trim=0.3cm 0.3cm 0.3cm 0.3cm, clip=false]{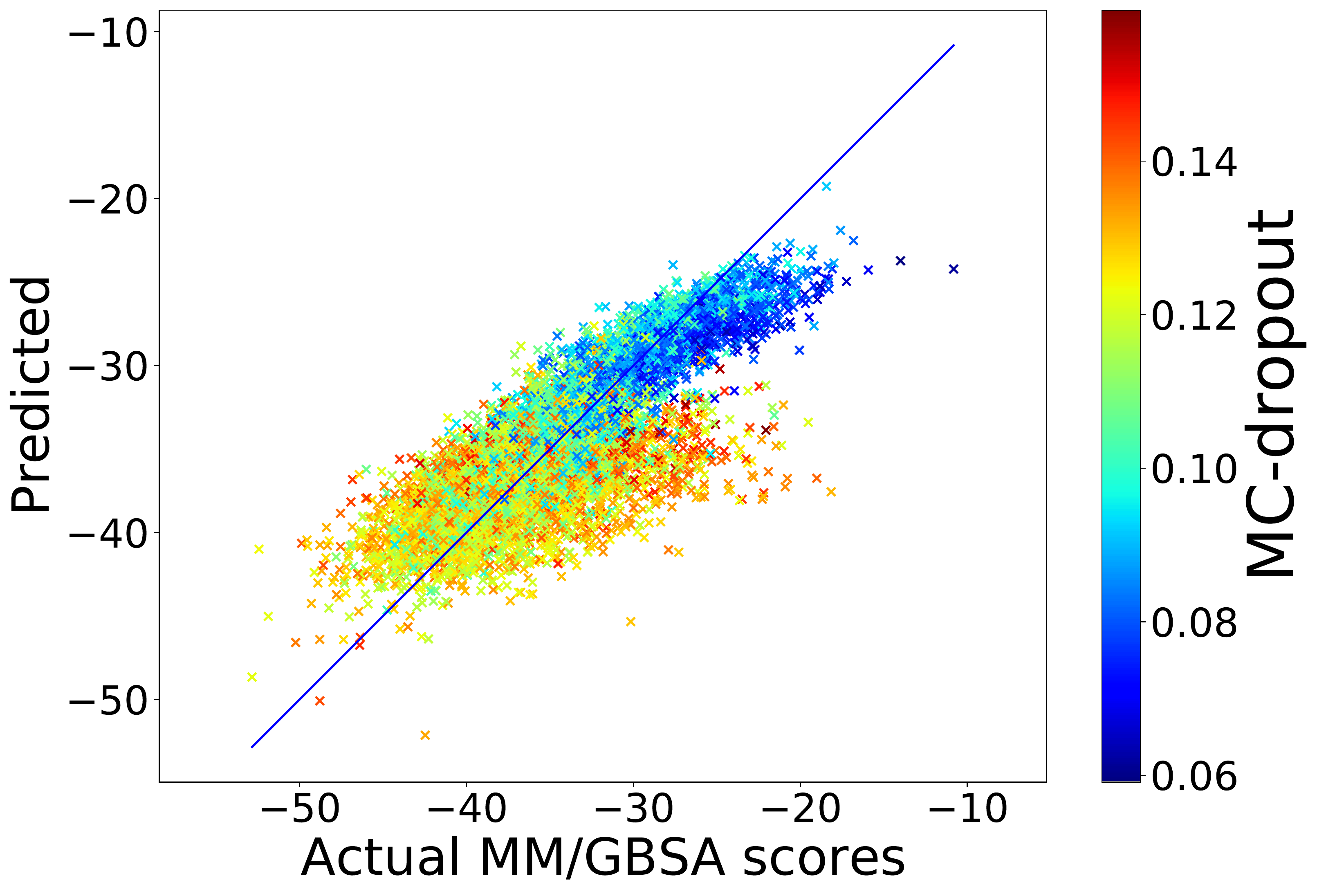}
&\includegraphics[width = 0.46\columnwidth, trim=0.3cm 0.3cm 0.3cm 0.3cm, clip=false]{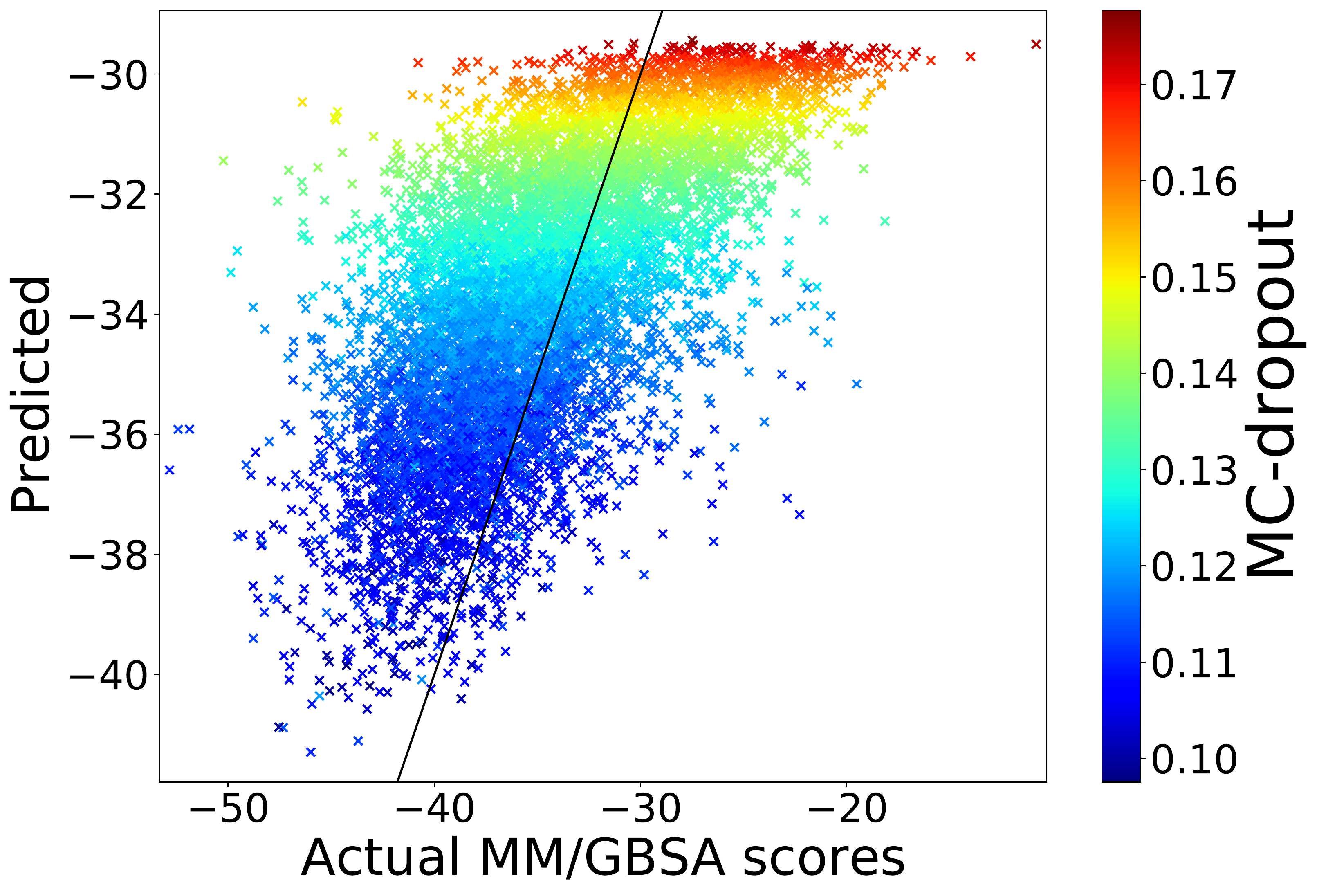}\\
\includegraphics[width = 0.46\columnwidth, trim=0.3cm 0.3cm 0.3cm 0.3cm, clip=false]{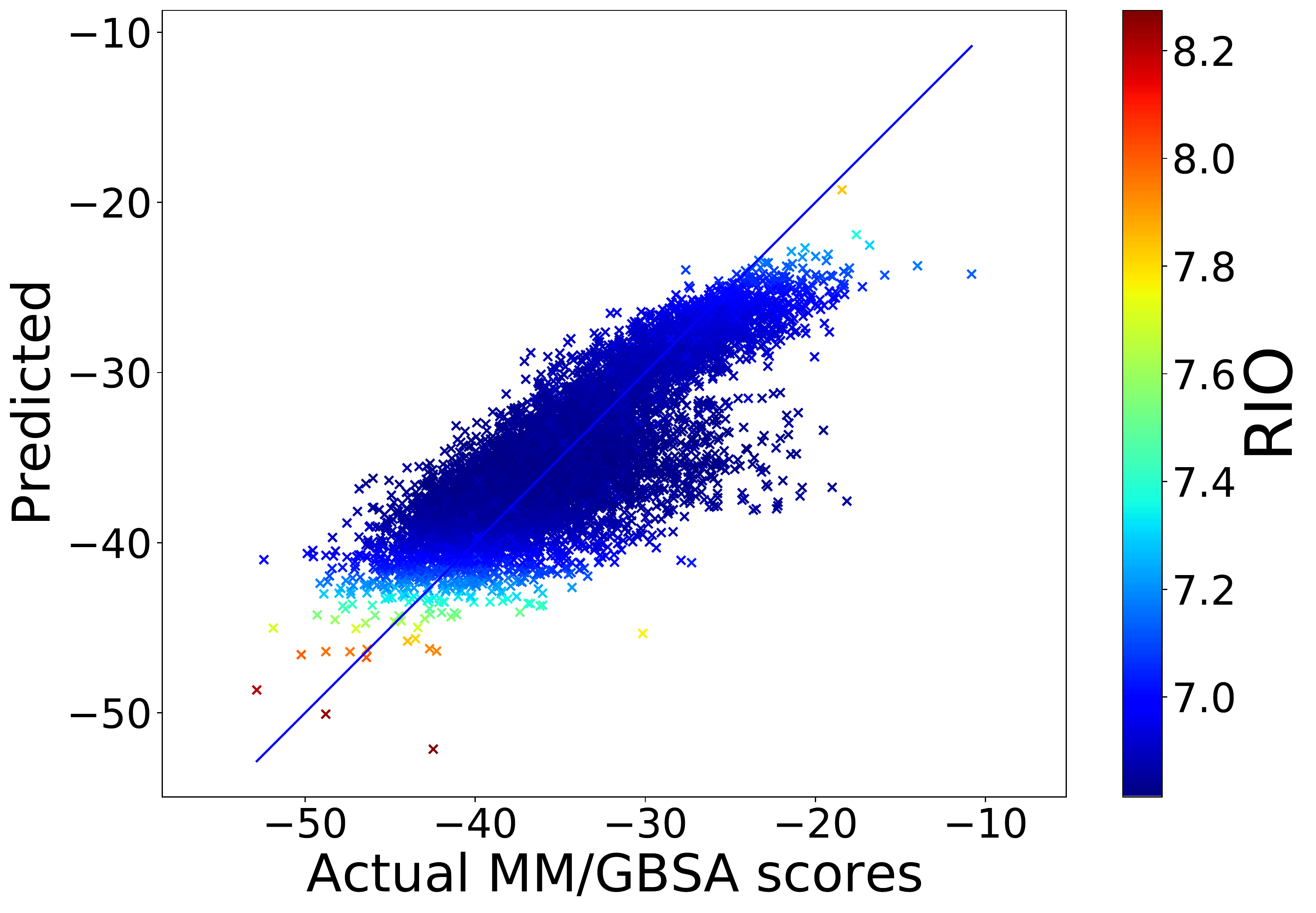}
&\includegraphics[width = 0.46\columnwidth, trim=0.3cm 0.3cm 0.3cm 0.3cm, clip=false]{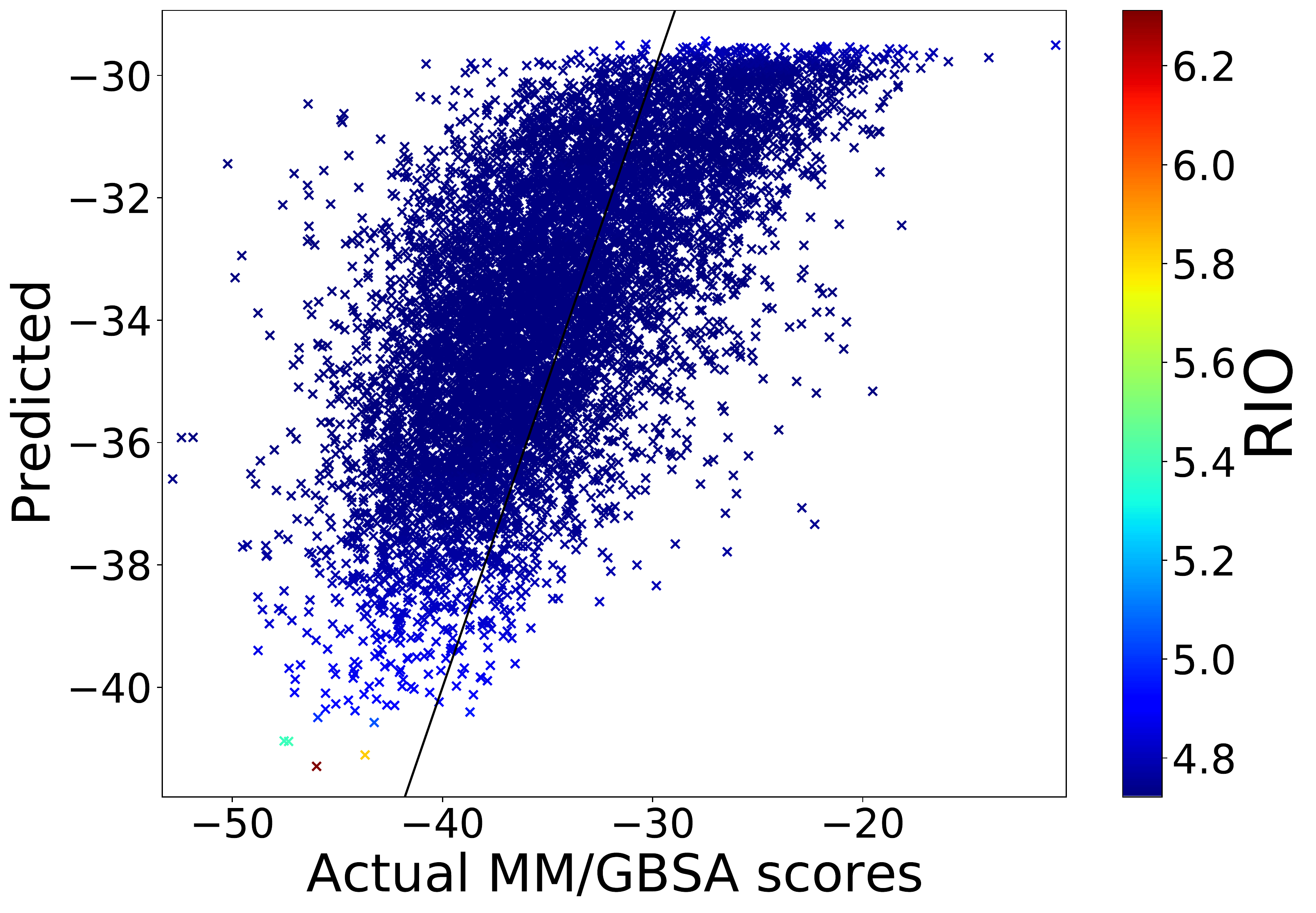}
\end{tabular}
\end{center}
\vspace{0in}
\caption{\label{fig:cluster3truePred} Test set actual MM/GBSA scores versus predicted plot from the model trained on cluster 3.}
\end{figure}
%
%------------------------------
%Cluster 4
%------------------------------

%
\begin{figure}[th]
\begin{center}
\begin{tabular}{cc}
{\large (a) Model trained on cluster 4 with MOE} & {\large (b) Model trained on cluster 4 with ECFP}\\ 
\includegraphics[width = 0.48\columnwidth, trim=0.3cm 0.3cm 0.3cm 0.3cm, clip=false]{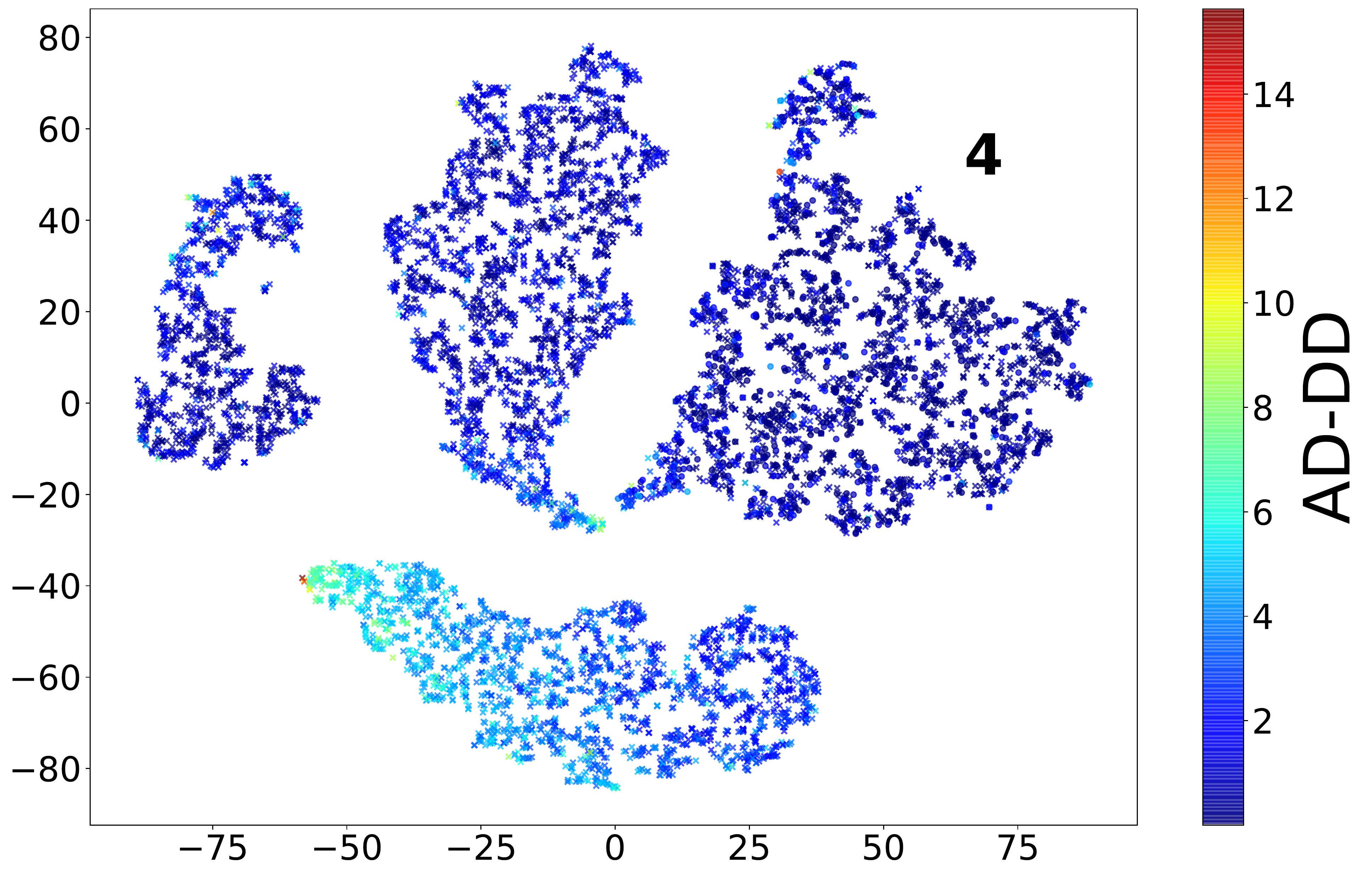}
&\includegraphics[width = 0.48\columnwidth, trim=0.3cm 0.3cm 0.3cm 0.3cm, clip=false]{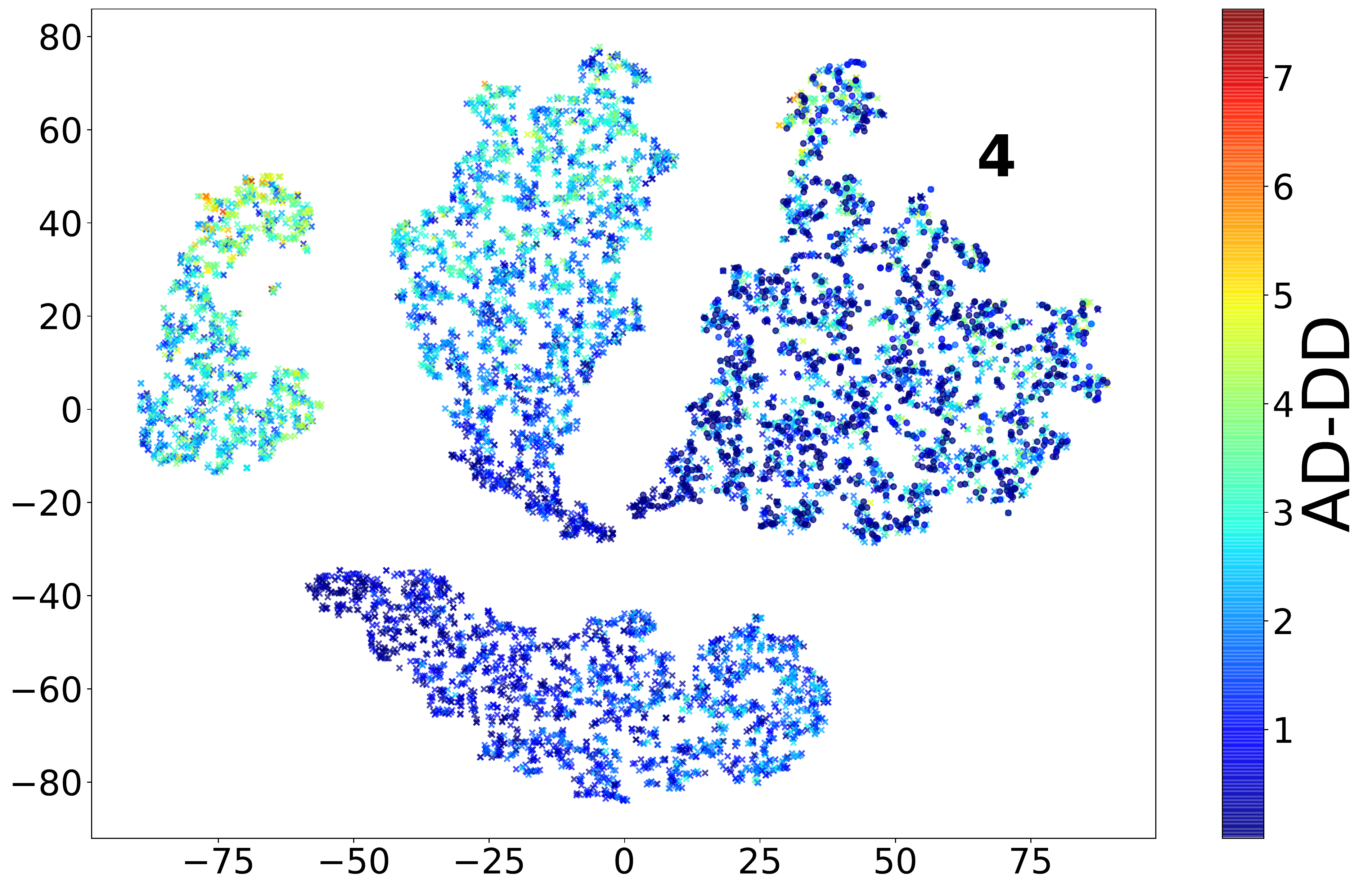}\\
\includegraphics[width = 0.48\columnwidth, trim=0.3cm 0.3cm 0.3cm 0.3cm, clip=false]{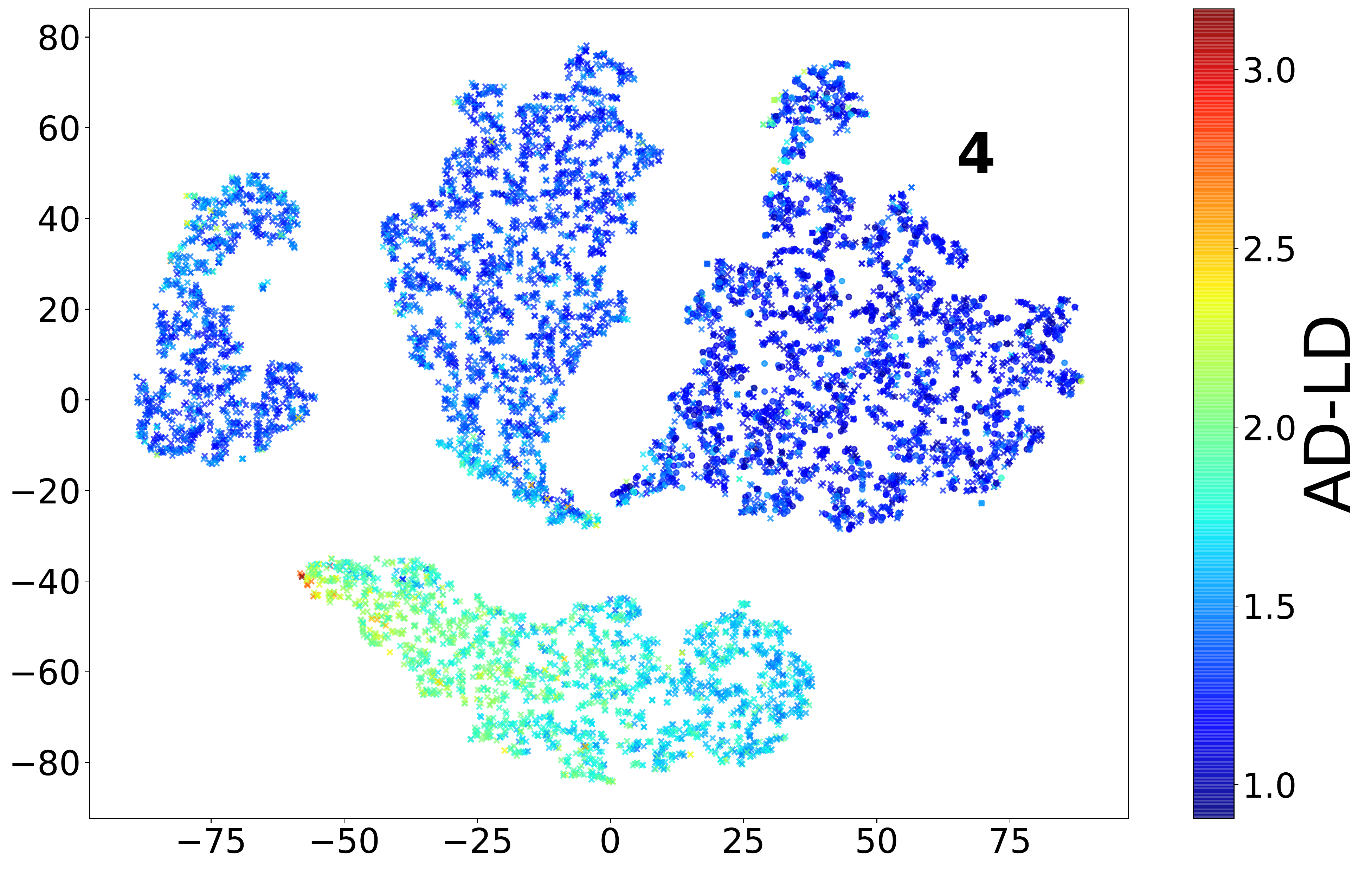}
&\includegraphics[width = 0.48\columnwidth, trim=0.3cm 0.3cm 0.3cm 0.3cm, clip=false]{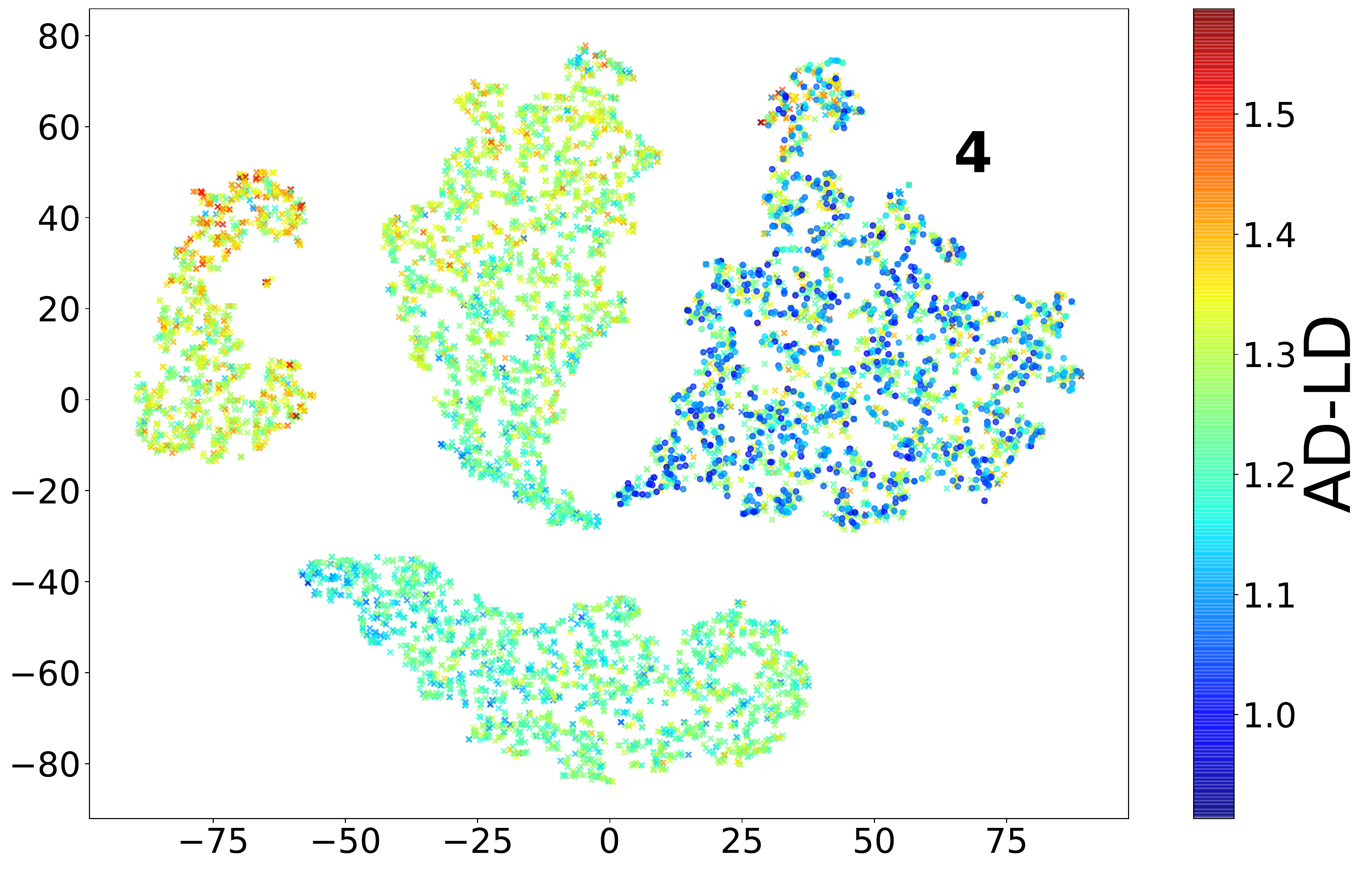}\\
\includegraphics[width = 0.48\columnwidth, trim=0.3cm 0.3cm 0.3cm 0.3cm, clip=false]{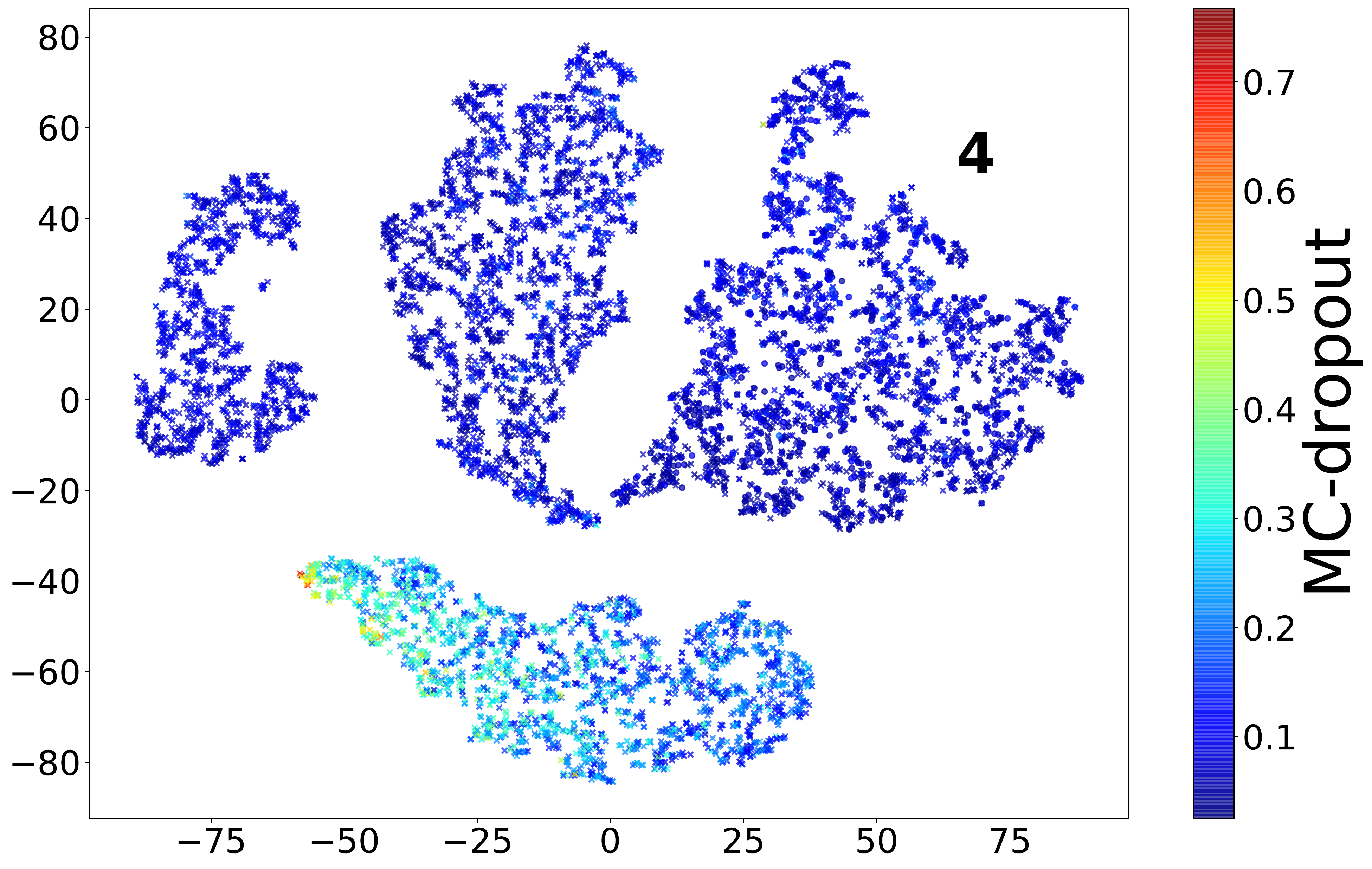}
&\includegraphics[width = 0.48\columnwidth, trim=0.3cm 0.3cm 0.3cm 0.3cm, clip=false]{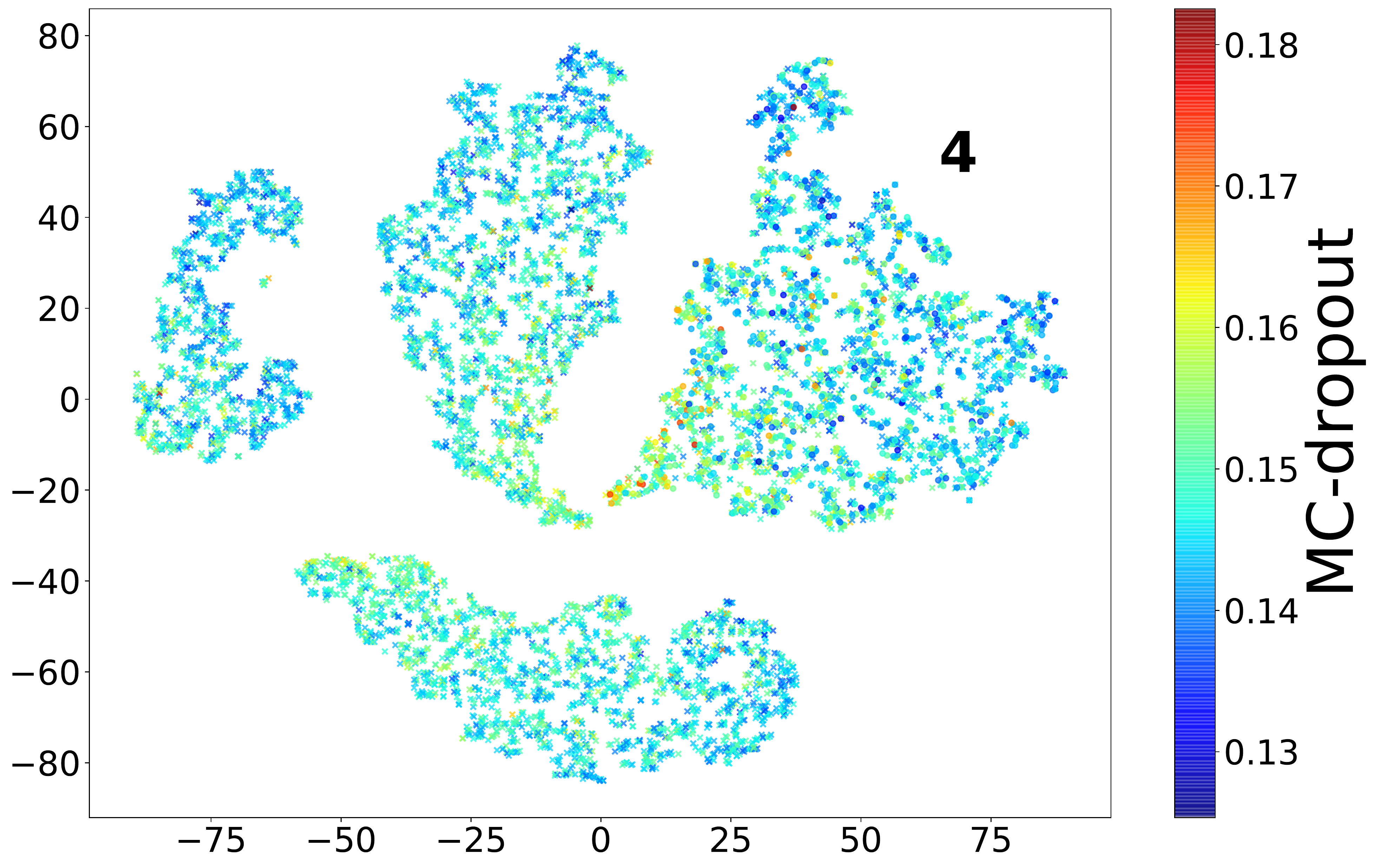}\\
\includegraphics[width = 0.48\columnwidth, trim=0.3cm 0.3cm 0.3cm 0.3cm, clip=false]{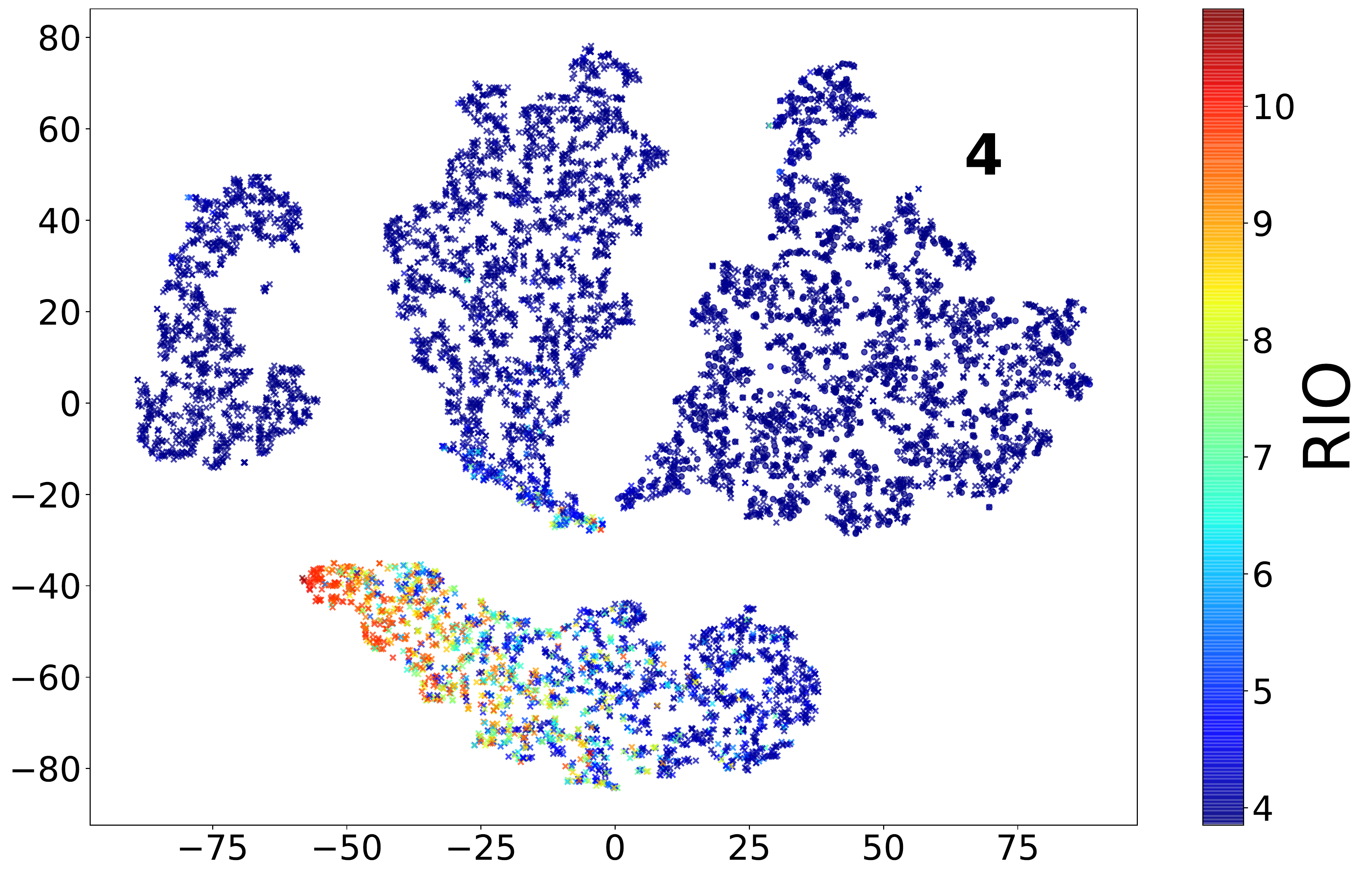}
&\includegraphics[width = 0.48\columnwidth, trim=0.3cm 0.3cm 0.3cm 0.3cm, clip=false]{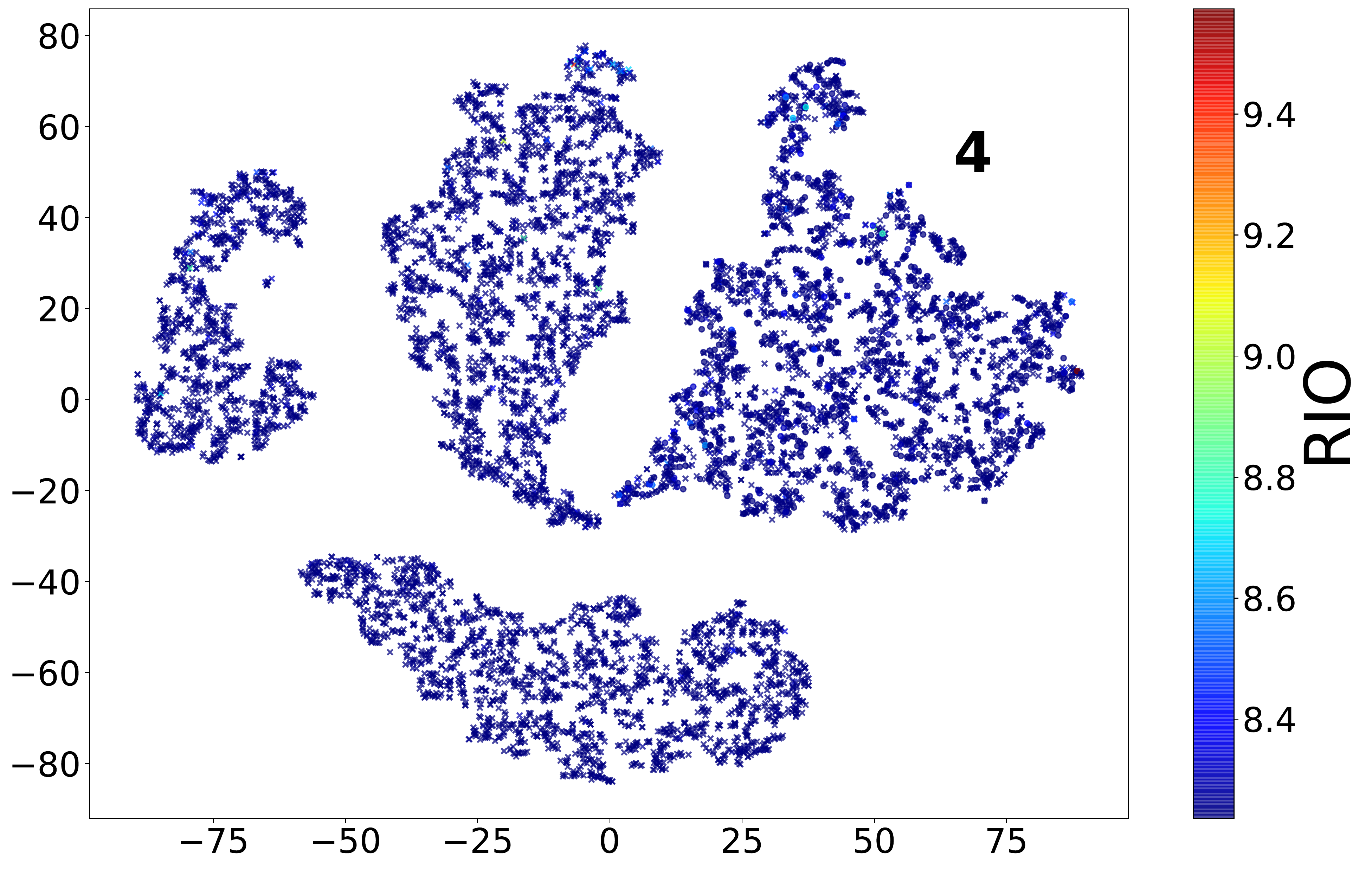}
\end{tabular}
\end{center}
\vspace{0in}
\caption{\label{fig:cluster4uq} Uncertainty values for the model trained on cluster 4.}
\end{figure}

\begin{figure}
\begin{center}
\begin{tabular}{cc}
{\large (a) Model trained on cluster 4 with MOE} & {\large (b) Model trained on cluster 4 with ECFP}\\ 
\includegraphics[width = 0.46\columnwidth, trim=0.3cm 0.3cm 0.3cm 0.3cm, clip=false]{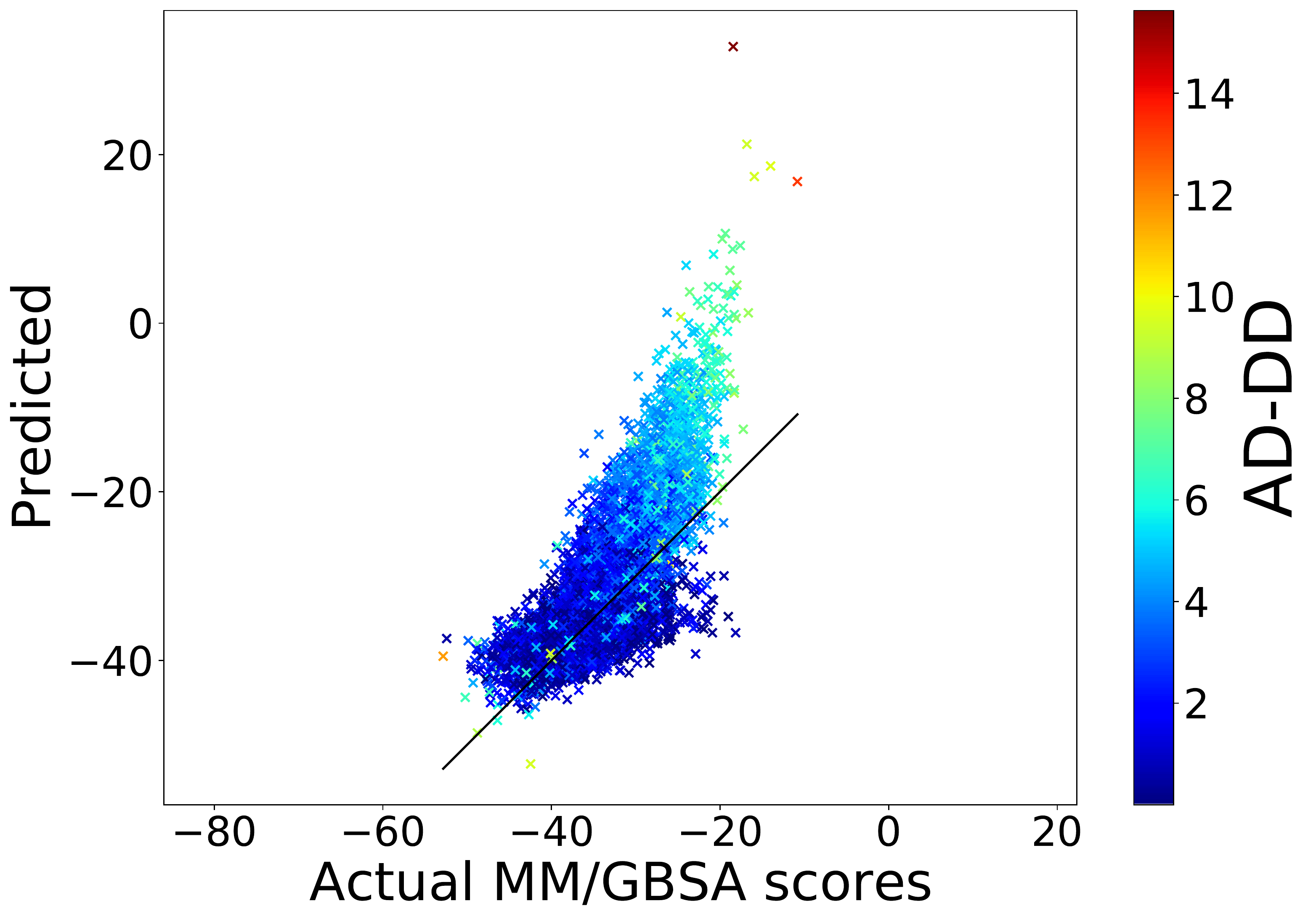}
&\includegraphics[width = 0.46\columnwidth, trim=0.3cm 0.3cm 0.3cm 0.3cm, clip=false]{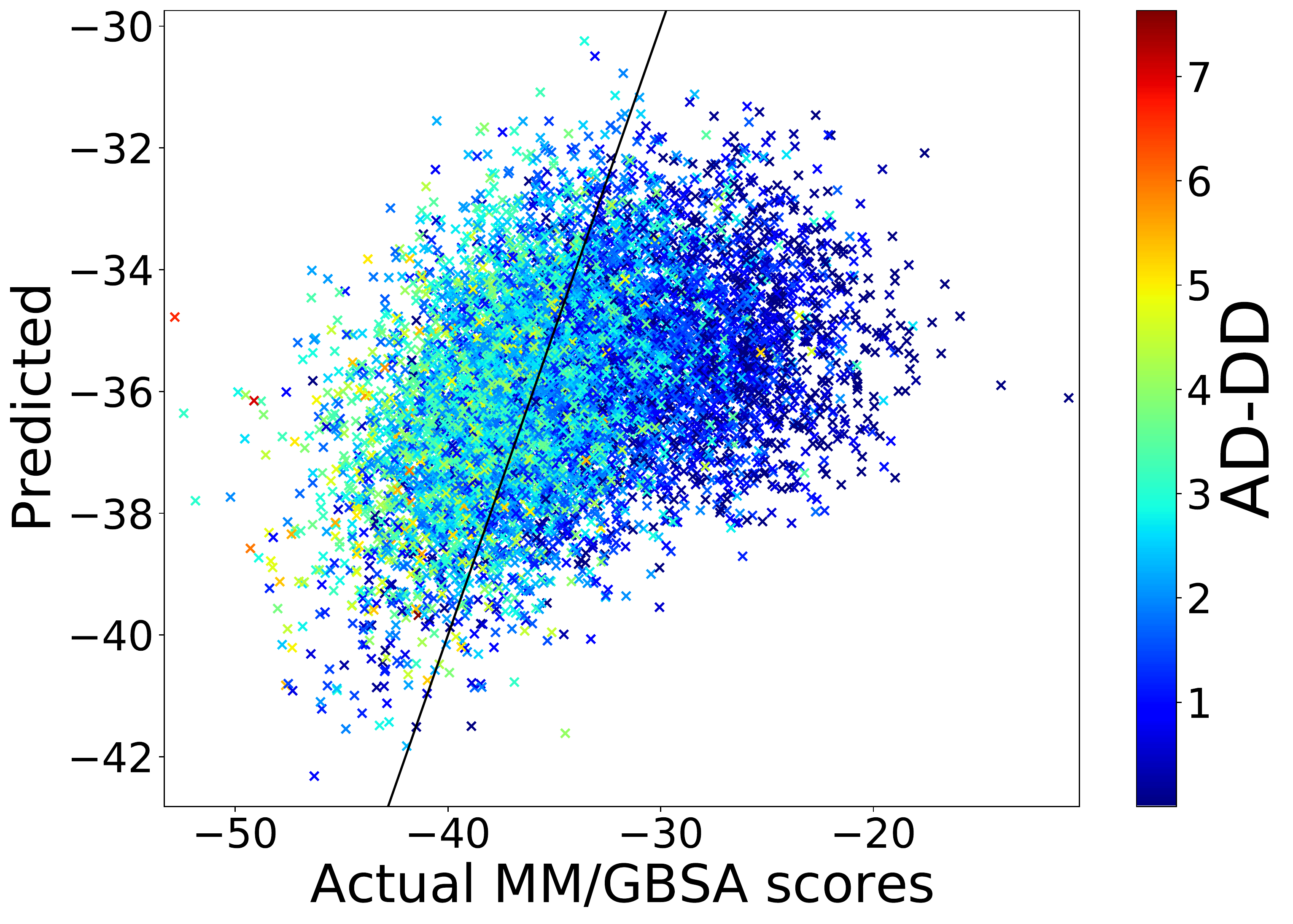}\\
\includegraphics[width = 0.46\columnwidth, trim=0.3cm 0.3cm 0.3cm 0.3cm, clip=false]{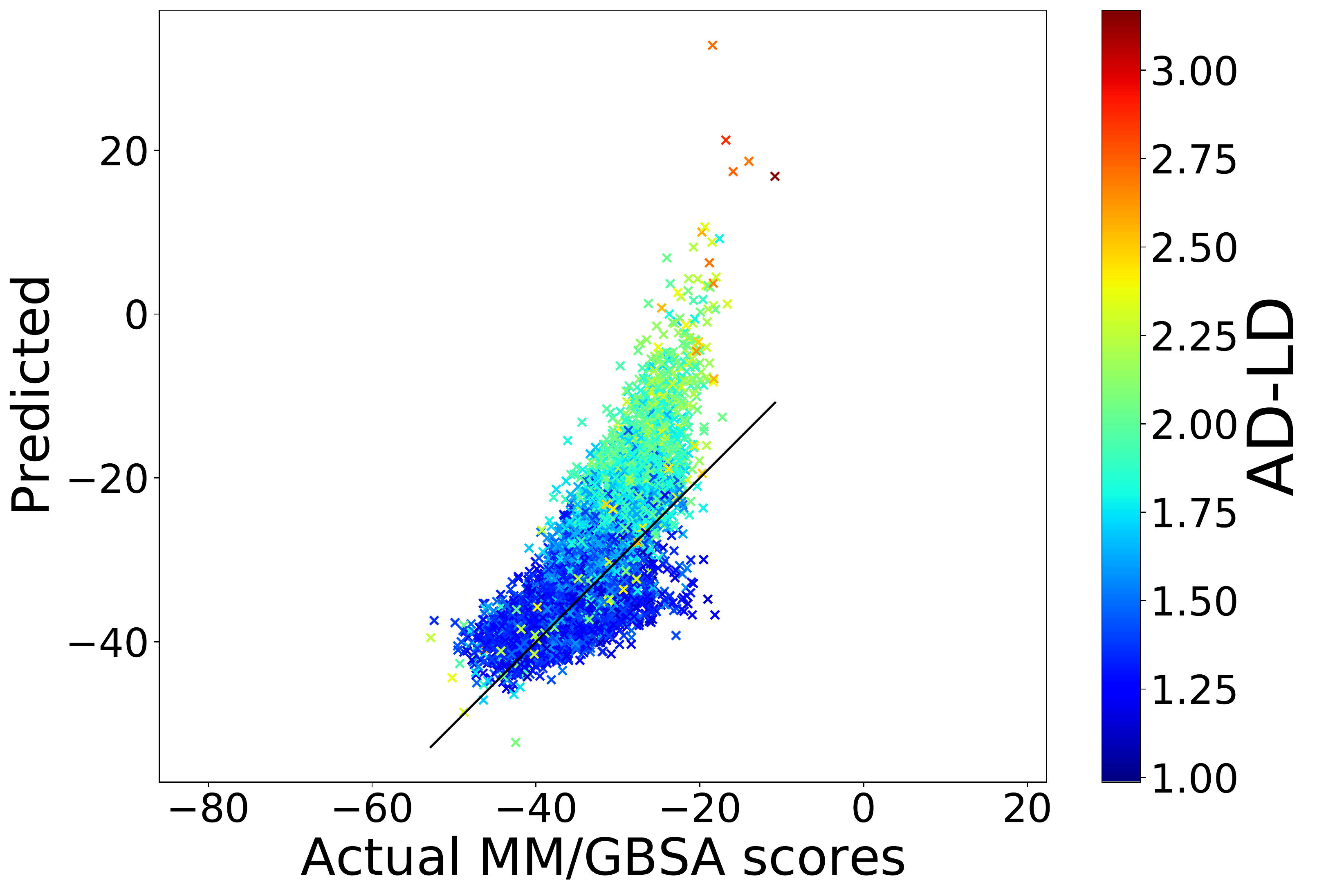}
&\includegraphics[width = 0.46\columnwidth, trim=0.3cm 0.3cm 0.3cm 0.3cm, clip=false]{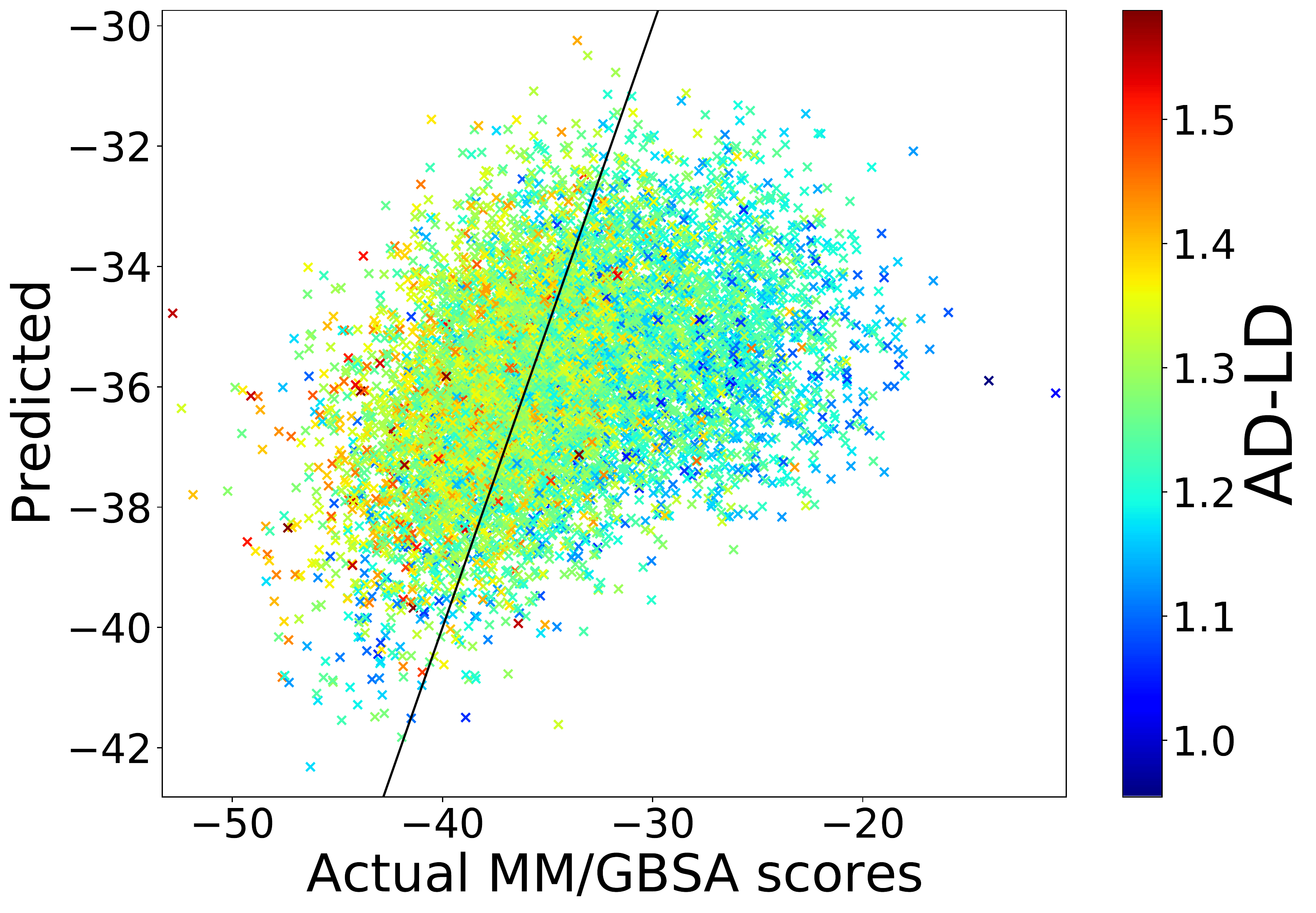}\\
\includegraphics[width = 0.46\columnwidth, trim=0.3cm 0.3cm 0.3cm 0.3cm, clip=false]{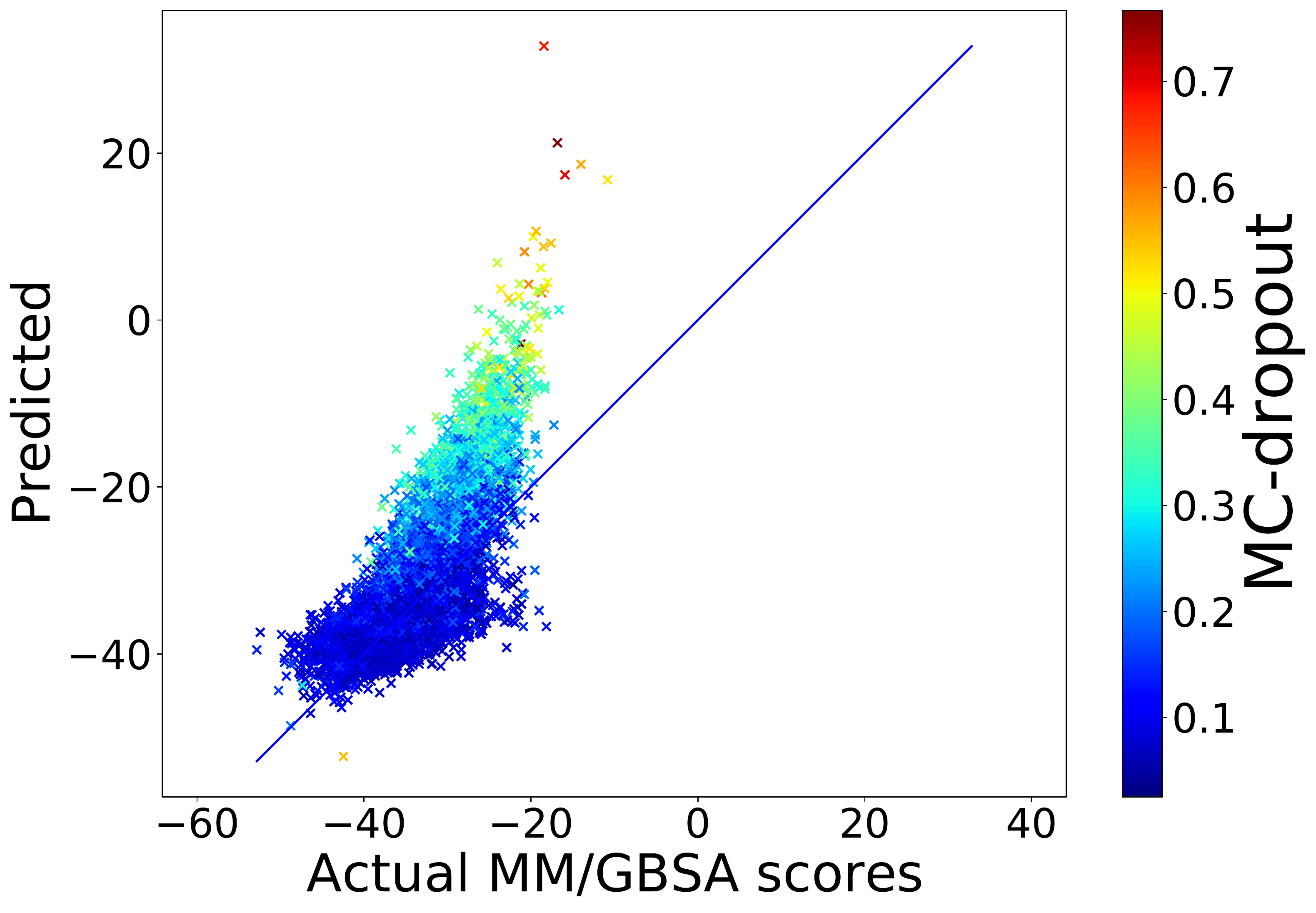}
&\includegraphics[width = 0.46\columnwidth, trim=0.3cm 0.3cm 0.3cm 0.3cm, clip=false]{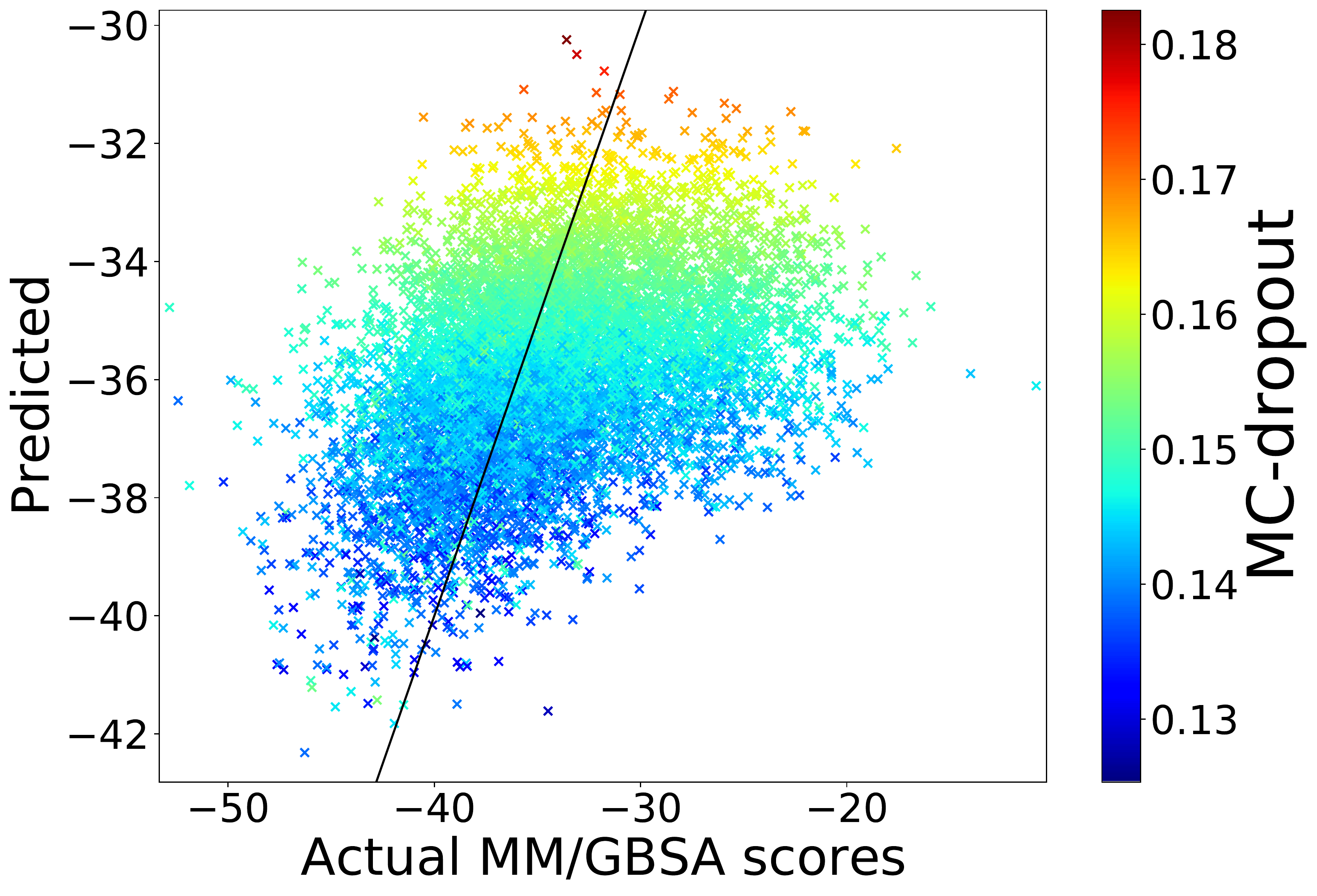}\\
\includegraphics[width = 0.46\columnwidth, trim=0.3cm 0.3cm 0.3cm 0.3cm, clip=false]{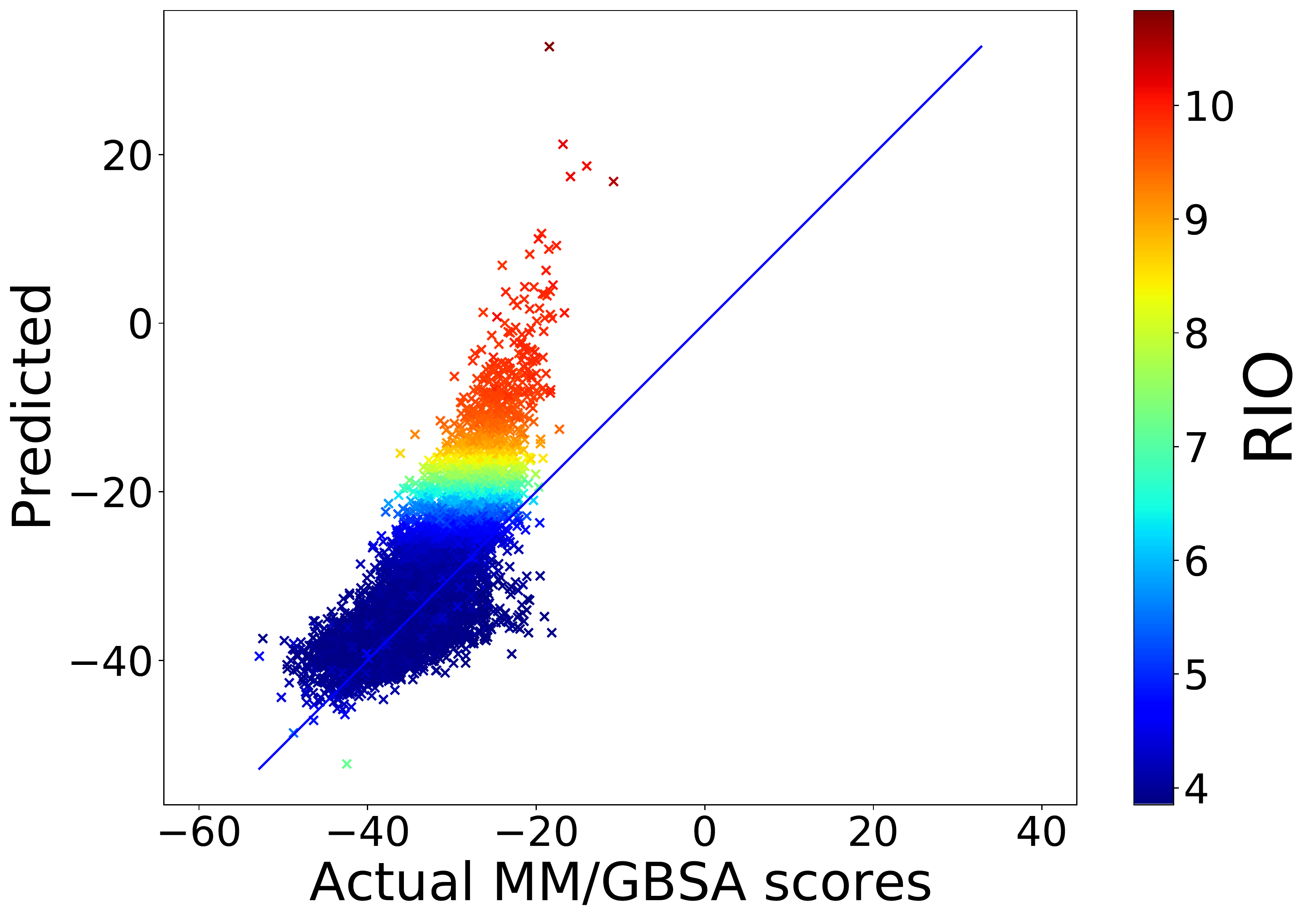}
&\includegraphics[width = 0.46\columnwidth, trim=0.3cm 0.3cm 0.3cm 0.3cm, clip=false]{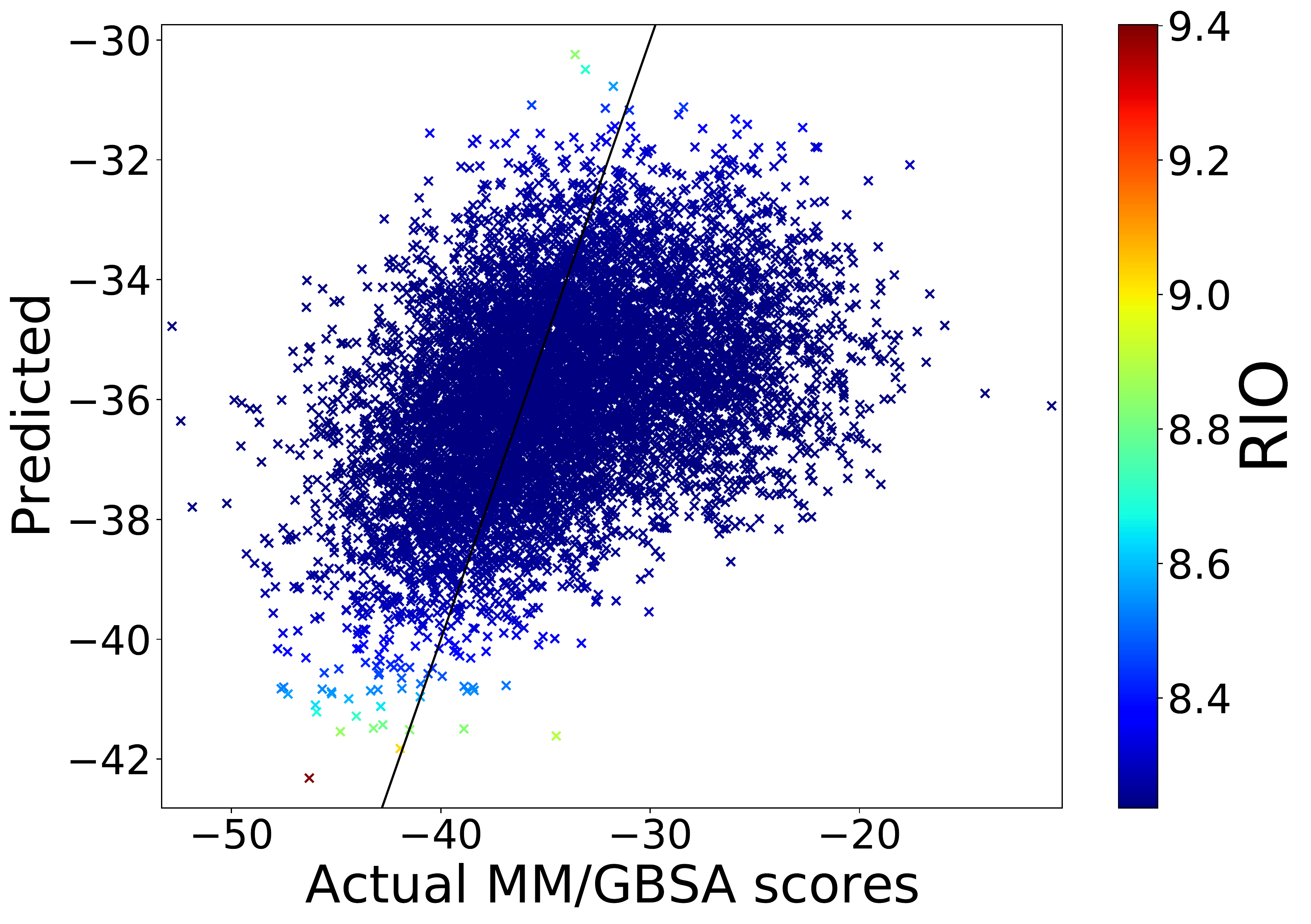}
\end{tabular}
\end{center}
\vspace{0in}
\caption{\label{fig:cluster4truePred} Test set actual MM/GBSA scores versus predicted plot from the model trained on cluster 4.}
\end{figure}

Since we use the t-SNE plot to generate the clusters where distances in the original data space are not preserved in the 2D feature space. The clusters in the plot may be shown to appear in non-clustered data. Values generated by the AD method using empirical distance distribution (AD-DD) measuring the dissimilarities to the training set can help realize the actual distances. Figure~\ref{fig:cluster1uq}, ~\ref{fig:cluster2uq}, ~\ref{fig:cluster3uq} and ~\ref{fig:cluster4uq} display the uncertainty estimations of the point predictions using the four uncertainty quantification methods for the models trained on cluster 1, 2, 3 and 4, respectively. The AD-DD uncertainty values are significantly lower in the training clusters than in the rest of the chemical compounds, especially in cluster 1, 2 and 3. When using the cluster 4 as the training set, AD-DD method does not give as distinctly smaller values for the training data. It appears that chemical compounds described by MOE descriptors in cluster 4 are close to data points in cluster 2 and 3, indicating that the cluster covers a wide range of chemical compounds. Furthermore, data points represented by ECFP features in cluster 4 spread out from each other, suggesting non-clustered data. 

Recall that distance-based novelty detection needs a threshold to decide whether a new compound is actually novel~\cite{Mathea2016}. If we use a 95\% quantile of the $k$-nearest neighbor within the training set distance distribution as a threshold, the corresponding z-score is around 1.6 in a standard normal distribution. We can use the value 1.6 for AD-DD values as a threshold to identify novel test compounds. We can see that this threshold distinguishes the training cluster 1, 2 and 3 from their corresponding test set clusters. 

The AD method using local density (AD-LD) highlights more data points with larger values than the AD-DD method. The distinctiveness between the training and the test sets using AD-LD is less significant than AD-DD. Only the models trained on cluster 1 with MOE descriptors gives significantly lower AD-LD values on the training cluster than the rest of chemical compounds. Moreover, the models trained with ECFP features yield lower ranges of AD-LD values, which indicates that data points in ECFP features spread out and may not form definite clusters.

Models using different compound representation may have opposite uncertainty values from MC-dropout. The model uncertainties (evaluated by MC-dropout) that appear high in one compound representations may be low in other compound representations. The third rows of the figures display the UQ values from MC-dropout used to present the model uncertainties.  For the models trained on cluster 1 and cluster 3, MC-dropout gives opposite degrees of uncertainties when using different compound representations. As shown in Figure~\ref{fig:cluster1uq} and~\ref{fig:cluster3uq} the MC-dropout method gives high uncertainty values with MOE descriptors while giving low uncertainty values with ECFP features for most predictions. To investigate which uncertainty values to trust, we compare the UQ values with the prediction errors in the Figure~\ref{fig:cluster1truePred} and~\ref{fig:cluster3truePred}, which display the actual MM/GBSA scores versus the predicted, colored by the UQ values. The distance from a data point to the diagonal indicate its prediction error. We found that the UQ values using MC-dropout on the MOE descriptors tend to be high on the prediction values that have data points far away from the diagonal and low on the predictions that are closer to the diagonal, indicating that MC-dropout on the MOE descriptors can reflect possible prediction errors. We also know that the models trained on the ECFP features do not perform well on the predictions. We may conclude that the UQ values from MC-dropout are more reliable on the MOE descriptors than on the ECFP features possibly due to better trained models with MOE descriptors. Furthermore, MC-dropout does not necessarily give lower uncertainties in the training than in the test set. There are high MC-dropout uncertainty values in the training clusters where the other three methods tend to have low uncertainty values. 

RIO method gives near constant uncertainty values to the majority of the chemical compounds, especially on the models trained with ECFP features. There are only a few data points that have relative high RIO uncertainty values. In general RIO picked up part of high uncertainties similar to AD-DD when the models are trained with MOE descriptors. Exceptionally, for the model trained on cluster 4 with MOE descriptors RIO marks particularly high uncertainty values on several chemical compounds; while all other methods mark similar high uncertainty values but they are less distinct than what RIO gives as shown in Figure~\ref{fig:cluster4uq} and~\ref{fig:cluster4truePred}.  

%the models trained on cluster 1 and cluster 3 using MOE descriptors give opposite high and low uncertainty values of the dropout method (that provides the model uncertainties) to the models using ECFP features. 

%\clearpage

%--------------------------------------------------------------
%
\subsection{Uncertainties v.s. prediction errors}
%
%--------------------------------------------------------------

When using NN model predictions to direct experimental design, unanticipated imprecision wastes valuable time and resources for drug discovery applications~\cite{Hirschfeld2020}. We examine whether the UQ methods are able to identify the prediction errors when using NN models for new chemical compounds. We plot the actual MM/GBSA scores versus the predicted values of the chemical compounds from the test sets. We color their uncertainty values from the four evaluation methods for the models trained on cluster 1, 2, 3 and 4 in Figure~\ref{fig:cluster1truePred}, ~\ref{fig:cluster2truePred}, ~\ref{fig:cluster3truePred} and ~\ref{fig:cluster4truePred}, respectively. The diagonal line on the plots represents the values where the predicted is equal to the actual MM/GBSA score. Data points that are far away from the diagonal have large prediction errors. We examine the results for each cluster in the following.

The majority of the chemical compounds in cluster 1 have MM/GBSA scores ranging from around -15 to -35, indicating weaker binding. It is reasonable that larger prediction errors fall on the stronger binding side, ranging from -30 to -40 as shown in Figure~\ref{fig:cluster1truePred}. This range is near the tail of the training set. Predicted values that fall outside the training set can be less reliable. The AD-LD, MC-dropout and RIO uncertainty values are high for those less reliable predictions for the model trained with MOE descriptors, while only the AD-DD and AD-LD values are high for ECFP. For the model trained on cluster 1 with ECFP the model uncertainty from MC-dropout appears opposite to the desired values, rendering low values on the high prediction error ranges and high values on the low prediction error ranges. For the same model the RIO method gives relatively high evaluations only to a few predicted values at the two ends of the predicting range, around -20 and -35. 

Cluster 2 has strong binding free energy with the MM/GBSA scores approximately ranging from -25 to -55 as the majority. For the model trained on cluster 2 with MOE descriptors the AD-DD and AD-LD methods give relatively high values on the chemical compounds that also have large prediction errors as shown in Figure~\ref{fig:cluster2truePred}. For the same model with the MOE descriptors, the model uncertainties from MC-dropout are moderately high on the lower half of the prediction values, while the RIO uncertainties are distinctly higher at the points far away from the diagonal. The four uncertainty evaluation methods do not capture any prediction errors for the model trained on cluster 2 with the ECFP features, whose training set performance was poor with $R^2$ at 0.412. The model predicts strong binding for all data points, and has a small range of prediction values, from -32 to -44.  

The chemical compounds in cluster 3 and cluster 4 cover a wider range of MM/GBSA scores than cluster 1 and cluster 2. Between the two clusters, cluster 3 contains more high value MM/GBSA scores (meaning weaker binding) while cluster 4 contains more low value MM/GBSA scores (meaning stronger binding). The models trained on cluster 3 have the best performance on the test set prediction. This also reflects on the actual versus predicted plots in Figure~\ref{fig:cluster3truePred} where most data points are along the diagonal. AD-DD, AD-LD and RIO give high values at the two ends of the predicting range for the model trained with MOE descriptors. RIO uncertainty gives distinct large uncertainty values at the lowest predicted MM/GBSA scores and only a few at the highest, reflecting the uncertainty due to the lack of low MM/GBSA scores in the training set. For the models trained on cluster 3 model uncertainties from MC-dropout are high at the lower half of the predicted range with MOE descriptors, but high at the highest predicted values with ECFP. The model trained with ECFP again does not perform well and has a narrow range of predicted values, ranging from -30 to -43. The AD-DD and AD-LD values are spread out while the RIO uncertainties are close to a constant for the model trained with ECFP.   

The model trained on cluster 4 with MOE descriptors predicts high MM/GBSA scores with large errors. All four methods successfully indicate these high prediction errors. Particularly the RIO method gives distinctively high uncertainties on the predictions larger than -20 where the large errors occur. Similar to the model trained on cluster 3, the model trained with ECFP does not perform well and has a narrow range of prediction values, ranging from -30 to -43. The AD-DD and AD-LD values are spread out. The model uncertainties are high at the high predicted values. The RIO uncertainties are close to a constant for most chemical compounds except the lowest and the highest ends of the predicted values.

\begin{figure}[h!]
\begin{center}
\small
\begin{tabular}{cccc}
\multicolumn{4}{c}{MOE}\\
\multicolumn{1}{c}{(A) MC dropout} & \multicolumn{1}{c}{(B) RIO} & \multicolumn{1}{c}{(C) RIO} & \multicolumn{1}{c}{(D) $y_i-\hat{y_i}$} \\
\includegraphics[width = 0.23\columnwidth, trim=0.cm 0.1cm 0.cm 0.cm, clip=true]{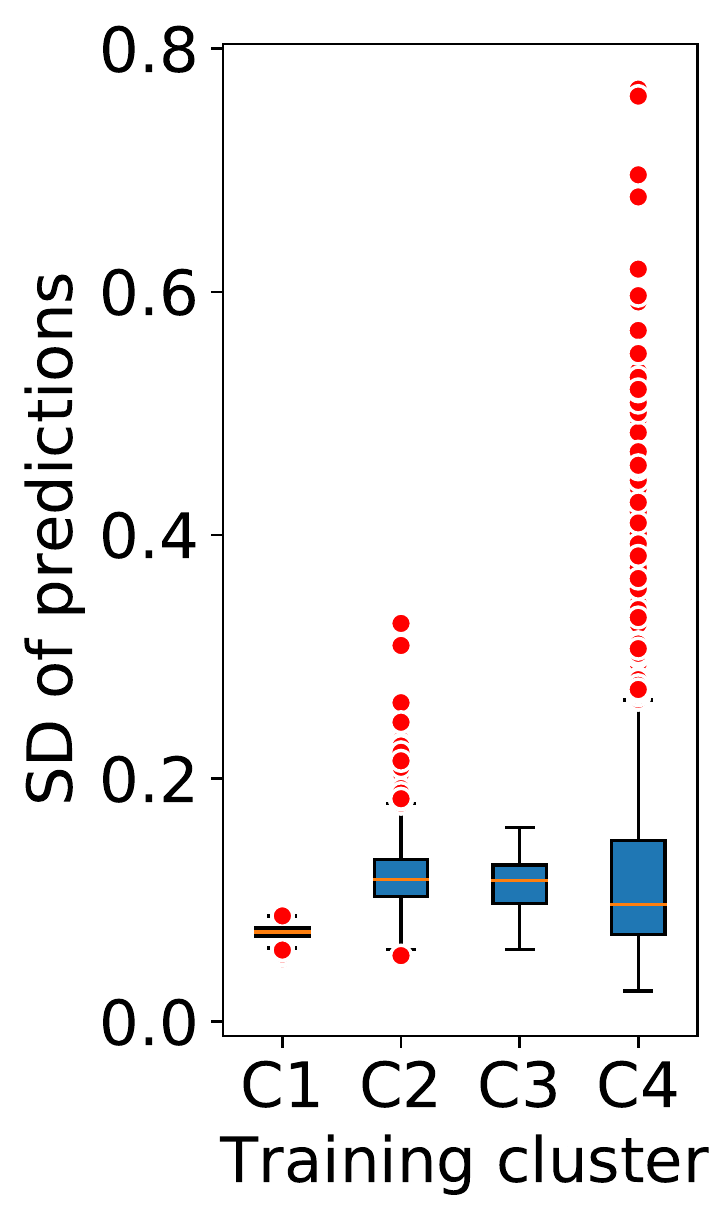}
&\includegraphics[width = 0.23\columnwidth, trim=0.cm 0.1cm 0.cm 0.cm, clip=true]{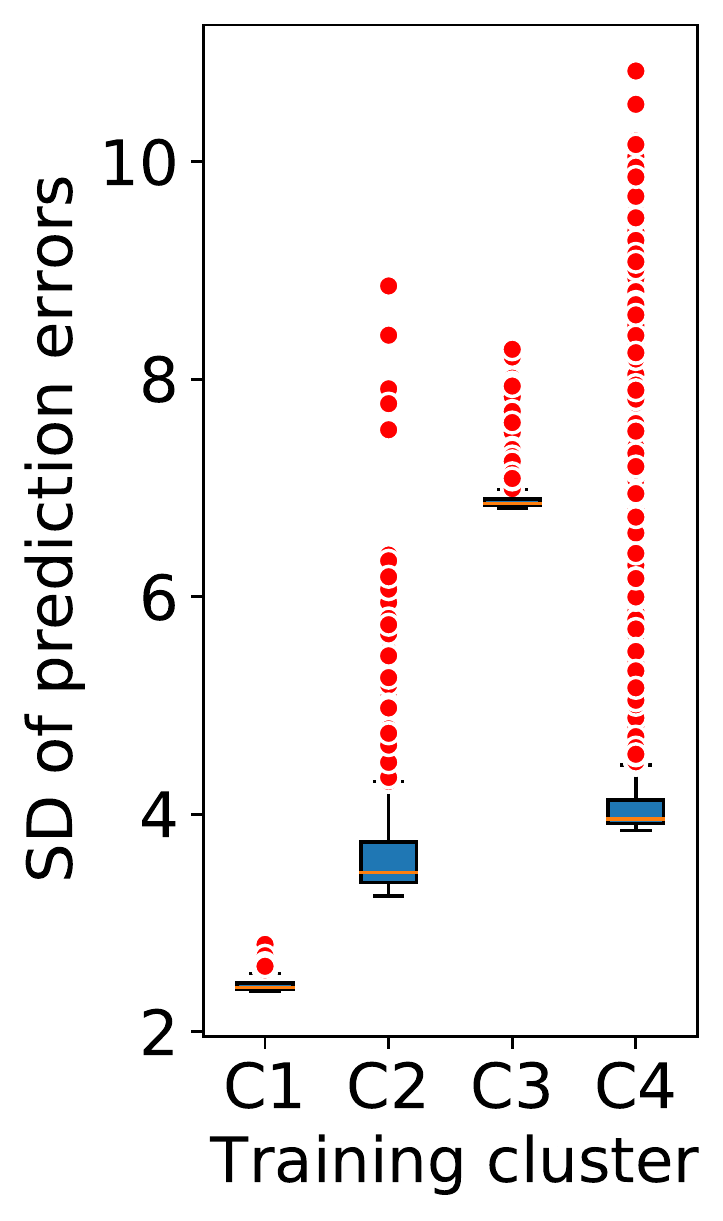}
&\includegraphics[width = 0.23\columnwidth, trim=0.cm 0.1cm 0.cm 0.cm, clip=true]{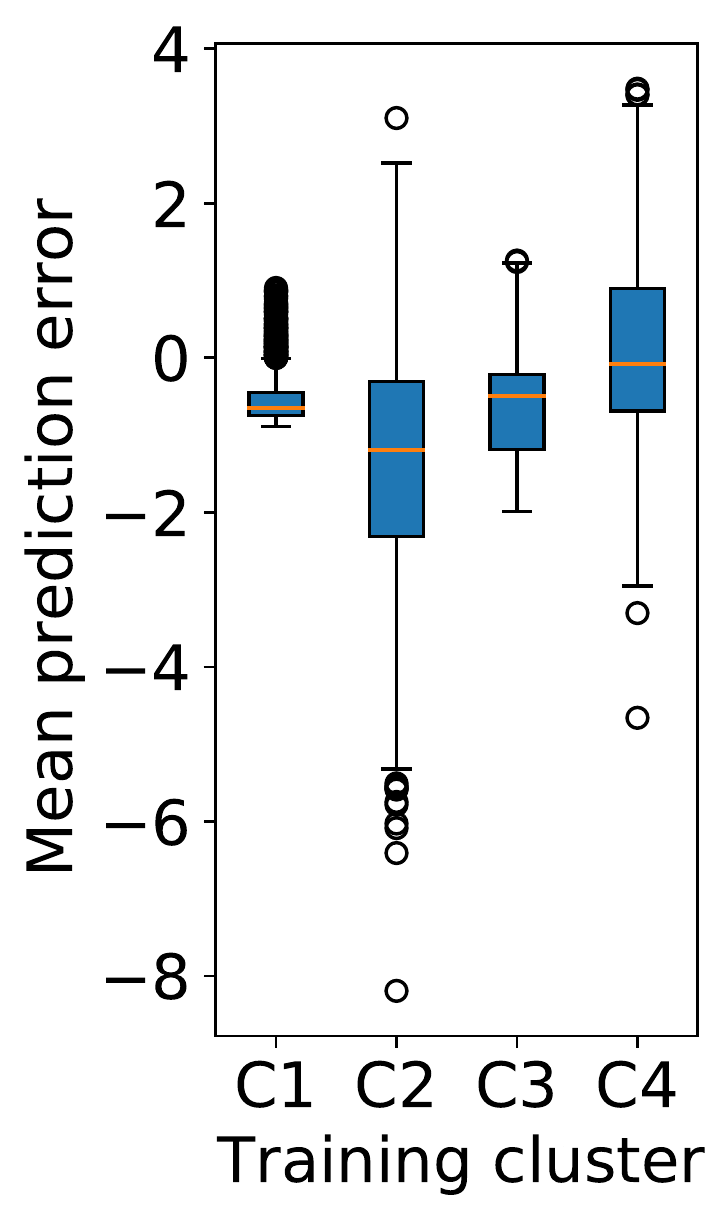}
&\includegraphics[width = 0.23\columnwidth, trim=0.cm 0.1cm 0.cm 0.cm, clip=true]{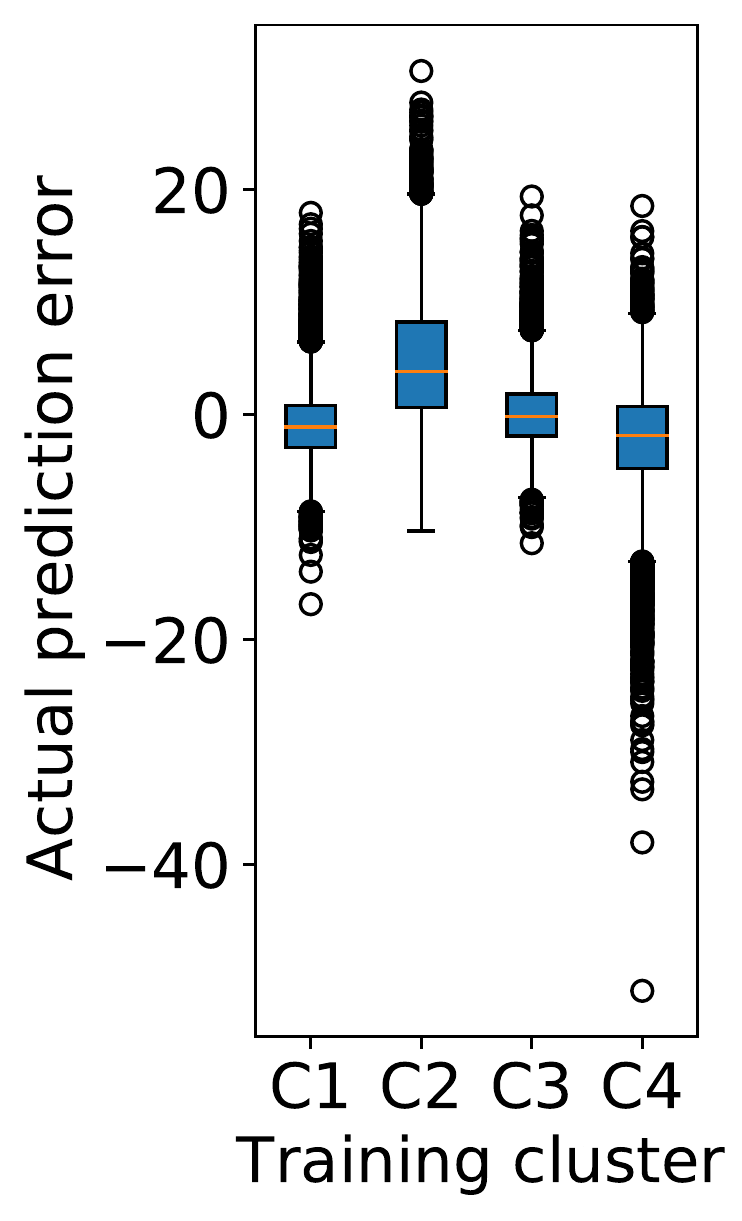}\\
\multicolumn{4}{c}{ECFP}\\
\multicolumn{1}{c}{(A) MC dropout} & \multicolumn{1}{c}{(B) RIO} & \multicolumn{1}{c}{(C) RIO} & \multicolumn{1}{c}{(D) $y_i-\hat{y_i}$} \\
\includegraphics[width = 0.25\columnwidth, trim=0.cm 0.1cm 0.cm 0.cm, clip=true]{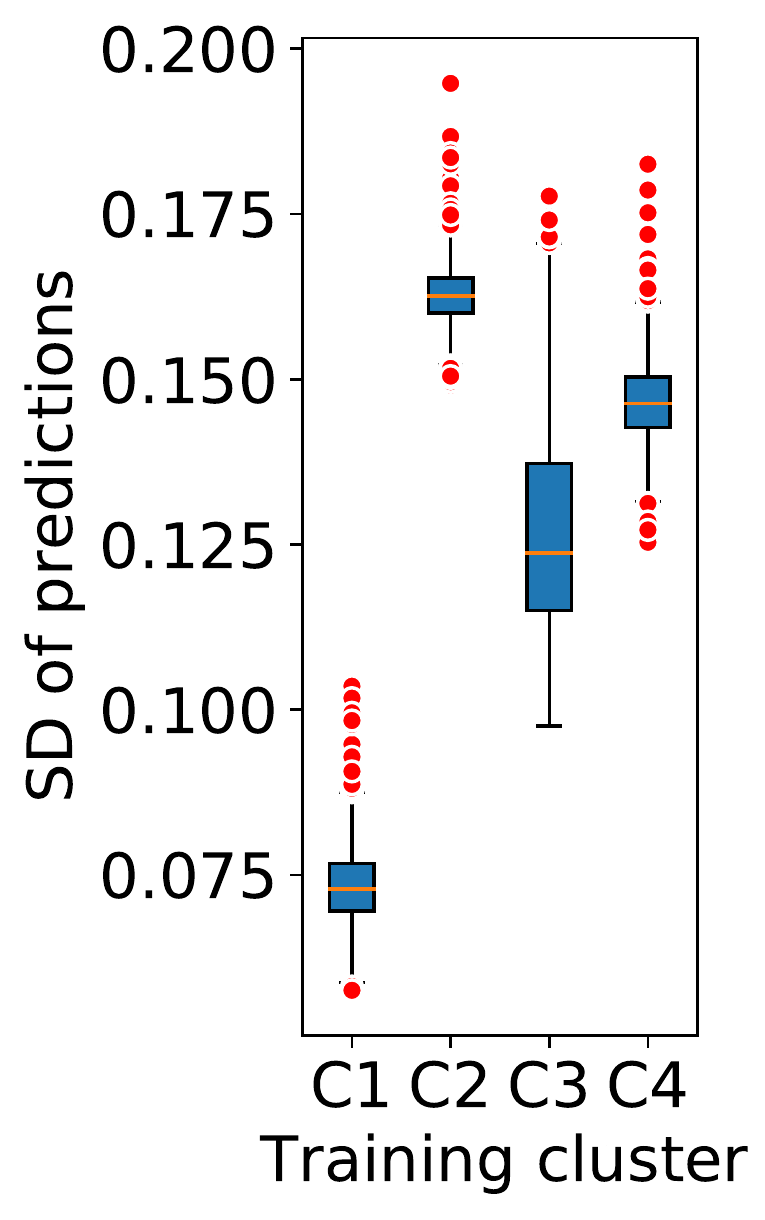}
&\includegraphics[width = 0.23\columnwidth, trim=0.cm 0.1cm 0.cm 0.cm, clip=true]{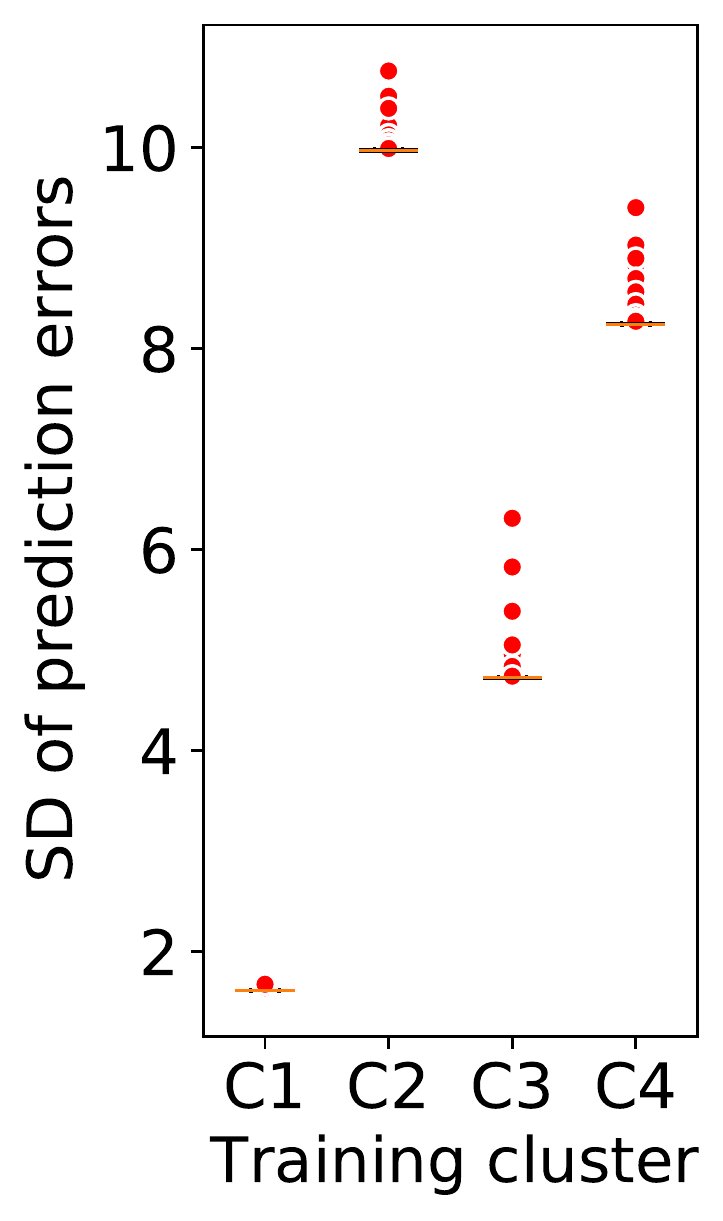}
&\includegraphics[width = 0.23\columnwidth, trim=0.cm 0.1cm 0.cm 0.cm, clip=true]{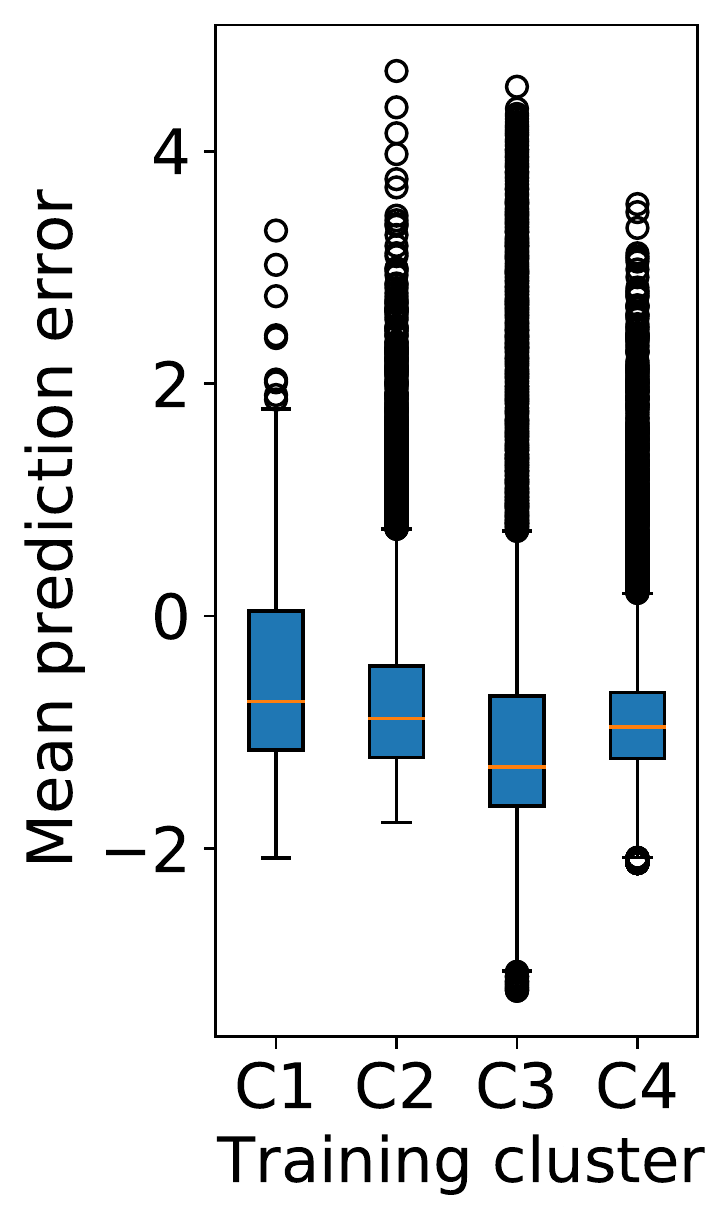}
&\includegraphics[width = 0.23\columnwidth, trim=0.cm 0.1cm 0.cm 0.cm, clip=true]{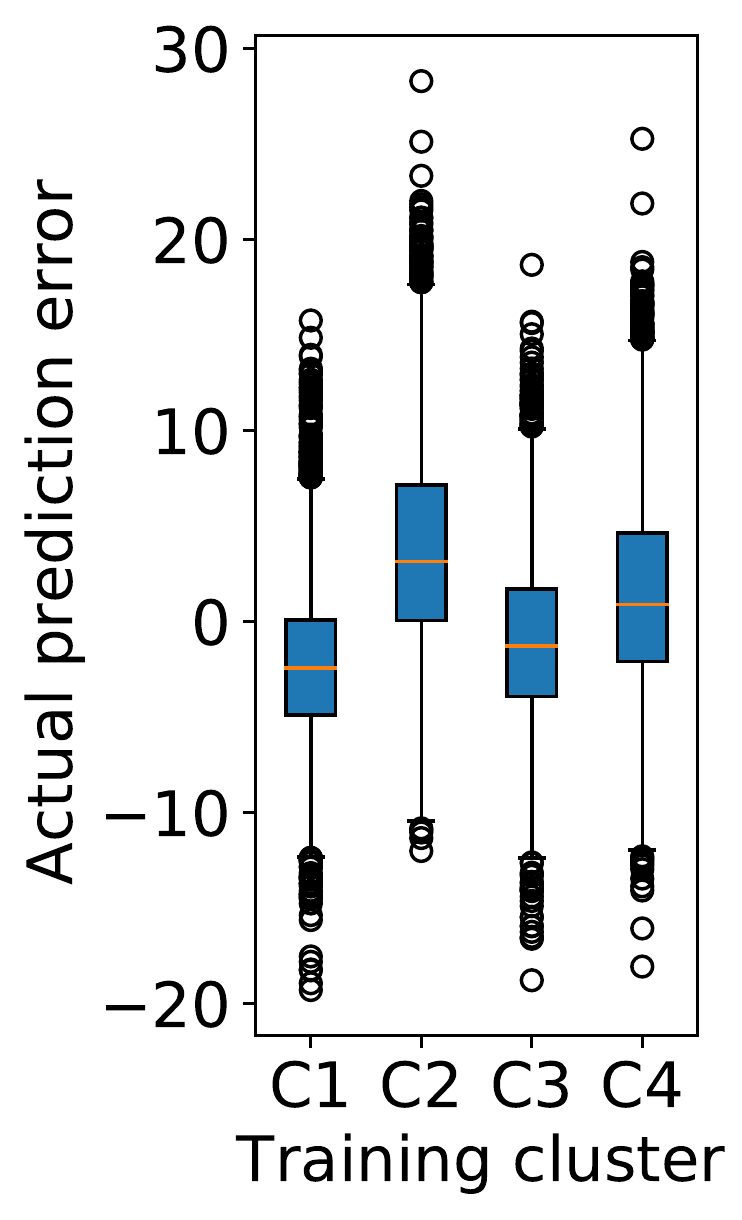}
\end{tabular}
\end{center}
\vspace{-0.17in}
\caption{\label{fig:boxPlots} Box plots for the test set uncertainties and prediction errors from the models trained with MOE features and ECFP features.The first column C1 indicates the model trained on cluster 1 and the corresponding data points in the box plot are from the test set, composed of cluster 2, 3 and 4. (A) MC dropout uses the standard deviation of the predictions as its uncertainty estimation for the NN point prediction. For all the test sets MC dropout gives the uncertainty values in a narrow range. (B) RIO provides the standard deviation of the prediction errors (the residuals) to the NN point prediction. (C) RIO also estimates the mean of the prediction errors to the NN point prediction. (D) The actual prediction error is the actual MM/GBSA score subtracted by the NN predicted value. The actual prediction errors are close to zero with some outliers that have large errors. Although RIO estimates most mean prediction errors close to zero, it gives high standard deviations to the predictions made by the models trained on cluster 2, 3 and 4, indicating high uncertainties in some data points.}
\end{figure}
%

\begin{comment}
%
\begin{figure}[htb]
\begin{center}
\small
\begin{tabular}{cccc}
\multicolumn{1}{c}{(A) MC dropout} & \multicolumn{1}{c}{(B) RIO} & \multicolumn{1}{c}{(C) RIO} & \multicolumn{1}{c}{(D) $y_i-\hat{y_i}$} \\
\includegraphics[width = 0.25\columnwidth, trim=0.cm 0.1cm 0.cm 0.cm, clip=true]{GBSA_ecfp_dropout_var.pdf}
&\includegraphics[width = 0.23\columnwidth, trim=0.cm 0.1cm 0.cm 0.cm, clip=true]{GBSA_ecfp_rio_var.pdf}
&\includegraphics[width = 0.23\columnwidth, trim=0.cm 0.1cm 0.cm 0.cm, clip=true]{GBSA_ecfp_rio_mean.pdf}
&\includegraphics[width = 0.23\columnwidth, trim=0.cm 0.1cm 0.cm 0.cm, clip=true]{GBSA_ecfp_dropout_actualError.pdf}
\end{tabular}
\end{center}
\vspace{-0.17in}
\caption{\label{fig:uqECFP} Box plots for the test set uncertainties and prediction errors from the models trained with ECFP features.}
\end{figure}
%
\end{comment}

%\caption{\label{fig:rio} Box plot for RIO estimated mean and standard deviation of the prediction errors for the point predictions in the test set. The x-axis indicates the clusters the model trained on. The y-axis shows the mean and standard deviation estimated using RIO on the point predictions of the test set. The two plots on the left are from models trained with MOE descriptors, and the two on the right are with ECFP. The UQ value that RIO estimates for a point prediction is the standard deviation of the prediction error. Models trained on cluster 2 and cluster 4 do not perform well on their test set predictions, and RIO gives high uncertainty estimations to most of them.}
 
Although RIO gives almost all predictions a near constant uncertainty estimation, we observed that these constants reflect possible prediction errors. Figure~\ref{fig:boxPlots} display the box plots for the test set uncertainties and prediction errors. Figure~\ref{fig:boxPlots}(C) and Figure~\ref{fig:boxPlots}(B) contain the estimated mean and standard deviation of the prediction errors from the Gaussian Processes used in RIO. On the x-axis are the trained models where the first column C1 indicates the model trained on cluster 1. The corresponding data points in the box plot are from C1's test set, composed of cluster 2, 3 and 4. The UQ value that RIO estimates for a point prediction is the standard deviation of the prediction errors. RIO gives high uncertainty values to most of their test set predictions except C1 with MOE descriptors, reflecting the possibility of large test set prediction failures.

\begin{figure}[htb]
\begin{center}
\small
\begin{tabular}{cccc}
\multicolumn{4}{c}{MOE}\\
\includegraphics[width = 0.23\columnwidth, trim=0.cm 0.2cm 0.3cm 0.2cm, clip=true]{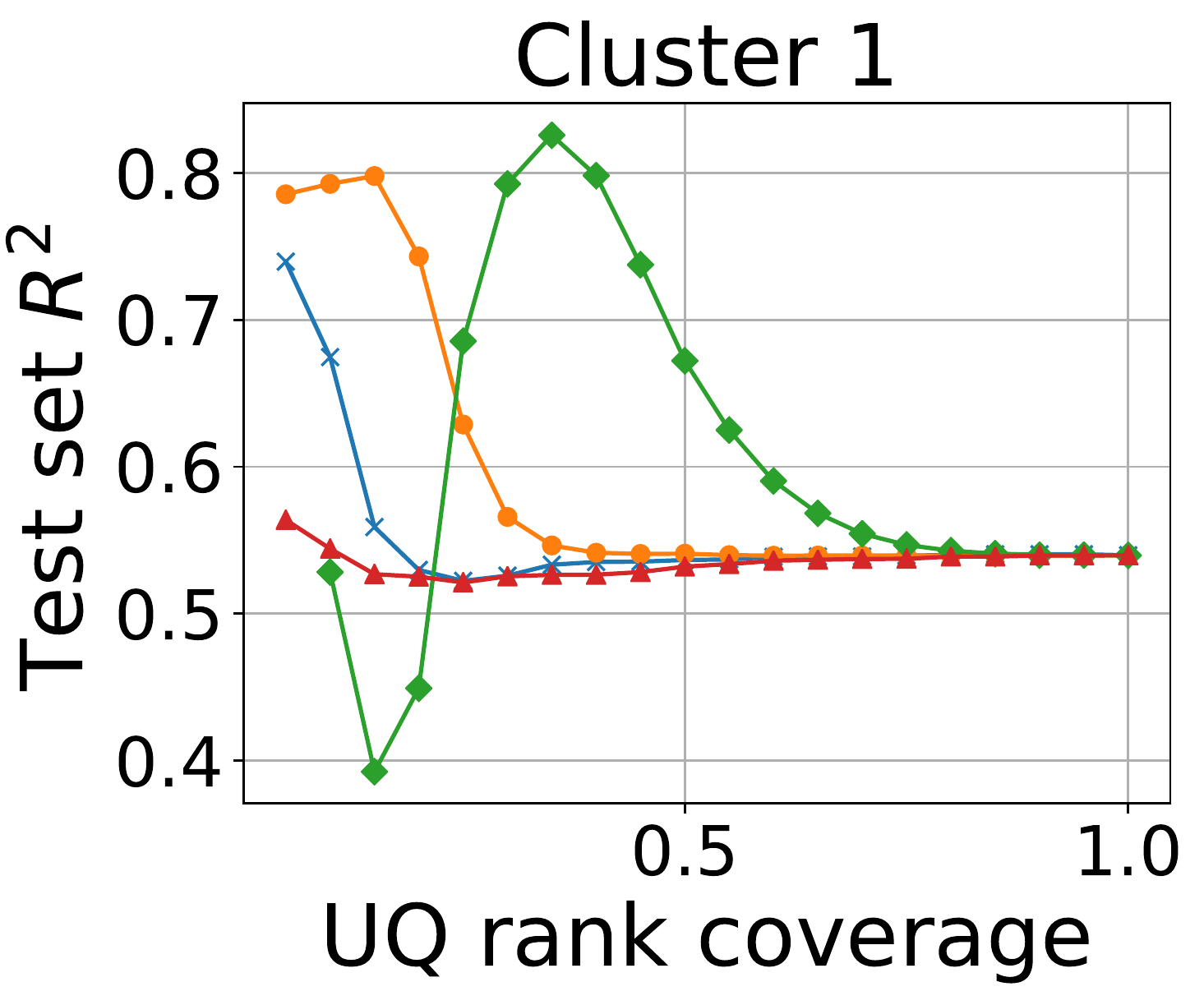}
&\includegraphics[width = 0.23\columnwidth, trim=0.cm 0.2cm 0.3cm 0.2cm, clip=true]{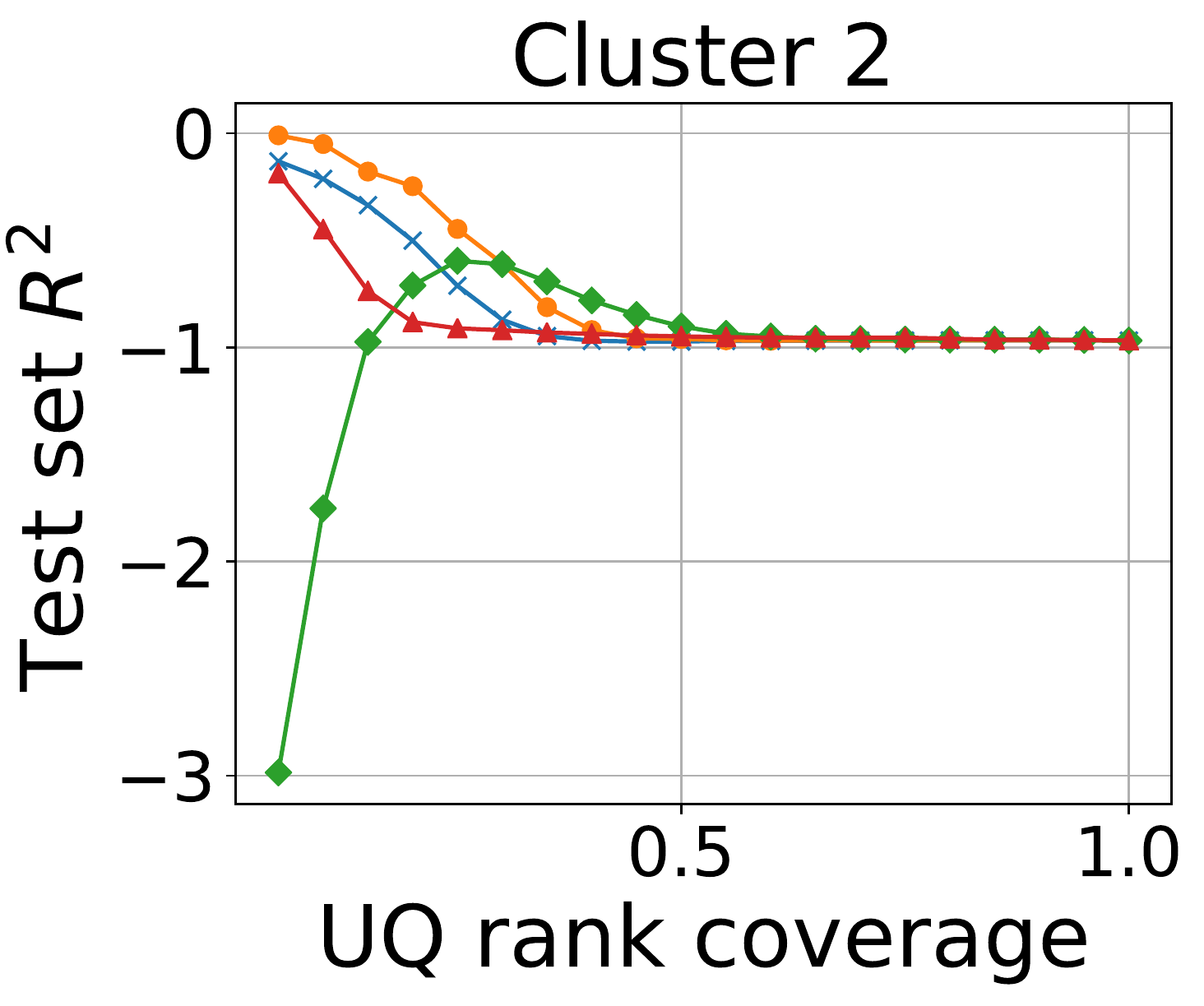}
&\includegraphics[width = 0.23\columnwidth, trim=0.cm 0.2cm 0.3cm 0.2cm, clip=true]{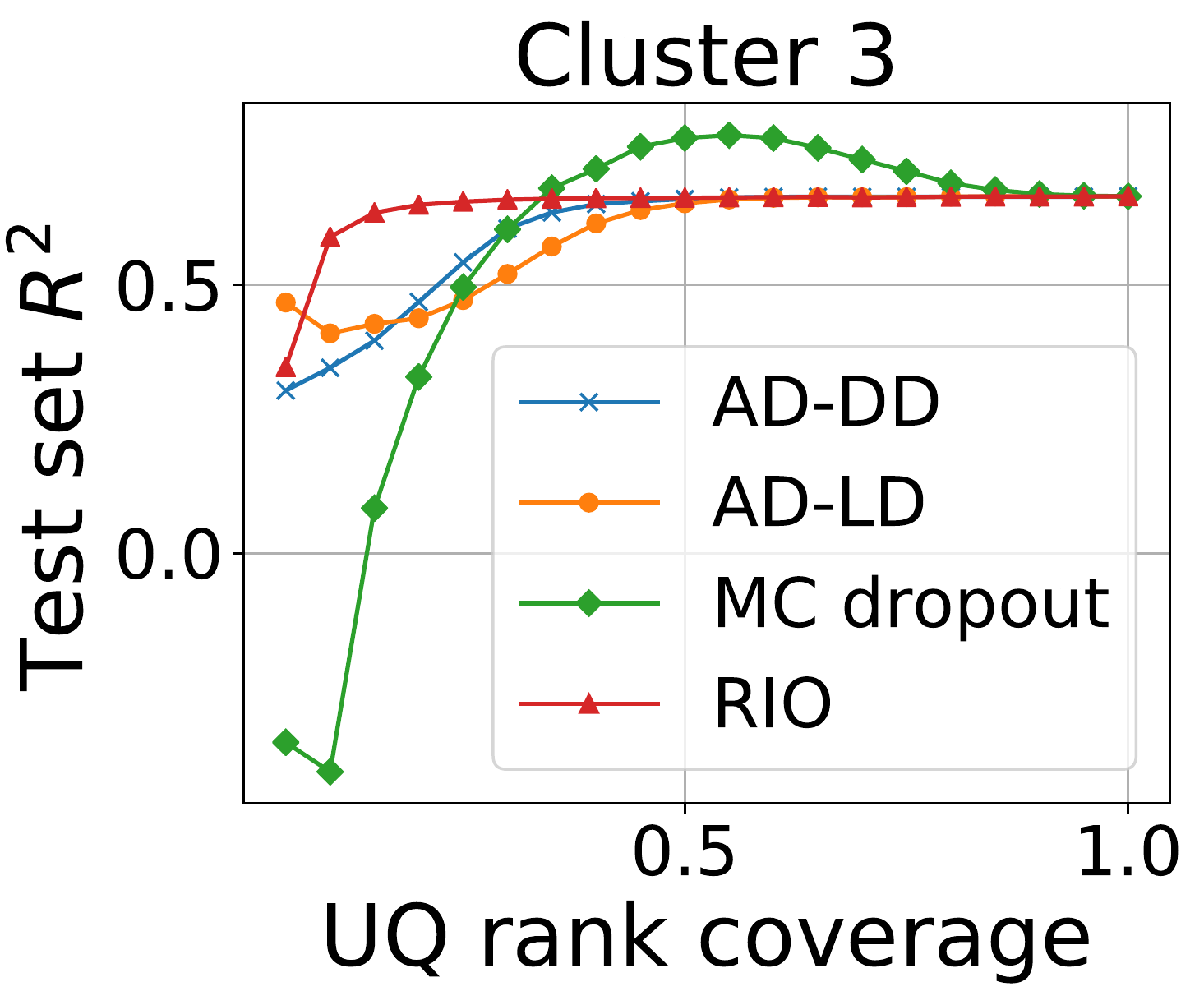}
&\includegraphics[width = 0.23\columnwidth, trim=0.cm 0.2cm 0.3cm 0.2cm, clip=true]{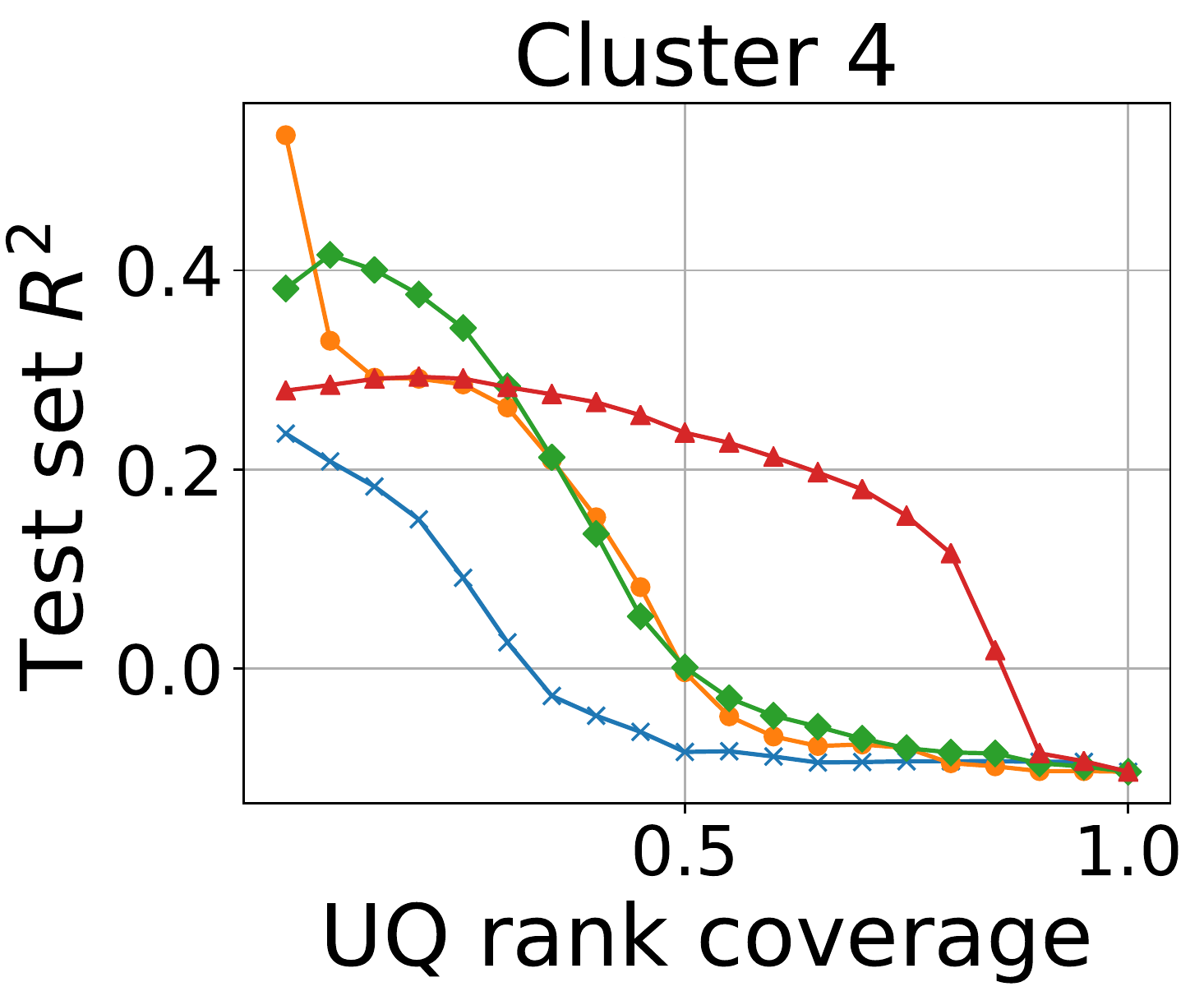}\\
\multicolumn{4}{c}{ECFP}\\
\includegraphics[width = 0.23\columnwidth, trim=0.cm 0.2cm 0.3cm 0.2cm, clip=true]{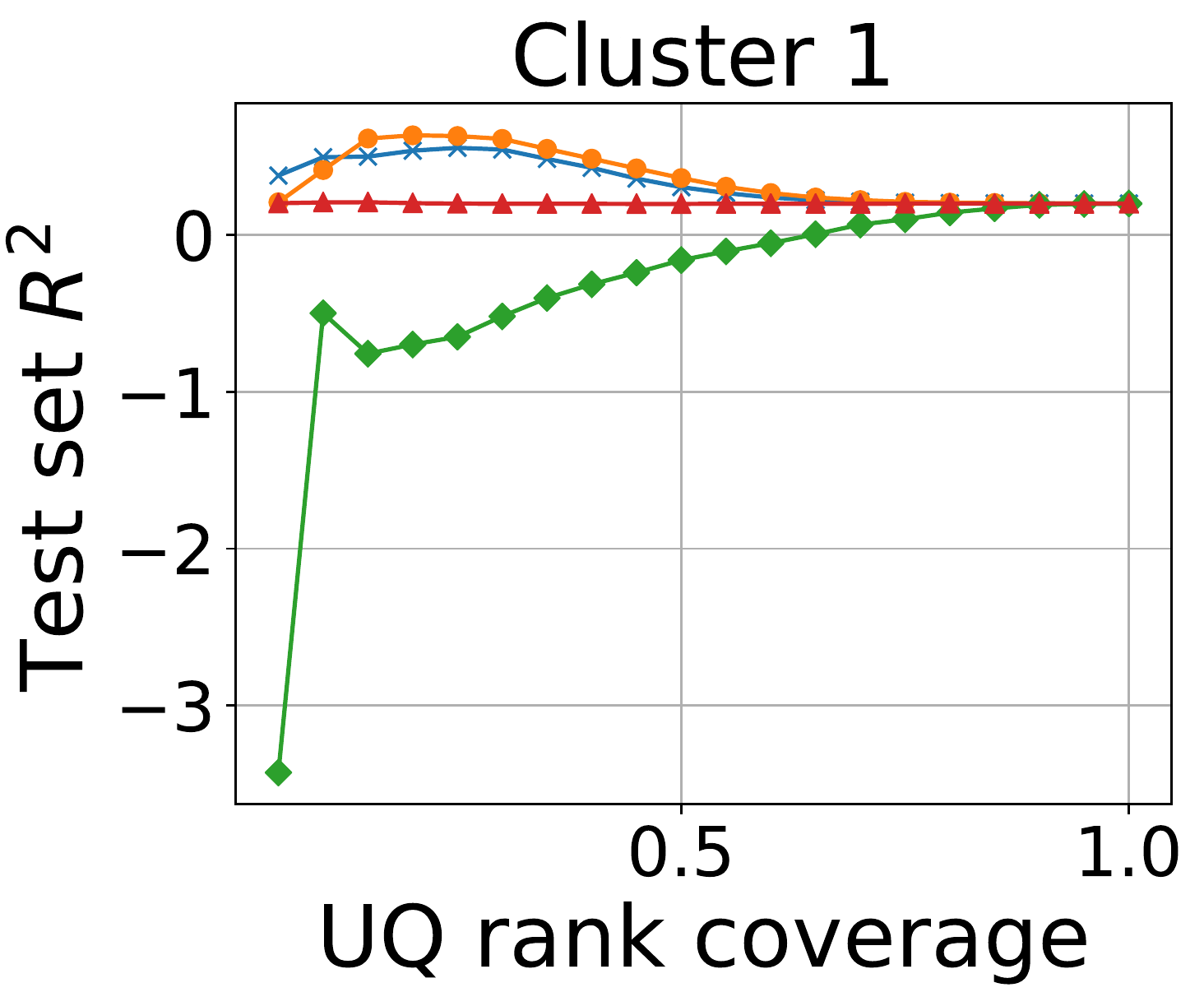}
&\includegraphics[width = 0.23\columnwidth, trim=0.cm 0.2cm 0.3cm 0.2cm, clip=true]{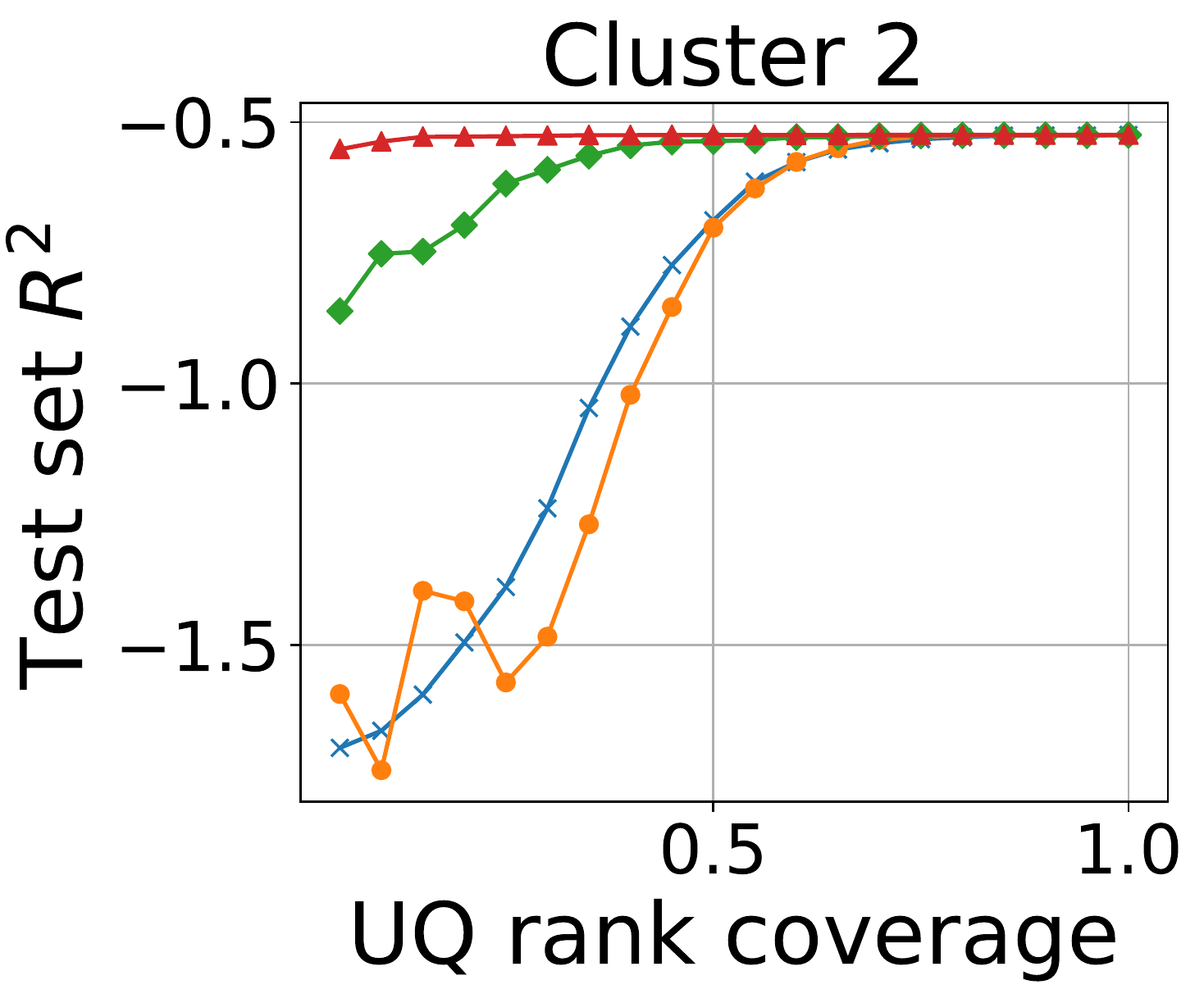}
&\includegraphics[width = 0.23\columnwidth, trim=0.cm 0.2cm 0.3cm 0.2cm, clip=true]{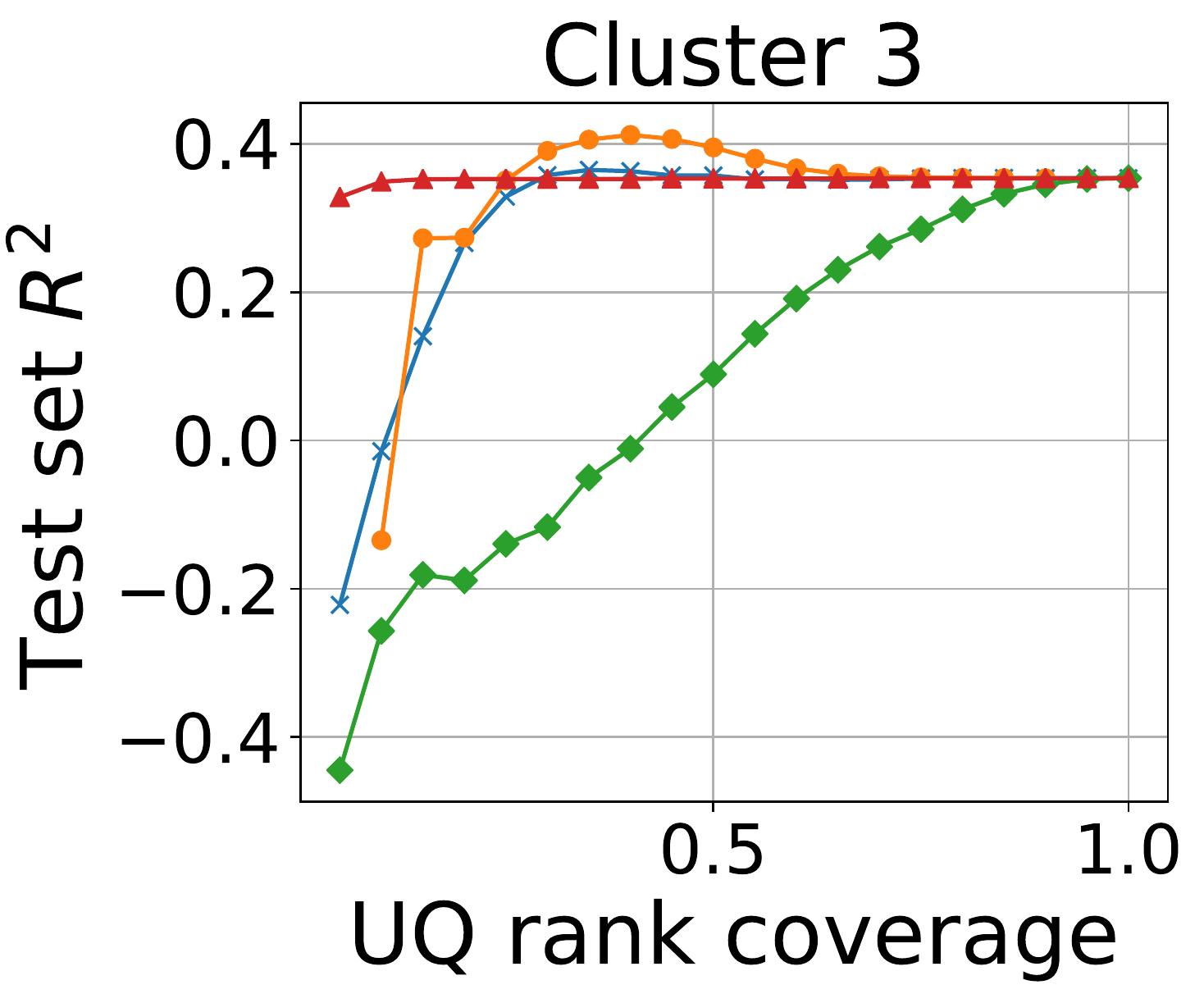}
&\includegraphics[width = 0.23\columnwidth, trim=0.cm 0.2cm 0.3cm 0.2cm, clip=true]{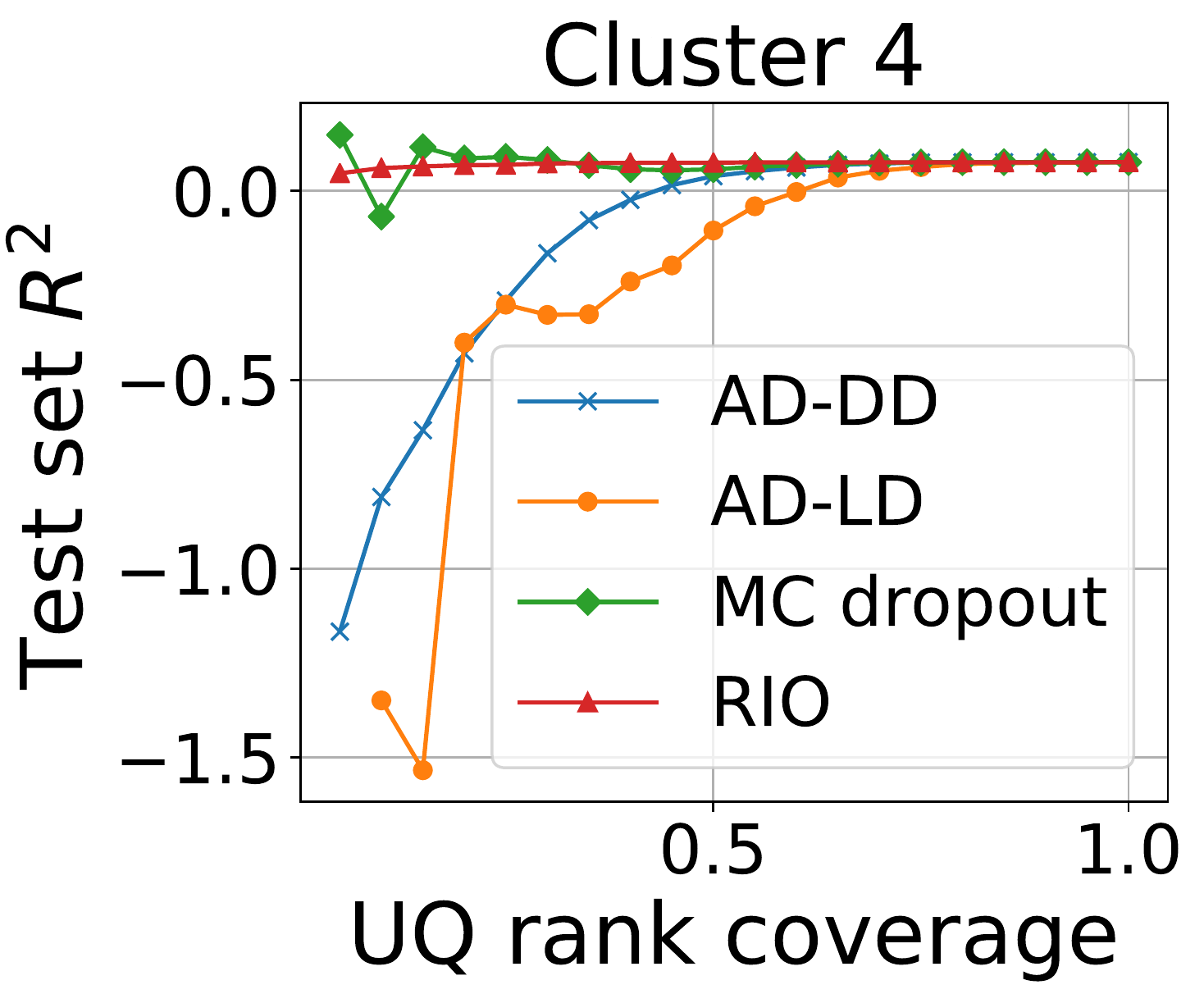}
\end{tabular}
\end{center}
\vspace{-0.15in}
\caption{\label{fig:uqPercent} Ranked UQ values versus the prediction performance on the test sets from the models trained with MOE and ECFP features.}
\end{figure}

Remarkably by ordering the uncertainty values and examining the test set performance, we found that the magnitude of the uncertainties does not necessarily reflect the level of prediction errors. We sort the uncertainty values in a decreasing order in the test sets for each of the UQ methods. We compute the test set coefficient of determination (also known as the $\text{R}^2$ score) and remove the top 5\% largest uncertainty values in the test set, iteratively, until no samples are left. These computed values form the performance curves shown in Figure~\ref{fig:uqPercent}. If the magnitude of the uncertainty values reflect the level of the prediction errors, the test set performance will increase as the top 5\% largest uncertainty values are removed from the test set. The model trained on cluster 4 with MOE descriptors is the only one where all four UQ methods are able to reflect the level of prediction errors. These UQ methods evaluating other NN models barely echo the level of prediction errors.

\clearpage

%======================
%
\section{Conclusion}
\label{sec:conclusion}
%
%======================

In this article, we investigate selected uncertainty quantification (UQ) methods that provides estimates of different sources of predictive uncertainty for neural networks (NN) aimed at drug discovery. The applicability domain (AD) methods capture distributional uncertainties. MC-dropout estimates model uncertainties. RIO perceives NN behavior by estimating the NN prediction residuals. The selected methods can be directly applied to any standard NN without having to modify the model formulation or training procedure, making them more accessible to analysts. 

To test how these UQ methods work on NN model predictions for drug discovery tasks, we carry out a series of carefully designed experiments. We randomly select drug compounds in the form of SMILES strings from public available database. We rescore the binding affinities of the chemical compounds by employing the molecular mechanics generalized Born surface area (MM/GBSA) continuum solvent approach. Since the MM/GBSA scores are calculated and not from the experimental measurements, it reduces the otherwise irreducible aleatoric uncertainty in the prediction. We use two featurization schemes, the MOE descriptors and the ECFP4 bit vectors, as model features. We design the splits of the training and test set using our prior knowledge on the correlation between MOE descriptors and the MM/GBSA scores. The visualization algorithm, t-distributed stochastic neighbor embedding (t-SNE), gives four visible clusters on the subset of MOE features that are highly correlated to the MM/GBSA scores. The clusters contain different distributions of the MM/GBSA scores. We take each cluster as a training set and the rest of the chemical compounds as the corresponding test set. We apply hyperparameter search to build optimal NN models. Finally, the selected UQ methods provide uncertainty values for the point predictions made by the trained NN models.    

Our experiments show that the selected UQ methods describe different sources of uncertainty for point predictions of NN models. The AD method using distance distribution (AD-DD) retain the distance scale of the original data space, which the projected 2-dimensional t-SNE space has lost. Our results show that the AD-DD method can highlight novel chemical compounds, which gives the data in the other clusters higher uncertainties. It can also identify non-clustered data in the t-SNE plot, such as the cluster 4. When using cluster 4 as the training cluster, AD-DD gives high uncertainty values to several data points in the cluster, indicating the chemical compounds are very different from each other. The AD method using local density (AD-LD) verifies the clarity of the training set clusters. 

MC-dropout points to the sensitivity of the training parameters to the final predictions. MC-dropout can give opposite levels of uncertainty estimations when using different compound representations (i.e. MOE or ECFP). Our results show that MC-dropout is more reliable with MOE descriptors than with ECFP, possibly due to the better trained models with MOE descriptors. MC-dropout can reflect possible prediction errors on some of the predictions made by the models trained with MOE descriptors on cluster 1 and cluster 3. However, MC-dropout provide high uncertainty values to the predictions in the training clusters where the other methods tend to give low values. 

RIO method gives near constant uncertainty values to the majority of chemical compounds. We observed that these constants reflect possible prediction errors made by poorly performing models. When the model is not performing well, the RIO gives high near-constant uncertainty values to most predictions. Only in the case where the chemical compounds are very different from the applicable domain and at the same time the model causes large prediction errors in the test set, RIO gives high uncertainty values to the data points that are distinctly larger than the near-constant uncertainty values. 

Moreover, we observed that not all UQ methods can capture high prediction errors invariably. Only for cluster 4 with MOE descriptors do all four models capture the high prediction errors. Furthermore, the magnitude of the uncertainties does not clearly reflect the level of prediction errors. Again, only the models trained on cluster 4 with MOE descriptors form increasing performance curves when the top 5\% largest uncertainty values are removed iteratively. Generally, the AD methods can only detect the novelty of the chemical compounds without any knowledge from the NN model. MC-dropout reflects the prediction variance due to changes in training parameters. RIO gives extra high uncertainties to extreme predicted values. These scenario do not always come with high prediction errors. 

The differences in the uncertainty estimations made by the selected UQ methods provide more insight into the behavior of NN models for drug discovery. We suggest to model the different types of predictive uncertainty separately. Knowing the assorted types of uncertainty in a point prediction may assist practitioners in taking an appropriate UQ aware selection strategy.

%======================
%
\section*{Acknowledgment}
%
%======================
LLNL-JRNL-839676. This work represents a multi-institutional effort and is supported by the Accelerating Therapeutics for Opportunities in Medicine (ATOM) Consortium under CRADA TC02349.9. Funding sources include the following: Lawrence Livermore National Laboratory internal funds; the National Nuclear Security Administration; DTRA under Award HDTRA1036045 and federal funds from the National Cancer Institute, National Institutes of Health, and the Department of Health and Human Services, Leidos Biomedical Research Contract No. 75N91019D00024, Task Order 75N91019F00134. This work was performed under the auspices of the U.S. Department of Energy by Lawrence Livermore National Laboratory [Contract No. DE-AC52-07NA27344].

%\pagebreak
%\bibliographystyle{acm}
\bibliographystyle{plain}
\bibliography{pdf_rio_uq}  
%\pagebreak
\end{document}